\documentclass[]{acmart}
\AtBeginDocument{%
  }

\setcopyright{acmlicensed}
\copyrightyear{2026}
\acmYear{2026}
\acmDOI{XXXXXXX.XXXXXXX}

\usepackage[utf8]{inputenc} 
\usepackage{url}            
\usepackage{amsfonts}       
\usepackage{nicefrac}       
\usepackage{subcaption}
\usepackage{multirow}
\usepackage{array}

\begin{document}

\title{A Survey on Human-AI Collaboration with Large Foundation Models}

\author{Vanshika Vats}
\authornote{Shared First Authorship}
\email{vvats@ucsc.edu}
\author{Marzia Binta Nizam}
\authornotemark[1]
\email{manizam@ucsc.edu}

\author{Minghao Liu}
\author{Ziyuan Wang}
\authornote{Equal Contribution}
\author{Richard Ho}
\authornotemark[2]

\author{Mohnish Sai Prasad}
\authornotemark[2]

\author{Vincent Titterton}
\authornotemark[2]

\author{Sai Venkat Malreddy}
\authornotemark[2]

\author{Riya Aggarwal}
\authornotemark[2]

\author{Yanwen Xu}
\authornotemark[2]

\author{Lei Ding}
\authornotemark[2]

\author{Jay Mehta}
\authornotemark[2]

\author{Nathan Grinnell}
\authornotemark[2]

\author{Li Liu}
\authornotemark[2]

\author{Sijia Zhong}
\authornotemark[2]

\author{Devanathan Nallur Gandamani}
\authornotemark[2]

\author{Xinyi Tang}
\authornotemark[2]

\author{Rohan Ghosalkar}
\authornotemark[2]

\author{Celeste Shen}
\authornotemark[2]

\author{Rachel Shen}
\authornotemark[2]

\author{Nafisa Hussain}
\authornotemark[2]

\author{Kesav Ravichandran}
\authornotemark[2]

\author{James Davis}
\affiliation{%
  \institution{University of California, Santa Cruz}
  \city{Santa Cruz}
  \state{California}
  \country{USA}}

\renewcommand{\shortauthors}{Vats and Nizam et al.}

\begin{abstract}
As the capabilities of artificial intelligence (AI) continue to expand rapidly, Human-AI (HAI) Collaboration, combining human intellect and AI systems, has become pivotal for advancing problem-solving and decision-making processes. The advent of Large Foundation Models (LFMs) has greatly expanded its potential, offering unprecedented capabilities by leveraging vast amounts of data to understand and predict complex patterns. At the same time, realizing this potential responsibly requires addressing persistent challenges related to safety, fairness, and control. This paper reviews the crucial integration of LFMs with HAI, highlighting both opportunities and risks. We structure our analysis around: human-guided model development, collaborative design principles, ethical and governance frameworks, and applications in high-stakes domains. Our review shows that successful HAI systems are not the automatic result of stronger models but the product of careful, human-centered design. By identifying key open challenges, this survey aims to give insight into current and future research that turns the raw power of LFMs into partnerships that are reliable, trustworthy, and beneficial to society.
\end{abstract}

\ccsdesc[500]{Computing methodologies~Artificial intelligence}
\ccsdesc[500]{Human-centered computing~Human computer interaction (HCI)}

\keywords{Human-AI Collaboration, Artificial Intelligence, Large Foundation Models, Large Language Models}


\acmJournal{TIST}
\acmYear{2026} 
\acmMonth{8} \acmDOI{10.1145/3841472}

\maketitle

\section{Introduction}
\begin{figure}
    \centering
    \begin{subfigure}[b]{0.49\textwidth}
        \includegraphics[width=\textwidth]{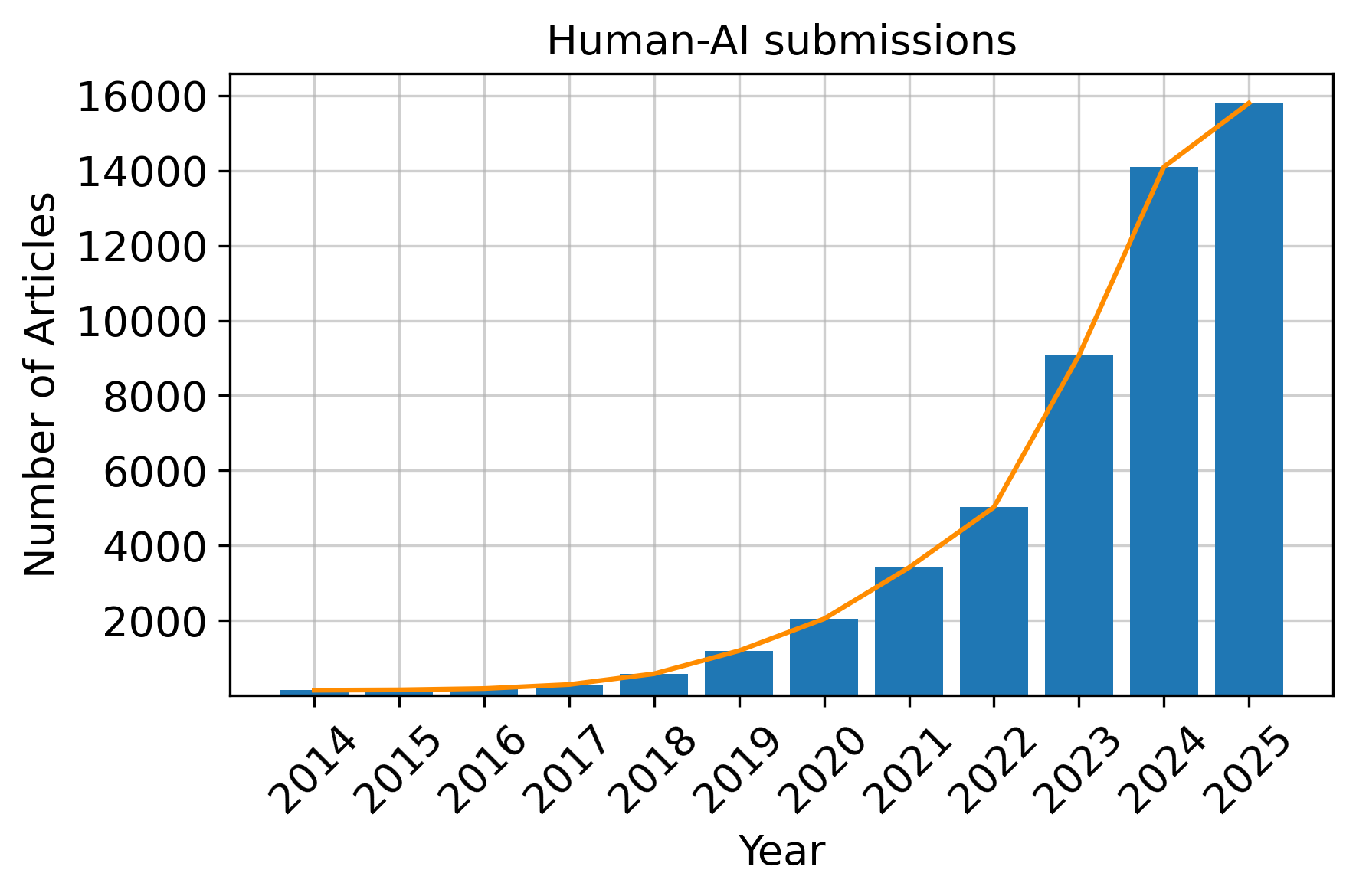}
        \caption{}
        \label{fig:HAI}
    \end{subfigure}
    \begin{subfigure}[b]{0.49\textwidth}
        \includegraphics[width=\textwidth]{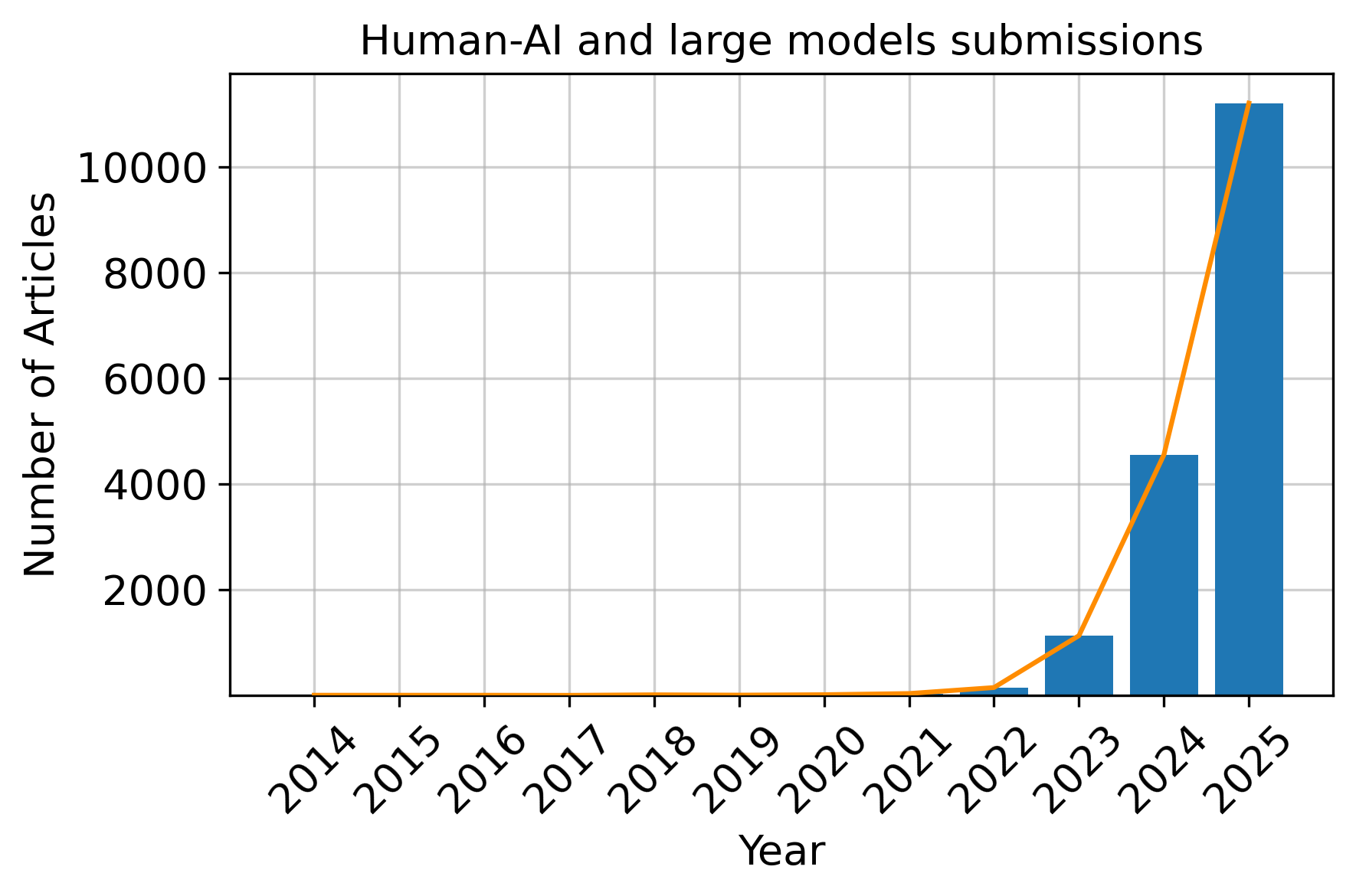}
        \caption{}
        \label{fig:HAI+LM}
    \end{subfigure}
    \caption{A graphical depiction of the number of relevant articles submitted from 2014-2025. (a) illustrates the increasing engagement of research communities in collaborations between humans and AI over the years. (b) observes a notably sharp increase in submissions related to human-AI and large-scale models attributed to the recent emergence of large models.}
    \Description[Two charts show the growth of Human-AI research from 2014 to 2025.]{Two side-by-side bar charts plot the number of articles by year from 2014 to 2025. The left chart shows Human-AI research increasing gradually through 2020 and then sharply after 2021, reaching nearly 16,000 articles in 2025. The right chart shows research combining Human-AI topics with large models remaining limited until 2022, followed by rapid growth in 2023-2025 and more than 11,000 articles in 2025.
}
    \label{fig:submissions}
\end{figure}

People have long been fascinated by the idea of creating machines that think like humans. One notable example from the 1770s is the "Mechanical Turk," a machine that appeared to play chess autonomously but was, in fact, operated by a person concealed inside it \cite{mech_turk}. This early effort, while not \emph{"Artificial Intelligence" (AI)} in the modern sense, reflects the enduring fascination with creating technology that can mimic or complement human cognitive abilities. The formal realization of integrating human-like intelligence into machines peaked in 1956 at Dartmouth College, USA, marking the official birth of Artificial Intelligence as a research field \cite{Russell2021}. As AI technology has advanced and become more prevalent over the last decade, researchers have identified limitations within purely automated systems \cite{mittelstadt2016, goodfellow2015, rodrigues2020} leading to renewed focus on augmenting AI with human expertise, aiming to utilize the best of both worlds (Fig. \ref{fig:HAI}). This approach seeks to enhance AI applications by combining automation with human judgment, creating a more symbiotic collaboration between human and artificial intelligence.

Navigating the rapidly evolving AI landscape, the integration of human cognition with Large Foundation Models (LFMs), including Large Language Models (LLM) \cite{Tom2020, touvron2023llama} and Large Vision Models (LVM) \cite{kirillov2023segany, pmlr-v139-radford21a, jia2022vpt}, has initiated an exciting shift. These models are pre-trained on vast web-scale datasets and act as "foundations" before being fine-tuned for specific tasks. This has opened new avenues for collaborative problem-solving and decision-making. When humans add domain insight, ethics, and creativity to these generic large models, they return rapid pattern-finding and specialized outputs at scale. In turn, AI amplifies human capabilities by processing data at large scales, offering insights, and augmenting decision-making. This interaction paves the way to creating an advanced form of HAI collaboration. In the LFM era, this collaboration spans both build-time settings, where humans shape what models learn, and run-time settings, where humans and AI systems divide initiative, control, and responsibility during task execution.

For this survey paper, we formally define the term `Human-AI Collaboration'. Human-AI (HAI) collaboration is a cooperative partnership in which humans and AI systems coordinate their complementary strengths, exchange information (and sometimes control), and pursue a shared goal, without the tight interdependence and joint accountability \cite{kamar2016directions, shneiderman2020human,WilsonDaugherty2018HBR}. Each partner contributes what it does best and iteratively refines the joint output through feedback. This paper undertakes an extensive survey of HAI collaboration, analyzing the complex interactions between human agents and sophisticated pre-trained models across various fields. Our study aims to uncover the progress made, confront the challenges, and understand the implications of this evolving partnership. To synthesize this literature, we use recurring collaboration patterns as a guiding lens: human steering of AI, AI augmentation of human work, mixed-initiative collaboration, dynamic delegation or control handoff, and emerging supervisory orchestration in agentic systems.

\begin{figure}[t]
    \centering
    \begin{subfigure}[b]{0.49\textwidth}
        \includegraphics[width=\textwidth]{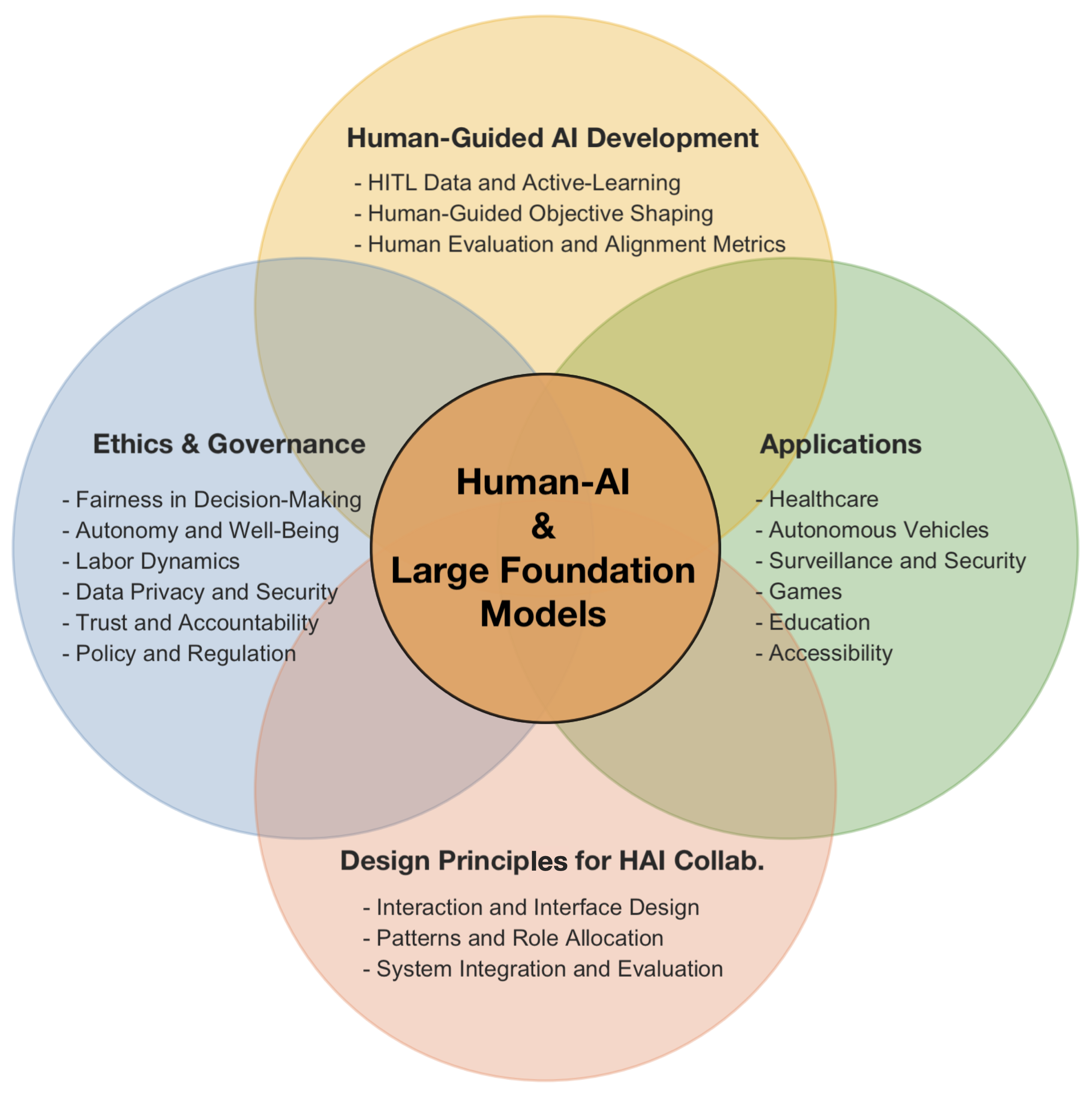}
        \caption{}
        \label{fig:scope}
    \end{subfigure}
    \begin{subfigure}[b]{0.49\textwidth}
        \includegraphics[width=\textwidth]{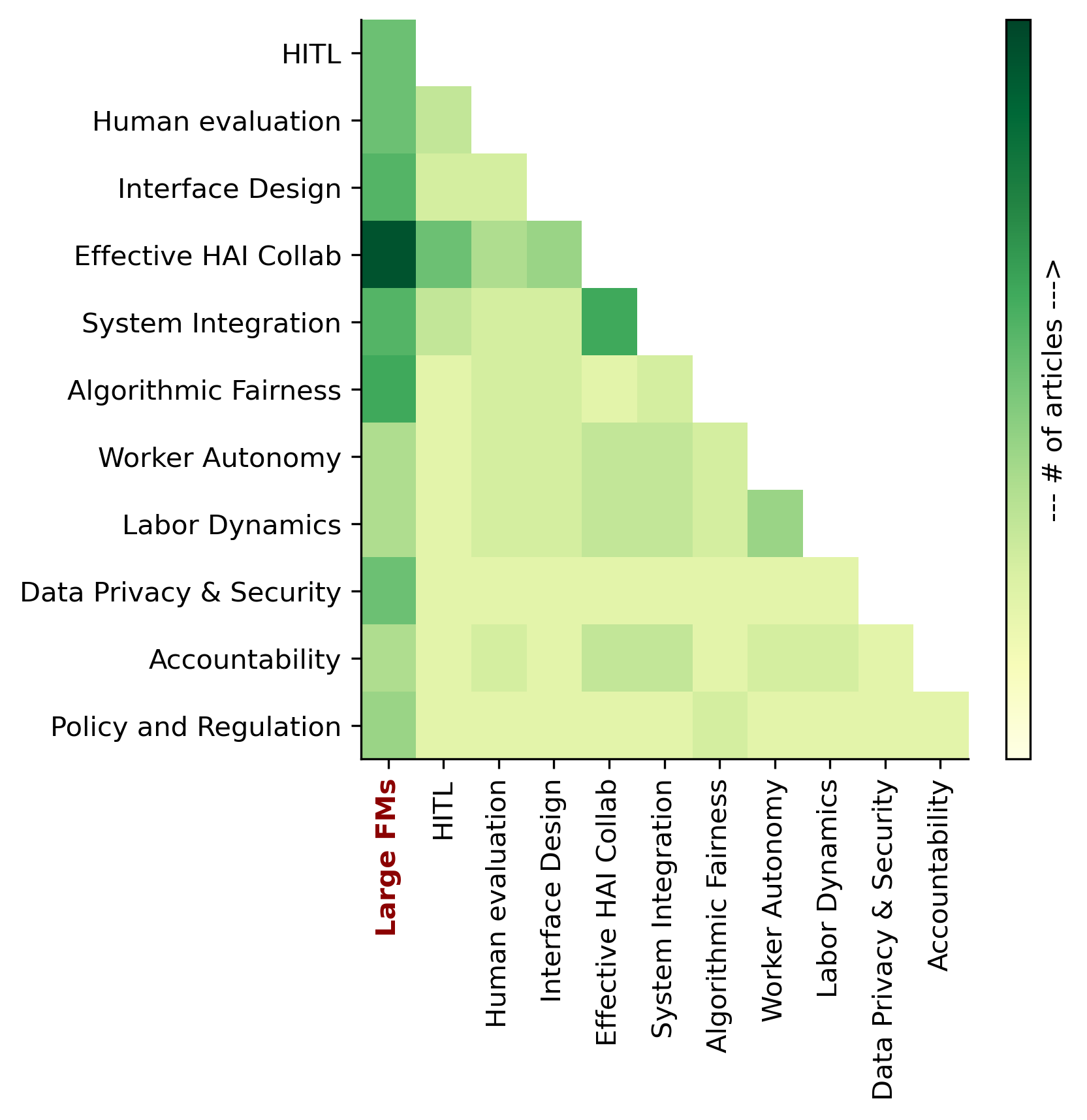}
        \caption{}
        \label{fig:heatmap}
    \end{subfigure}
    \caption{A representation of the scope of this survey. (a) We cover four broad categories - Human Guided AI Development, Design Principles for HAI Collaboration, Ethics and Governance, and their Applications - in the human-AI domain, along with the recent developments in them with the help of LFMs. The overlap between the four petals illustrates the fact that the four categories are not mutually exclusive; the articles may belong to two or more categories at the same time. (b) The heatmap shows a general trend of the number of articles co-existing within subcategories, including the articles discussing large foundation models within each subject area. The overlap between the applications is not analyzed, given their specific and varied nature. \emph{HAI}: Human-AI, \emph{HITL}: Human-in-the-Loop.}
    \label{fig:scope_heatmap}
    \Description[The survey scope is shown through four overlapping categories and a heatmap of article overlap.]{Panel (a) places Human-AI and Large Foundation Models at the center of four overlapping areas: Human-Guided AI Development at the top, Ethics and Governance on the left, Applications on the right, and Design Principles for HAI Collaboration at the bottom. Their overlap indicates that papers may address multiple areas. Panel (b) is a heatmap whose rows and columns represent survey subcategories, including human-in-the-loop learning, evaluation, interface design, collaboration, system integration, fairness, autonomy, labor, privacy, accountability, and regulation. Darker cells indicate more articles shared by a pair of subcategories, while the first column highlights articles involving large foundation models.
}
\end{figure}

\subsection{Scope of the Survey}
Our survey focuses on the articles describing the developments of human-AI collaboration through the years and how the introduction of large models is reforming the field. The search for articles was conducted on Google Scholar focusing on the years between 2014-2025, using keywords such as "Human-AI", and "Human-AI collaboration". For research focusing on the collaboration between humans and AI involving large models, additional keywords, including "large models", "foundation models", "large language models", and "large vision models" were utilized.  Because the field has rapidly shifted toward agentic and multi-agent AI systems, we also include a targeted update of selected 2025-early 2026 works on tool-use agents, multi-agent orchestration, computer-use agents, and human supervision of long-horizon AI workflows. The content was then distributed among the authors to filter based on the theme of our review to be structured into human-guided AI development, design principles for HAI collaboration, ethics and governance, and finally, applications. Some of the selected papers fall into two or more categories, as illustrated in Fig. \ref{fig:scope_heatmap}. Since human-AI collaboration and large models have only recently gained a lot of interest from the research community given the recent popularity of large foundation models (Fig. \ref{fig:HAI+LM}), some portion of studies cited in this survey could also be from arXiv preprints. We then sample the articles according to the structure desired for this survey, i.e., primarily covering Human-AI collaboration and its large foundation model counterparts in better model training, effective human-AI joint systems, safe and secure HAI collaboration, and their applications (Fig. \ref{fig:scope}). Although not systematic, this review captures the most visible work across AI, HAI, and collaboration.

In order to show support for the theme of our survey article, we try to implement HAI collaboration with the help of large language models to prepare this review. For the \emph{human} part, the authors survey the existing literature about human-AI collaboration, collect and filter a subset of articles according to the desired structure, organize the survey into relevant sections, study submission statistics, put together tables and visualizations, and finally, prepare the first complete draft of the manuscript. Using the \emph{AI} part in human-AI collaboration, we ask ChatGPT \cite{OpenAI2023ChatGPT} for its feedback regarding the language flow of each subsection, placement of the citations, and revising the language wherever necessary. This helps to ease in identifying the missing elements and for the authors to make necessary revisions. The final manuscript is, therefore, a product of human and AI collaboration.

\begin{figure}[t]
\centering
\includegraphics[width=0.9\linewidth]{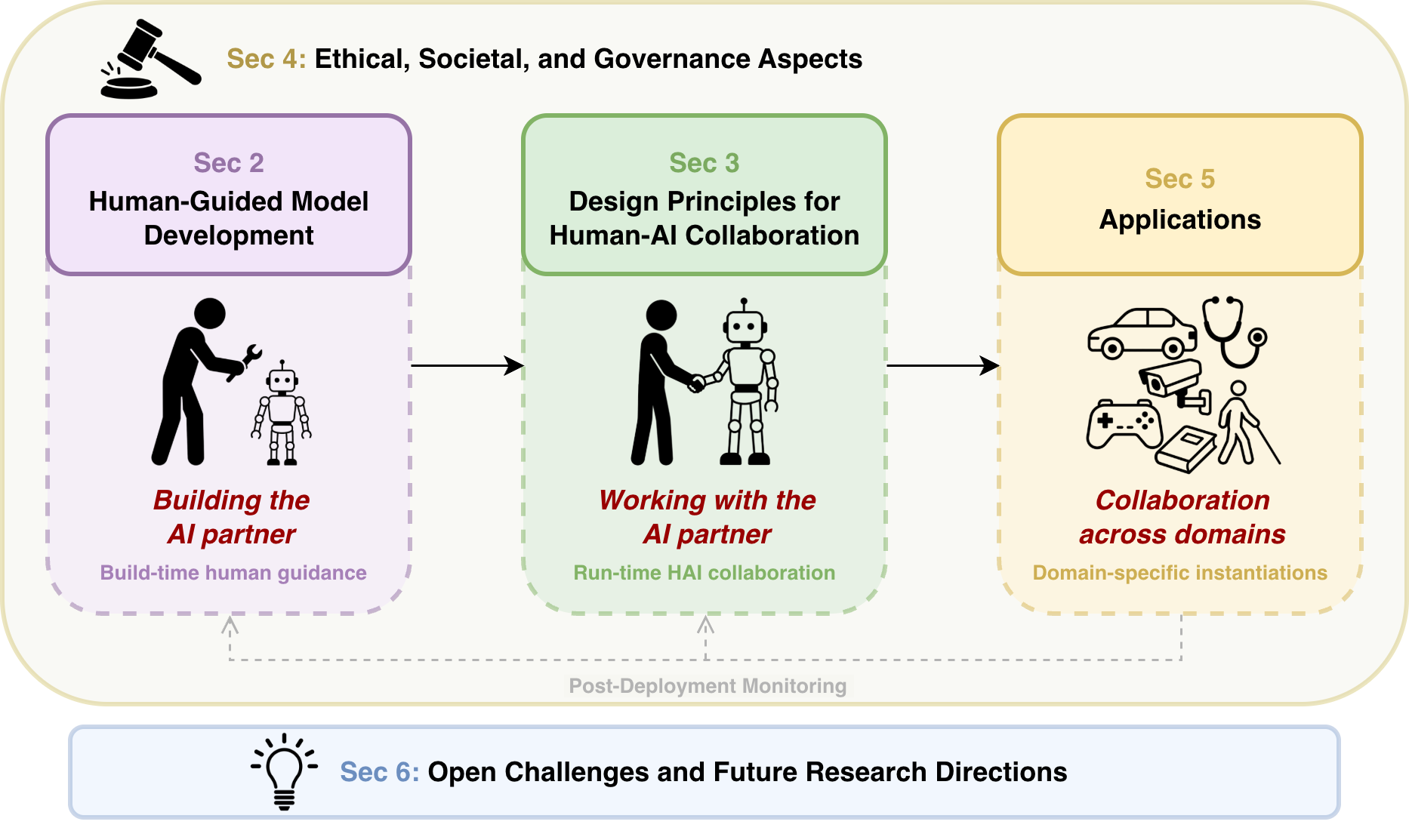}
\caption{This figure summarizes the lifecycle of Human–AI collaboration with LFMs. Section 2 covers build-time human guidance of models; Section 3 organizes run-time collaboration into recurring patterns such as AI augmentation, mixed initiative, control handoff, and emerging supervisory orchestration; Section 5 shows how these patterns appear across application domains. Section 4 acts as a cross-cutting governance layer, while Section 6 identifies open challenges that arise across the lifecycle.}
\label{fig:lifecycle}
\Description[The Human-AI collaboration lifecycle connects model development, run-time collaboration, applications, governance, and open challenges.]{The central flow moves from Section 2, Human-Guided Model Development, to Section 3, Design Principles for Human-AI Collaboration, and then to Section 5, Applications. These stages respectively represent building the AI partner, working with the AI partner, and applying collaboration across domains. Section 4 forms an enclosing governance layer around all three stages. A dashed post-deployment monitoring path feeds application experience back into model development and collaboration design. Section 6 appears below the lifecycle and represents open challenges and future research directions spanning all stages.
}
\end{figure}

%

\begin{table}
\centering
\small
\setlength{\tabcolsep}{4pt}
\renewcommand{\arraystretch}{1.2}
\caption{A tabular representation of the four broader focus topics covered in this survey, with their relevant subtopics. Each category contains its cited articles for the ease of reader reference.}
\begin{tabular}{|>{\centering\arraybackslash}m{0.180\linewidth}|@{}l@{}|}
\hline
\textbf{Human-AI Topics} &
\begin{tabular}{@{\hspace{4pt}}>{\raggedright\arraybackslash}m{0.238\linewidth}@{\hspace{4pt}}|@{\hspace{4pt}}>{\raggedright\arraybackslash}m{0.492\linewidth}@{\hspace{4pt}}}
\centering\arraybackslash\textbf{Subtopics} & \centering\arraybackslash\textbf{Articles Cited}
\end{tabular} \\ \hline
Section \ref{section:betterModel}: Human-Guided Model Development &
\begin{tabular}{@{\hspace{4pt}}>{\raggedright\arraybackslash}m{0.238\linewidth}@{\hspace{4pt}}|@{\hspace{4pt}}>{\raggedright\arraybackslash}m{0.492\linewidth}@{\hspace{4pt}}}
HITL and Active-Learning & \cite{Whittlestone2019}, \cite{hu2023}, \cite{Memmert2023}, \cite{Settles2009}, \cite{Pries2023}, \cite{Lu2022}, \cite{Bao2018}, \cite{Lertvittayakumjorn2020}, \cite{Yao2023} \\ \hline
Human-Guided Objective Shaping & \cite{christiano2017}, \cite{ouyang2022}, \cite{stiennon2020}, \cite{rafailov2023}, \cite{bai2022}, \cite{lee2023rlaif}, \cite{kaelbling1996reinforcement}, \cite{nakano2021}, \cite{align-anything-200k}, \cite{vodrahalli2025canonical}, \cite{hu2023}, \cite{Mirchandani2023}, \cite{Wei2022}, \cite{Zha2023}, \cite{zhang2021}, \cite{johnson2019no}, \cite{McNeese2021}, \cite{Zhang2025MMRLHF} \\ \hline
Human Evaluation and Alignment Metrics & \cite{bansal2019}, \cite{Jianlong2019}, \cite{align-anything-200k}, \cite{vodrahalli2025canonical}, \cite{clark2021}, \cite{Chhibber2022}, \cite{hu2023}, \cite{Tabrez2020}, \cite{lu2023}, \cite{uchendu2023}, \cite{liu2025survey}
\end{tabular} \\ \hline
Section \ref{section:hai_systems}: Design Principles for Human-AI Collaboration &
\begin{tabular}{@{\hspace{4pt}}>{\raggedright\arraybackslash}m{0.238\linewidth}@{\hspace{4pt}}|@{\hspace{4pt}}>{\raggedright\arraybackslash}m{0.492\linewidth}@{\hspace{4pt}}}
Interaction and Interface Design & \cite{zhang2021}, \cite{Hauptman2023}, \cite{Bryan2023}, \cite{Qing2023}, \cite{Mingming2023}, \cite{Tongshuang2022}, \cite{Kozlowski2006}, \cite{Salas2017}, \cite{Zhang2023}, \cite{Chhibber2022}, \cite{Clark2016}, \cite{wang2019human}, \cite{Bosch2019}, \cite{Flathmann2021}, \cite{wei2023}, \cite{Lemaignan2017} \\ \hline
Collaboration Patterns and Role Allocation & \cite{wei2023}, \cite{Tongshuang2022}, \cite{Pflanzer2022}, \cite{dubey2020}, \cite{liu2023tag}, \cite{ngo2023tag}, \cite{Zhang2023}, \cite{Gopinath2022}, \cite{Munyaka2023}, \cite{rafailov2023}, \cite{nakano2021}, \cite{hu2023}, \cite{ouyang2022}, \cite{ziegler2019}, \cite{Jianlong2019}, \cite{henry2022}, \cite{Lv2021}, \cite{wang2023}, \cite{Veselovsky2023}, \cite{pal2023}, \cite{Memmert2023}, \cite{ahmad2023towards}, \cite{Fuchs2023}, \cite{bansal2021does}, \cite{Chakraborti2017}, \cite{Bosch2019}, \cite{liu2023humans}, \cite{Neerincx2018}, \cite{Nikolaidis2017}, \cite{Zhao2022}, \cite{Kaptein2016}, \cite{Rastogi2023}, \cite{Sharma2023}, \cite{Chakrabarty2023}, \cite{Sharifi2022}, \cite{fourney2024magenticone}, \cite{motwani2024malt}, \cite{yu2025table}, \cite{vats2026guideline}, \cite{schoembs2026conversation}, \cite{masters2025orchestrating}, \cite{yao2024tau}, \cite{xie2024osworld}, \cite{abhyankar2025osworld}, \cite{xu2026theagentcompany}, \cite{jiang2025medagentbench} \\ \hline
System Integration and Evaluation & \cite{Weisz2021}, \cite{Chhibber2022}, \cite{Bau2019}, \cite{wang2023}, \cite{Eloundou2023}, \cite{harrer2023}, \cite{McNeese2021}, \cite{lyons2019}, \cite{Hou2023}, \cite{bansal2021does}, \cite{Tongshuang2022}, \cite{Veselovsky2023}, \cite{henry2022}, \cite{liu2023humans}, \cite{Munyaka2023}, \cite{Memmert2023}, \cite{chang2023}
\end{tabular} \\ \hline
Section \ref{section:safe}: Ethical, Societal and Governance Aspects &
\begin{tabular}{@{\hspace{4pt}}>{\raggedright\arraybackslash}m{0.238\linewidth}@{\hspace{4pt}}|@{\hspace{4pt}}>{\raggedright\arraybackslash}m{0.492\linewidth}@{\hspace{4pt}}}
Fairness in Collaborative Decision-Making & \cite{Ben2019}, \cite{Stahl2021}, \cite{Nicholas2023}, \cite{Chhibber2022}, \cite{Morrison2023}, \cite{kirk2021}, \cite{huang2023bias}, \cite{kotek2023}, \cite{meyer2023chatgpt}, \cite{li2023bias}, \cite{gallegos2023bias}, \cite{ohi2024bias}, \cite{bi2023bias}, \cite{ling2024applicants} \\ \hline
Empowering Human Autonomy and Well-Being & \cite{Chhibber2022}, \cite{Konstantis2023}, \cite{pal2023}, \cite{despotovic2024}, \cite{wang2023}, \cite{woodruff2023} \\ \hline
Collaborative Labor Dynamics & \cite{Hemmer2023}, \cite{CHOWDHURY202231}, \cite{Chhibber2022}, \cite{pal2023}, \cite{Eloundou2023}, \cite{Purdy2016}, \cite{despotovic2024}, \cite{wang2023}, \cite{walkowiak2023} \\ \hline
Data privacy and Security & \cite{Ezer2019}, \cite{Yin2019}, \cite{Kaissis2020}, \cite{Admin2022}, \cite{Lepri2021}, \cite{Hacker2023}, \cite{ullah2023}, \cite{gupta2023}, \cite{de2023}, \cite{thapa2023}, \cite{sebastian2023} \\ \hline
Building Trust and Shared Accountability & \cite{Hou2023}, \cite{Pflanzer2022}, \cite{caldwell2022agile}, \cite{choudhury2024large}, \cite{kim2023help}, \cite{bansal2019} \\ \hline
Policy and Regulation & \cite{jobin2019}, \cite{cath2018}, \cite{Stahl2021}, \cite{Bender2021}, \cite{Chinonso2023}, \cite{mit2023law}, \cite{gdpr2023law}
\end{tabular} \\ \hline
Section \ref{section:applications}: Applications &
\begin{tabular}{@{\hspace{4pt}}>{\raggedright\arraybackslash}m{0.238\linewidth}@{\hspace{4pt}}|@{\hspace{4pt}}>{\raggedright\arraybackslash}m{0.492\linewidth}@{\hspace{4pt}}}
Healthcare & \cite{henry2022}, \cite{Bienefeld2023}, \cite{Memmert2022}, \cite{Carrie2019}, \cite{McKinney2020}, \cite{BUDD2021}, \cite{CHOUDHURY2022}, \cite{lyu2023}, \cite{kung2023}, \cite{johnson2023}, \cite{strong2024deferral}, \cite{biswas2024clinicaldoc} \\ \hline
Autonomous Vehicles & \cite{Atakishiyev2021}, \cite{Lv2021}, \cite{Jianlong2019}, \cite{Fuchs2023}, \cite{yang2024}, \cite{park2024}, \cite{cui2024drive}, \cite{tian2024critical}, \cite{wang2023empowering}, \cite{wen2023}, \cite{wang2023drive} \\ \hline
Surveillance and Security & \cite{killcrece_2003}, \cite{Pazho2023}, \cite{Hauptman2023}, \cite{Guo2018}, \cite{chen2023}, \cite{jain2023}, \cite{baruwalchhetri2024alertfatigue}, \cite{oliver2024carbonfilter} \\ \hline
Games & \cite{MarioKart8Deluxe2017}, \cite{frans2021}, \cite{xu2023werewolf}, \cite{sobieszek2022}, \cite{akata2023}, \cite{sidji2024codenames}, \cite{todd2023}, \cite{vartinen2022}, \cite{hu2023}, \cite{white2024communicate} \\ \hline
Education & \cite{nwana1990}, \cite{vanlehn2011}, \cite{whitaker2013}, \cite{kasinathan2017}, \cite{eicher2018}, \cite{hartle2019}, \cite{dikli2006}, \cite{extance2023}, \cite{milano2023}, \cite{rose2023a}, \cite{hellas2023}, \cite{chang2023}, \cite{kong2025synergy}, \cite{schotter2025spiral} \\ \hline
Accessibility & \cite{kumar2022}, \cite{Ozarkar2020}, \cite{Khan2020}, \cite{Wen2021}, \cite{Ghazal2021}, \cite{taheri2023}, \cite{gadiraju2023}, \cite{brilli2024airis}, \cite{tokmurziyev2025llmglasses}
\end{tabular} \\ \hline
\end{tabular}
\label{tab:topics}
\end{table}

\subsection{Outline of the Survey}
The survey follows the lifecycle summarized in Fig. \ref{fig:lifecycle}, moving from human-guided model development to run-time collaboration, governance, applications, and open challenges. In each segment of our study, we commence by delineating traditional methodologies employed in human-AI collaboration, subsequently delving into the contributions of extensive pre-trained foundation models in the fulfillment of these tasks (Table \ref{tab:topics}). Section \ref{section:betterModel} explores the involvement of incorporating human expertise into the AI model training cycle, fostering a cooperative relationship between humans and AI in active learning scenarios, enhancing learning through human feedback, and involving human experts in the thorough evaluation of machine learning models \cite{hu2023, Settles2009, ziegler2019}. Section \ref{section:hai_systems} shifts to run-time collaboration, where the trained model becomes a partner whose interface, role, initiative, and control boundaries must be designed \cite{Lemaignan2017, Mingming2023}. 

A major concern in Human-AI collaboration is the safety, security, and trustworthiness of AI systems. Section \ref{section:safe} of our comprehensive analysis covers multiple dimensions: mitigating algorithmic biases to uphold fairness, assessing the impact of Human-AI collaboration on workers' autonomy, well-being, and job satisfaction, examining economic repercussions on employment and wages, addressing data privacy and security concerns, building trust in AI systems, and navigating the legal and regulatory frameworks governing Human-AI interactions \cite{Konstantis2023, Hemmer2023, Braun2021}. These factors are integral to creating an ethical and responsible environment for AI's integration into human-centric workflows. Section \ref{section:applications} explores the broad spectrum of Human-AI collaboration applications across various sectors. We investigate the unique challenges and opportunities in fields such as healthcare \cite{Bienefeld2023}, autonomous vehicles, surveillance systems \cite{Pazho2023, Guo2018}, gaming \cite{frans2021}, education, and accessibility. Understanding the specific characteristics and benefits of Human-AI collaboration in these areas is key to influencing its future direction and maximizing its societal impact. Finally, Section \ref{sec:future} summarizes the core findings and identifies key open challenges and future research directions, providing a roadmap for building the next generation of effective, fair, and trustworthy human-AI partnerships.

\section{Human-Guided AI Model Development: \textit{Building the AI Partner}} 
\label{section:betterModel}

 \begin{figure}[t]
\centering
\includegraphics[width=0.65\linewidth]{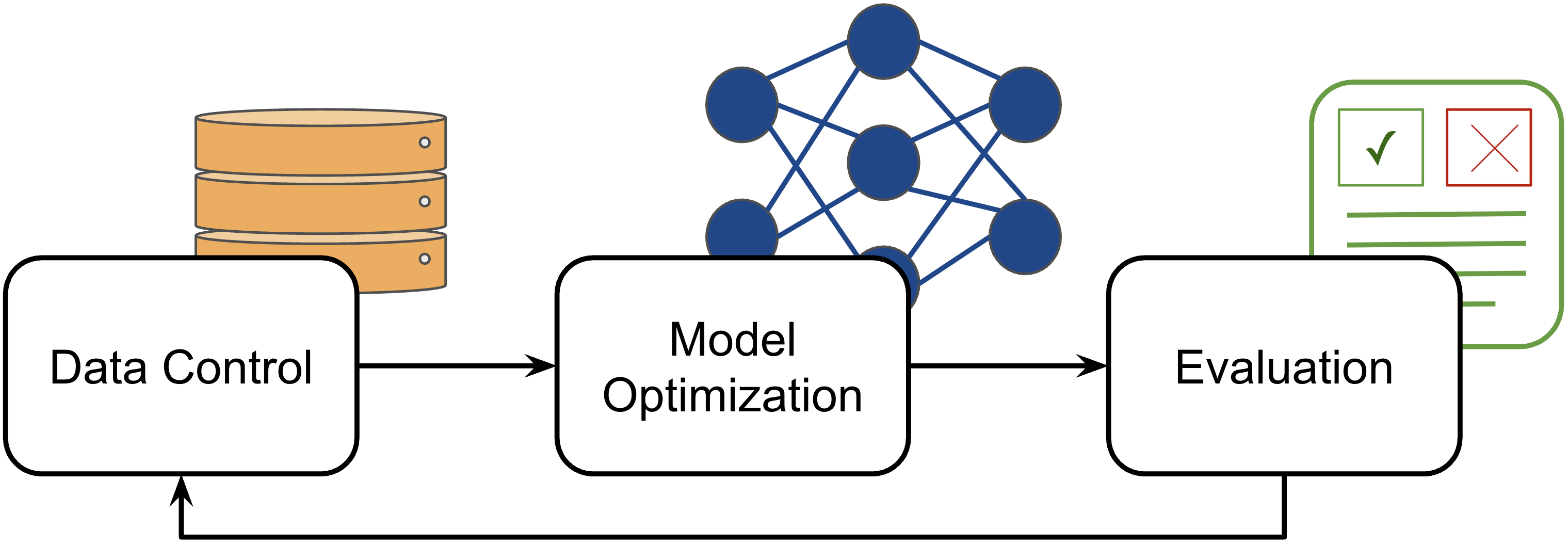}
\caption{The three-phase iterative framework for human-guided model development - comprising data control, model optimization, and evaluation, as described by Maadi et al. \cite{Maadi2021}.}
\label{fig:datapipeline}
\Description[An iterative framework links data control, model optimization, and evaluation.]{Three stages of Data Control, Model Optimization, and Evaluation are arranged from left to right. Forward arrows show data being used to optimize a model and the resulting model being evaluated. A return arrow from Evaluation to Data Control closes the loop, indicating that evaluation results inform the next round of data selection or correction.}
\end{figure}

This section examines build-time collaboration, where humans shape LFMs through data, objectives, and evaluation before the model is deployed as a collaborative partner. We adapt the three-phase view of Maadi et al. \cite{Maadi2021}, \textit{data control $\rightarrow$ model optimization $\rightarrow$ evaluation} (Fig. \ref{fig:datapipeline}), to review the points where human insight most clearly improves an FM:
\begin{itemize}
    \item Sec. \ref{HITL}, Human-in-the-Loop Data Curation \& Active Learning: People curate or label the right examples and, for LFMs, craft or filter instruction prompts that guide later fine-tuning.
    \item Sec. \ref{SecRLHF}, Human-Guided Objective Shaping: Preference-based methods such as RLHF and DPO let human feedback reshape the model’s reward or loss surface during optimization.
    \item Sec. \ref{HumanEval}, Human Evaluation and Alignment Metrics: Domain experts and end-users judge outputs, supplying both task scores and subjective signals of trust, fairness, and usability, which feed into the next iteration of the HAI collaboration pipeline.
\end{itemize}

\subsection{Human-in-the-Loop Data and Active-Learning} 
\label{HITL}

Human-in-the-Loop (HITL) methods exemplify human-AI collaboration by combining human judgment with automated efficiency. Humans guide model design by bringing ethical insight beyond technical metrics (Whittlestone et al. \cite{Whittlestone2019}), while AI accelerates data processing and suggests patterns that might not be easy to catch for the time-constrained annotators.  During data construction and fine-tuning, information flows bidirectionally. For example, NLP-based directives like InstructRL, as proposed by Hu and Sadigh \cite{hu2023}, steer AI toward more interpretable, aligned behaviors, and AI-generated prompts enrich human brainstorming, sparking novel ideas \cite{Memmert2023}. The diverse roles of humans, ranging from providing ethical oversight in the design phase to contributing to quality control and data labeling during model training and execution, are crucial for optimizing AI performance.

\subsubsection{\textbf{Human-AI Collaboration with Active Learning}}
\begin{figure}[t]
\centering
\includegraphics[width=0.57\linewidth]{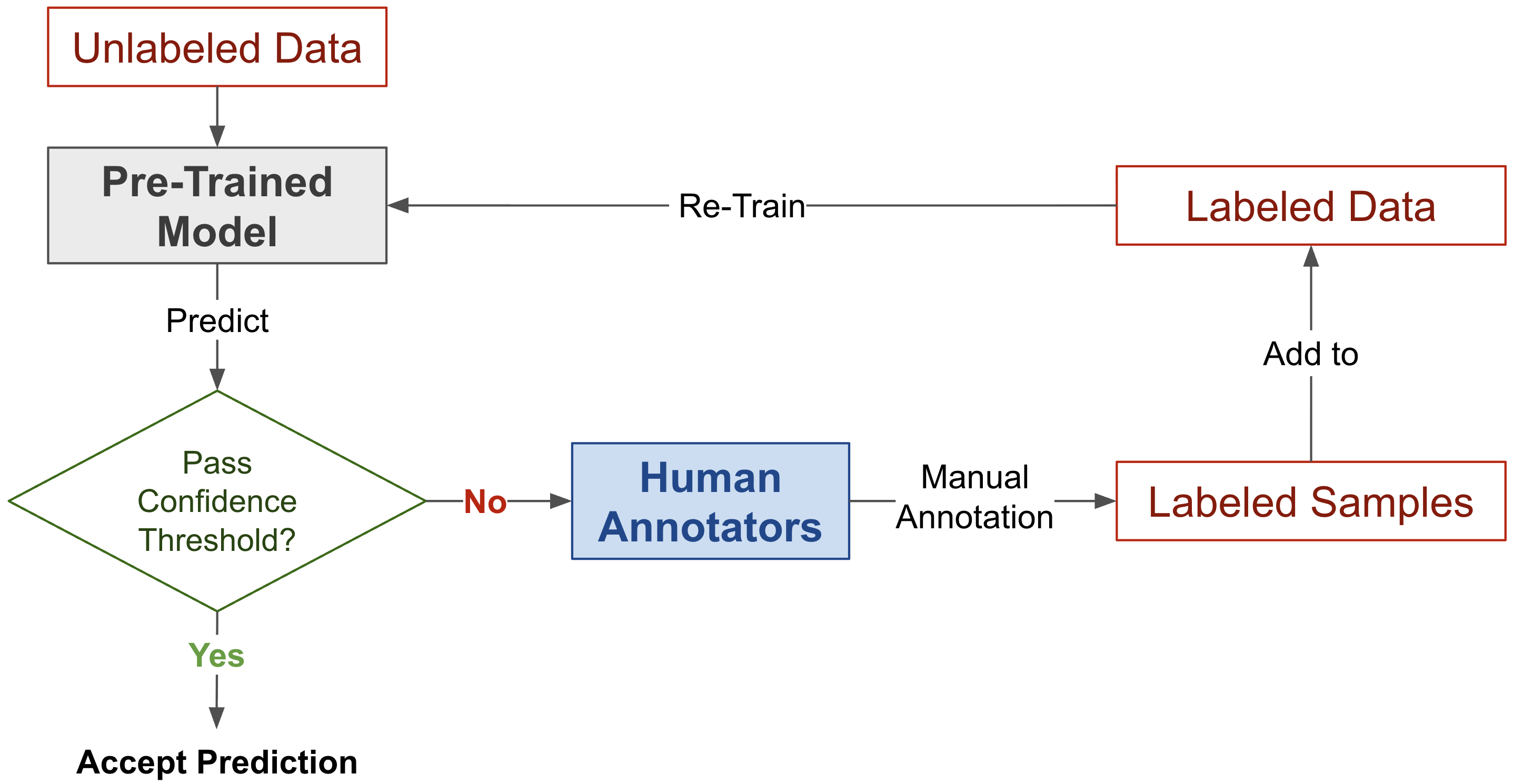}
\caption{A visual representation of an active learning workflow. If the model predictions do not pass the required confidence value thresholds required for a task, the data is sent to humans for manual annotations, further improving the model by retraining it with the now-labeled data.}
\label{fig:activelearning}
\Description[An active-learning workflow routes uncertain predictions to human annotators and uses their labels to retrain the model.]{Unlabeled data enters a pre-trained model, which produces a prediction that is checked against a confidence threshold. If the prediction passes, it is accepted. If it does not pass, the sample is sent to human annotators for manual annotation. The resulting labeled samples are added to the labeled dataset, which is then used to retrain the pre-trained model, forming an iterative improvement loop.}
\end{figure}

Unlike traditional machine learning concepts, active learning involves an iterative process where a model selectively identifies data requiring labeling, optimizing system performance with minimal training data. Particularly, this approach is crucial for fine-tuning large pre-trained foundation models that require substantial labeled data for specific user scenarios. Active learning efficiently utilizes human expertise to pinpoint areas of uncertainty, enabling more targeted training with less annotation effort \cite{Settles2009}. The active learning workflow begins with an unlabeled dataset and a pre-trained model. The model predicts labels for each sample, outputting confidence levels (Fig. \ref{fig:activelearning}). When predictions fall below a quality threshold, human annotators step in for manual annotation. This iterative process of re-training the model with new labeled data continues until satisfactory confidence levels are achieved. Humans and AI thus collaborate in efficient data labeling and continuous model improvement.

Recent research in this area has introduced novel methods to leverage human expertise effectively. For instance, Pries et al. \cite{Pries2023} propose a method for efficient data labeling using precise distance measurements to filter data samples, streamlining the presentation of data to experts. Lu et al. \cite{Lu2022} discuss human-guided interventions for continuous model improvement, focusing on dissociating biases. Bao et al. \cite{Bao2018} explore the transformation of human-annotated rationales into continuous attention mechanisms, enhancing the learning of domain-invariant representations. Additionally, Lertvittayakumjorn et al. \cite{Lertvittayakumjorn2020} present a generalizable approach for debugging deep text classification models with human input, applicable to larger models. Finally, Yao et al. \cite{Yao2023} propose an explainable-generation active learning framework for simpler tasks, potentially improving the sampling process in data augmentation. 
\\

\noindent Taken together, research in this area shows a shift in the human role in data curation. Traditional active learning balances annotation cost against model performance, but with Large Foundation Models, humans act less as mass annotators and more as targeted teachers during fine-tuning. This shift brings a difficult trade-off: small, carefully chosen datasets must be diverse enough to reduce, not reinforce, the biases already present in the model. A key open challenge remains to design bias-aware active learning methods that identify not only areas of model uncertainty but also regions where pre-trained knowledge is skewed or incomplete.


\subsection{Human-Guided Objective Shaping}
\label{SecRLHF}

After curating the \emph{data} that an LFM sees, collaborators can intervene a second time by reshaping the \emph{objective} it tries to optimize. This extends human-guided model development beyond labeling examples: humans express preferences, rankings, or critiques over model outputs, indicating which behaviors should be preferred when multiple responses are plausible. These signals are then converted into a learnable reward or direct policy update. The canonical pipeline, Reinforcement Learning from Human Feedback (RLHF) \cite{christiano2017,ouyang2022,stiennon2020}, trains a reward model on pairwise preference judgements and then fine-tunes the policy to maximize that reward, producing assistants that are measurably more helpful and safer. Simpler, gradient-based alternatives such as Direct Preference Optimization (DPO) \cite{rafailov2023} and Constitutional AI \cite{bai2022} bypass reinforcement yet achieve similar alignment by treating human or rule-based critiques as a differentiable signal. This section surveys these preference-based algorithms, the types of human feedback they require, and open challenges in keeping objectives both scalable and faithful to human intent.

\subsubsection{\textbf{RLHF Pipeline.}} Typically, the RLHF approach involves three stages \cite{ouyang2022}: initially, the pre-trained foundation model generates candidate responses; human annotators then evaluate these outputs, providing explicit preference-based feedback; subsequently, a reward model is trained on these human-generated evaluations. Finally, policy optimization methods, often proximal policy optimization (PPO) or direct preference optimization (DPO) \cite{rafailov2023}, fine-tune the foundation model to maximize the learned reward. This iterative human–AI feedback loop enables continuous and targeted improvements in model alignment.

\subsubsection{\textbf{Collaborative Advantages.}} RLHF’s reliance on human evaluation significantly enhances the alignment of models with ethical and functional expectations. Human annotators can actively identify and mitigate harmful biases or unsafe outputs early in the optimization process, producing fairer and more trustworthy AI systems \cite{lee2023rlaif}. Additionally, the iterative nature of human feedback facilitates rapid adaptation to evolving domain-specific or user-specific preferences, overcoming the rigidity inherent in traditional supervised learning methods. Transparency is also notably improved, as human-generated feedback and reward insights provide explicit visibility into the rationale behind model decision-making \cite{kaelbling1996reinforcement}. The adaptability of RLHF frameworks to user and domain-specific contexts further enhances their practical effectiveness, enabling personalized and contextually significant AI behaviors.

Despite these advantages, implementing RLHF involves several practical challenges. Ensuring consistency and accuracy of human feedback, scaling RLHF effectively to large datasets, and aligning human insights with algorithmic processes are key considerations that must be addressed to fully leverage RLHF's potential \cite{lee2023rlaif}.

\subsubsection{\textbf{Extensions of Preference-Based Alignment}}
Recent methodological advances have expanded preference-based alignment. OpenAI’s InstructGPT \cite{ouyang2022} introduced a structured three-phase training process comprising supervised policy training, reward model training with human feedback, and subsequent reinforcement learning-based policy optimization. Similarly, Direct Preference Optimization (DPO) simplifies the training by directly mapping human feedback into policy improvements, thereby reducing complexity and enhancing the quality of outputs \cite{rafailov2023}. A recent survey of DPO further shows that preference optimization is broadening beyond the basic RLHF-free objective, with variants organized around data strategy, learning frameworks, constraint mechanisms, and model properties \cite{liu2025survey}. Further illustrating RLHF's potential, Nakano et al. \cite{nakano2021} introduced web browser-assisted agents trained via RLHF, demonstrating significant improvements in real-world usability. Recent studies have further expanded RLHF frameworks through innovative approaches. Ji et al. \cite{align-anything-200k} introduced a dataset enabling models to integrate multimodal human feedback, spanning text, images, audio, and video, thereby significantly enriching the interaction between humans and AI. Additionally, Vodrahalli et al. \cite{vodrahalli2025canonical} proposed a canonical basis of human preferences, identifying a compact yet expressive set of fundamental categories that efficiently represent diverse human judgments, potentially reducing annotation overhead and enhancing model interpretability.

Extending beyond language models, RLHF methodologies have influenced fields like robotics and interactive decision-making systems. Hu and Sadigh’s InstructRL \cite{hu2023} employs natural-language instructions to effectively guide robotic actions, underscoring the versatility of RLHF approaches. Additionally, research by Mirchandani et al. \cite{Mirchandani2023} highlights the utility of large language models (LLMs) in robotics, leveraging their in-context learning abilities to facilitate complex sequential tasks. Chain-of-thought prompting techniques \cite{Wei2022} and correction-based learning frameworks such as DROC \cite{Zha2023} further exemplify RLHF’s expanding scope.

Nevertheless, RLHF models still face some hurdles, notably in bridging communication barriers between humans and AI. Studies by Zhang et al. \cite{zhang2021} and Johnson and Vera \cite{johnson2019no} emphasize these challenges, highlighting the need for more sophisticated human-AI communication frameworks. McNeese et al. \cite{McNeese2021} further underscore the limitations of current RLHF models in managing real-world complexities, particularly regarding nuanced human communication and the ability to generalize from simulated environments to real-world settings.
\\

\noindent To summarize, the work on objective shaping marks a shift from programming explicit goals to learning implicit human values through feedback. RLHF and its variants demonstrate that human preference signals can reliably steer models toward safer and more useful behaviors. However, the practical limits of noisy, costly, and potentially narrow annotation remain. Recent advances \cite{vodrahalli2025canonical, Zhang2025MMRLHF} point toward concrete solutions: datasets that capture multimodal human feedback (text, images, audio, video) and frameworks that distill a canonical basis of preferences reduce annotation overhead while broadening coverage of diverse values. Together, these directions suggest that scalable, bias-resilient objective shaping is achievable, not by collapsing human preferences into a single worldview, but by structuring them into compact, expressive, and extensible feedback spaces.

\begin{figure}[t]
\centering
\includegraphics[width=0.5\linewidth]{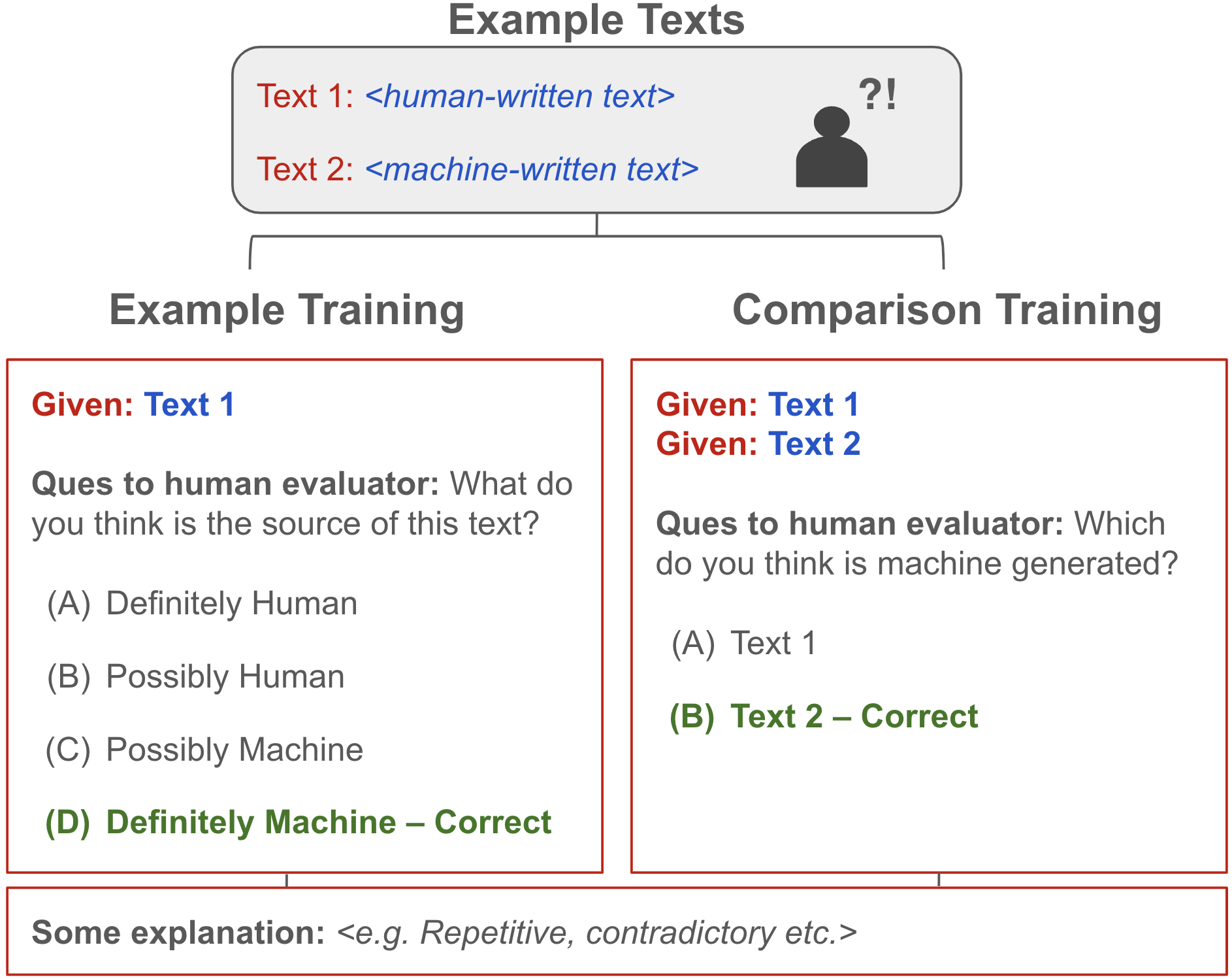}
\caption{Evaluation training strategies based on Clark et al. \cite{clark2021}. They employ training methods by giving example and comparison tests, along with their correct answers. This is followed by an explanation of why the answer is correct.  This way, the evaluators gradually learn about how to distinguish between human and machine-generated texts.}
\label{fig:clark30}
\Description[Two evaluator-training tasks teach people to distinguish human-written and machine-written text.]{The figure begins with two example texts, one labeled as human-written and the other as machine-written. In Example Training, an evaluator sees one text and selects among four source judgments ranging from definitely human to definitely machine; the correct answer is shown as definitely machine. In Comparison Training, the evaluator sees both texts and identifies which one is machine-generated; Text 2 is marked correct. Both tasks are followed by an explanation of cues such as repetitive or contradictory writing.
}
\end{figure}

\subsection{Human Evaluation and Alignment Metrics}
\label{HumanEval}
Human evaluation is the final collaborative safeguard that ensures foundation models satisfy both task requirements and human expectations.  Rather than serving as a single `final report card' human evaluation forms a feedback loop: people surface errors, voice preferences, and refine rubrics, which in turn steer data curation (\ref{HITL}) and objective shaping (\ref{SecRLHF}).  In this sense, evaluation becomes a form of post-hoc guidance rather than passive measurement. The result is an iterative alignment cycle that balances raw performance with usability, trust, and safety.

\subsubsection{\textbf{Trust, Transparency, and Safety}}
Establishing calibrated trust begins with making model limits visible.  Bansal et al.\cite{bansal2019} show that users build more accurate mental models when an AI system discloses its own error likelihood.  Zhou and Chen’s Uncertainty‑Performance Interface (UPI) \cite{Jianlong2019} couples confidence scores with outcome heat‑maps so that decision makers can dynamically adjust their thresholds.  Recent work extends these ideas to multimodal settings: Ji et al. \cite{align-anything-200k} let raters score image–text consistency, while Vodrahalli et al.\cite{vodrahalli2025canonical} distill 21 canonical preference axes that cover most harm‑related judgements, trimming annotation cost without losing coverage. Together, these studies frame transparency as an ongoing dialogue rather than a static disclosure.

\subsubsection{\textbf{Collaborative Approaches to Improve Evaluation}}
Human contribution is most powerful when it shapes evaluators, not just evaluations. Clark et al. \cite{clark2021} boost label accuracy by providing annotators with contrasting model outputs and rationale. They examine how human evaluators can differentiate between texts produced by humans and models like GPT-2 and GPT-3. They suggest methods to enhance evaluation accuracy, including detailed instructions, annotated examples, and text comparisons (Fig.~\ref{fig:clark30}).  Chhibber et al. \cite{Chhibber2022} demonstrate that teaching annotators the model’s decision heuristics increases both trust and delegation willingness.  InstructRL \cite{hu2023} and Tabrez et al. \cite{Tabrez2020} further show that natural‑language feedback can be folded back into the policy itself, reducing the evaluation–training gap.  Error‑analysis prompting (EAPrompt) with chain-of-thought reasoning \cite{lu2023} and collaborative detection workflows \cite{uchendu2023} route annotator attention to likely failure modes, lowering variance and cost.  
\\

\noindent Overall, the work on human evaluation shows a clear move away from static performance metrics toward an ongoing, people-centered checking of model behavior. As LFM outputs become more fluent and convincing, automated benchmarks alone cannot capture qualities like trust, safety, or transparency. This puts humans in a different position, not just as scorers, but as active partners who stress-test the system and surface the kinds of failures that models are best at hiding. The hard part is not only collecting those judgments at scale, but figuring out how to train and support evaluators so they can apply consistent rubrics, recognize subtle flaws, and probe these systems in ways that automated tests simply cannot. Together with data curation and objective shaping, human evaluation closes the build-time loop of human-guided model development: humans influence what the model sees, what it optimizes, and how its behavior is judged. This differs from the run-time collaboration patterns discussed in Sec. \ref{section:hai_systems}, where the model is already deployed and the central question becomes how humans and AI divide initiative, responsibility, and control during task execution.
\section{Design Principles for Human-AI Collaboration: \textit{Working with the AI Partner}}
\label{section:hai_systems}
We now turn to the question of how humans and AI systems collaborate after deployment. At this run-time stage, the model is no longer only an object to be trained or evaluated, but a partner whose interface, role, initiative, and control boundaries must be carefully designed.

Effective human-AI teams rely on more than raw model accuracy; they require systems intentionally designed for collaboration. This section lays out core design principles that turn powerful large pre-trained foundation models into reliable partners: (i) crafting interaction channels that surface the right information at the right moment, (ii) allocating initiative so humans and AI can fluently guide or assist one another, and (iii) embedding continuous evaluation and adaptation to keep the partnership usable, trustworthy, and resilient as tasks, data, and user expectations evolve.

\subsection{Interaction and Interface Design}
Before roles and control can be allocated, human-AI collaboration requires interaction channels through which people can understand, guide, correct, and trust the AI system. Interfaces, therefore, act as the practical surface where run-time collaboration becomes possible: they expose model behavior, support user input, and make the system's state and uncertainty legible.

Effective interactions and interfaces translate AI reasoning into human-understandable cues, enabling seamless and timely communication between humans and AI. To support human-AI collaboration effectively, it is crucial to focus on personalizing interactions, facilitating natural communication channels, and providing transparent explanations of AI behavior.

\subsubsection{\textbf{Personalization and Adaptive Displays}}
Integrating AI agents into collaborative teams has historically been challenging due to limited adaptability and personalization. Early AI systems, designed for narrowly defined tasks, struggled to grasp human subtleties or adjust dynamically to evolving team requirements, hindering effective collaboration \cite{zhang2021}. The advent of LLMs significantly enhanced these capabilities. Leveraging extensive training datasets and sophisticated algorithms, LLM-equipped AI agents now exhibit improved adaptability, dynamically adjusting autonomy levels to match team workflows and individual user preferences \cite{Hauptman2023}. Additionally, LLMs facilitate richer conversational interactions, providing context-sensitive, multi-turn responses and enabling AI to mediate complex interactions among human team members \cite{Bryan2023, Qing2023}. This enhanced personalization extends to mobile UI/UX evaluations, where LLMs analyze usability test videos to tailor user interactions effectively \cite{Mingming2023}. Techniques such as chaining LLM prompts further enhance transparency and user control, customizing collaboration experiences according to user needs \cite{Tongshuang2022}. This way, personalization is not only a usability feature, but also a mechanism for role calibration: the interface adapts what the AI shows, asks, or takes over based on the user's goals, expertise, and context.

\subsubsection{\textbf{Conversational and Multimodal Interaction}}
Effective human-AI collaboration relies heavily on clear and natural communication, influencing trust, information exchange, and team coordination \cite{Kozlowski2006, Salas2017}. Advances in AI conversational capabilities now enable responsive and contextually appropriate dialogue, bolstering user confidence and improving team effectiveness \cite{Zhang2023}. Notably, teachable conversational agents exemplify significant progress, learning dynamically from conversational interactions and adapting their knowledge to meet evolving task demands \cite{Chhibber2022}. These agents' perceived likability and human-likeness directly enhance their communication efficacy \cite{Clark2016}. Furthermore, natural language processing (NLP) continues to refine voice-based interactions, enhancing both verbal and textual exchanges within collaborative teams \cite{wang2019human}.

Efforts have also been devoted to establishing standardized communication frameworks within HAI collaboration teams. Initiatives like developing a shared vocabulary (Taxonomy Model) and comprehensive communication and explanation models facilitate clear, mutual understanding of AI actions and rationale, significantly boosting collaborative efficiency and team cohesion \cite{Bosch2019}. Conversational and multimodal interfaces, thus, provide a communication surface for mixed-initiative collaboration, where either the human or the AI can propose, clarify, revise, or redirect the task.

\subsubsection{\textbf{Transparency, Feedback and Explainability}}
The principles of user interface design (UI) and user experience design (UX) have progressively emphasized transparency, ethical interaction, and clear feedback mechanisms in collaborative human-AI systems. Early frameworks emphasized usability testing, clear communication, and synchronization between human and AI team members, laying foundations for trustworthy interactions \cite{Mingming2023, Flathmann2021}. Recent innovations leveraging large-scale pre-trained models have transformed these traditional UI/UX designs. For example, two-stage frameworks like ChatIE reframe zero-shot Information Extraction (IE) as interactive, multi-turn question-answering sessions, achieving superior performance and enabling transparent user interactions \cite{wei2023}. Empirical studies involving diverse user groups confirm these advances significantly enhance both interaction quality and emotional satisfaction, reinforcing AI's ability to adopt human-like behaviors and integrate seamlessly into complex social contexts \cite{Lemaignan2017}. Transparency and explainability form the control layer of human-AI interaction: they help users inspect the AI's reasoning, detect failure modes, and decide when to accept, revise, or override its output.
\\

\noindent This reveals a central tension in LFM-based collaboration. We are moving from rigid graphical interfaces to fluid, conversational interactions, which makes AI systems easier to approach but can also weaken precision, constraint enforcement, and user control. Natural language can hide system state and reasoning, while structured interfaces make parameters, validation, and recovery actions more explicit. The open challenge is therefore to design hybrid interfaces that combine natural-language flexibility with visible reasoning, structured controls, previews, confirmations, and undo mechanisms, so users retain clarity and agency as collaborative tasks become more complex.


\begin{figure}[t]
\centering
\includegraphics[width=\linewidth]{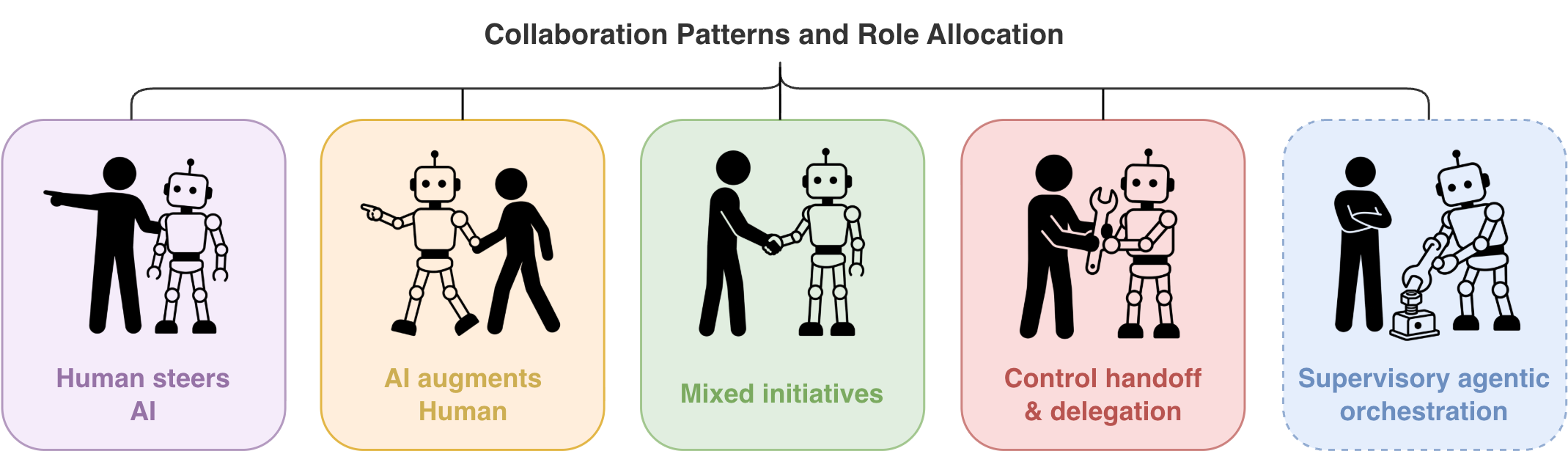}
\caption{Collaboration patterns and role allocation in HAI systems. Section 3.2 organizes run-time collaboration into human-steered AI, AI-augmented human work, mixed-initiative interaction, control handoff and delegation, and emerging supervisory orchestration of agentic systems. The dashed boundary highlights supervisory orchestration as a newer mode where humans monitor, guide, and intervene in multi-step AI workflows.}
\label{fig:design_patterns}
\Description[Five panels illustrate recurring Human-AI collaboration patterns.]{Five panels are arranged from left to right under the heading Collaboration Patterns and Role Allocation. Human Steers AI shows a person directing a robot. AI Augments Human shows a robot guiding or assisting a person. Mixed Initiatives shows a person and robot shaking hands. Control Handoff and Delegation shows them transferring a tool. Supervisory Agentic Orchestration shows a person monitoring a robot performing a task. The final panel has a dashed boundary to identify supervisory orchestration as an emerging collaboration pattern.}
\end{figure}

\subsection{Collaboration Patterns and Role Allocation}
The interaction channels described above become useful only when they are tied to clear collaboration patterns: who initiates action, who provides support, who makes final decisions, and when control shifts between partners. We organize this design space into recurring patterns that appear across LFM-based HAI systems: Human steers AI, AI Augments Human, Mixed-Initiative Collaboration, Dynamic Delegation/Control Handoff, and the emerging pattern of Supervisory Orchestration in agentic systems (Fig. \ref{fig:design_patterns}). They provide a practical vocabulary for connecting interface choices to role allocation, control, failure modes, and safeguards (Table \ref{tab:pattern_table}). Effective human-AI collaboration relies on clearly defining interaction patterns and thoughtfully allocating roles to maximize the strengths of both partners.

\subsubsection{\textbf{Human steers AI}}
This pattern captures cases where humans actively steer the AI during use, through prompting, in-context correction, and contextual oversight by adjusting its behavior and level of autonomy as the task unfolds, rather than only shaping it before deployment (Sec. \ref{section:betterModel}). Human expertise plays an essential role in enhancing AI capabilities throughout various development stages, from data preparation and algorithm refinement to ethical and contextual oversight \cite{wei2023, Tongshuang2022, Pflanzer2022}. Accurate data annotation and tailored algorithm adjustments guided by human feedback significantly improve AI precision and applicability \cite{dubey2020,liu2023tag,ngo2023tag}. Humans also ensure AI adherence to societal norms and ethical standards, aspects that purely algorithmic approaches might overlook \cite{Pflanzer2022}.

Further, intuitive interface designs facilitate effective human-AI communication, ensuring that sophisticated AI outputs remain accessible and practically useful \cite{Zhang2023}. Continuous human input and real-world behavioral data enhance specialized AI systems, such as adaptive driving systems, integrating subtle human behaviors into AI decision-making \cite{Gopinath2022, Munyaka2023}. At run time, this steering is expressed largely through natural-language instruction and preference signals, the deployment-side counterpart to the training-time preference methods covered in Sec. 2.2 (e.g., RLHF, DPO, InstructRL \cite{rafailov2023, hu2023}), letting users align AI outputs with their expectations without retraining the model.




\begin{table*}[t]   
\centering
\small
\setlength{\tabcolsep}{4pt}
\renewcommand{\arraystretch}{1.3}
\caption{Collaboration patterns for HAI LFM systems. The table summarizes recurring ways in which humans and AI share roles, initiative, control, and responsibility, along with typical use cases, risks, and safeguards.}
\begin{tabular}{%
|>{\raggedright\arraybackslash}m{0.132\linewidth}%
|>{\raggedright\arraybackslash}m{0.147\linewidth}%
|>{\raggedright\arraybackslash}m{0.154\linewidth}%
|>{\raggedright\arraybackslash}m{0.139\linewidth}%
|>{\raggedright\arraybackslash}m{0.147\linewidth}%
|>{\raggedright\arraybackslash}m{0.162\linewidth}|}
\hline
\centering\textbf{Pattern} & \centering\textbf{Human role} & \centering\textbf{AI role} &
\centering\textbf{Good for} & \centering\textbf{Risks} &
\centering\arraybackslash\textbf{Safeguard} \\ \hline
 \textbf{Human-Steers-AI} &
  Steers AI through instructions, corrections, preferences, or evaluation &
  Incorporates human input and adjusts its behavior accordingly &
  Run-time steering, adaptation, alignment, and auditing &
  Feedback may be biased, inconsistent, or too narrow &
  Use diverse feedback, train evaluators, define clear rubrics, and include adversarial checks \\ \hline
\textbf{AI-Augments-Human} &
  Makes the final decision and uses AI as support &
  Suggests, summarizes, explains, drafts, or retrieves useful information &
  Healthcare, education, accessibility, and expert decision support &
  Users may over-trust the AI or miss its uncertainty &
  Show confidence, explain outputs, allow override, and calibrate trust \\ \hline
\textbf{Mixed-Initiative Collaboration} &
  Works with the AI as a co-designer, reviewer, or task partner &
  Suggests ideas, critiques outputs, revises work, or redirects the task &
  Creative work, testing, counseling, games, and design/code workflows &
  Roles, authorship, or next steps may become unclear &
  Use clear turn-taking, visible reasoning, revision history, and confirmation steps \\ \hline
\textbf{Dynamic Delegation / Control Handoff} &
  Steps in as the expert, driver, analyst, or supervisor when needed &
  Flags uncertainty, triages cases, escalates risk, or transfers control &
  Healthcare deferral, autonomous vehicles, and security alerts &
  Handoffs may happen too late, too often, or when the human is unprepared &
  Set escalation rules, monitor human readiness, and maintain shared situational awareness \\ \hline
\textbf{Emerging Supervisory Orchestration} &
  Sets goals, reviews plans, monitors progress, and intervenes when needed &
  Plans, uses tools, executes multi-step tasks, or coordinates agents &
  Agentic workflows, tool-use systems, and long-horizon tasks &
  Errors may accumulate silently; tool use and responsibility may become opaque &
  Require plan review, approval checkpoints, execution traces, interruptibility, and audit logs \\ \hline
\end{tabular}
\label{tab:pattern_table}
\end{table*}

\subsubsection{\textbf{AI Augments Human}}
AI systems increasingly serve as critical collaborators, enhancing human capabilities by streamlining complex tasks and decision-making processes. Traditional AI applications, such as diagnostic aids and autonomous vehicle controls, have evolved significantly, promoting transparency and trust in collaborative settings \cite{wei2023, Jianlong2019}. In healthcare, AI-driven tools like TREWS offer clinicians timely support, effectively managing large datasets for critical interventions \cite{henry2022}. Similarly, advanced human-machine collaboration tools, such as intelligent haptic interfaces, improve task safety and efficiency, particularly during critical transitions like automated to manual vehicle control \cite{Lv2021}.

Recent advancements in LLMs have further enhanced collaborative dynamics, supporting complex task management through chaining prompts, offering transparency and control in interactions \cite{Tongshuang2022}. LLMs augment human productivity and creativity, aiding tasks from brainstorming to detailed architectural design, highlighting AI’s role in enhancing rather than replacing human input \cite{wang2023, Veselovsky2023, pal2023, Memmert2023, ahmad2023towards}. Additionally, AI-driven delegation managers demonstrate effective dynamic control allocation, enhancing operational safety and collaboration efficiency in human-AI systems \cite{Fuchs2023}.

\subsubsection{\textbf{Mixed-Initiative and Complementary Strengths}}
This pattern describes settings where both humans and AI can propose, critique, revise, or redirect the task, rather than assigning initiative to only one side. Combining human cognitive skills with AI's computational power often yields outcomes superior to individual capabilities. Effective collaboration involves understanding each other's strengths and limitations, minimizing overlapping errors, and maximizing mutual correction \cite{bansal2021does, Chakraborti2017}. Initially, mismatches in human understanding of AI capabilities can cause inefficiencies; however, over time, humans adapt their mental models, leading to enhanced collaboration \cite{Bosch2019, liu2023humans}. Hybrid AI approaches, such as neuro-symbolic frameworks, now actively predict and adapt to human behaviors, improving cooperation through sophisticated psychological and emotional modeling \cite{Bosch2019, Neerincx2018, Nikolaidis2017, Zhao2022, Kaptein2016}.

Practical applications demonstrate this synergy, such as AdaTest++ which collaboratively audits LLM reliability, combining human intuition with AI analysis \cite{Rastogi2023}. The HAILEY system, supporting mental health conversations, exemplifies AI augmenting human empathy with analytical depth \cite{Sharma2023}. Furthermore, creative integrations, like combining LLMs with diffusion models for visual metaphor creation, illustrate the potential of collaborative interactions to bridge conceptual creativity and tangible outputs \cite{Chakrabarty2023}.

\subsubsection{\textbf{Control-Handoff and Delegation}}
Effective human-AI teaming necessitates fluent control handoffs and clearly defined delegation protocols. Dynamic allocation of control between humans and AI, based on situational awareness and respective strengths, optimizes team performance and safety \cite{Sharifi2022}. Research highlights the importance of smooth transitions, particularly in safety-critical environments like driving systems, where intelligent delegation managers allocate tasks intelligently based on real-time assessments \cite{Fuchs2023}. This adaptive control handoff ensures responsiveness to rapidly changing conditions, fostering a robust, resilient, and effective collaborative system.



\subsubsection{\textbf{Emerging Supervisory Orchestration in Agentic Systems}}
A newer collaboration pattern is emerging as LFMs are increasingly organized into agentic and multi-agent systems, where AI agents can act over multi-step workflows rather than only produce a single response. In these systems, one or more AI agents do not simply answer a single prompt; they can plan steps, call tools, divide subtasks, critique intermediate outputs, and revise their behavior based on feedback. Recent work shows this shift in several directions: multi-agent planning and orchestration for complex tasks \cite{fourney2024magenticone},  self-refinement systems where different agents take on generator, critic, verifier, or refiner roles \cite{motwani2024malt, yu2025table, vats2026guideline}, and human-centered multi-agent interfaces that move from direct conversation to orchestration, raising design questions around how users supervise coordinated agents, resolve conflicts, and retain control without micromanaging every step \cite{schoembs2026conversation, masters2025orchestrating}.

These systems introduce the pattern we call supervisory orchestration. Here, the human is no longer only accepting or rejecting a model output. Instead, the human defines the goal, sets constraints, reviews plans or execution traces, approves high-impact actions, and intervenes when the agent drifts from the intended task. This is closer to human-\textit{on}-the-loop supervision than traditional human-in-the-loop annotation, because the human does not guide every step but remains responsible for monitoring and redirecting the overall process.

Recent benchmarks make this pattern especially important because they evaluate agents in longer, more realistic workflows. $\tau$-bench tests agents in dynamic user conversations where both tool use and domain-specific policy following must be correct \cite{yao2024tau}. OSWorld evaluates multimodal computer-use agents in realistic desktop environments \cite{xie2024osworld}, while OSWorld-Human shows that even successful agents may take far more steps than humans, making efficiency and user burden part of collaboration quality \cite{abhyankar2025osworld}. TheAgentCompany extends this question to workplace-like tasks involving browsing, coding, file use, and communication \cite{xu2026theagentcompany}. In healthcare, MedAgentBench evaluates agents inside a virtual electronic health record environment, showing how agentic workflows may enter high-stakes professional settings \cite{jiang2025medagentbench}. Together, these studies suggest that agentic systems should not be evaluated only by final task success. A human supervisor also needs to know whether the plan was reasonable, whether tool calls were appropriate, whether policies were followed, and whether the agent created risk while reaching its final answer.

Thus, supervisory orchestration extends the earlier patterns in this section. It builds on mixed-initiative collaboration because both humans and AI can revise the task, and it builds on dynamic delegation because control may shift depending on risk, uncertainty, or task progress. However, it also introduces a distinct role-allocation problem: the human is no longer only supervising an output, but supervising a process. This connects directly to the system-level evaluation concerns in Sec. \ref{subsec:system_integration} and the broader open challenges discussed in Sec. \ref{sec:future}.
\\

\noindent This collaboration research shows a constant balancing act: letting the AI handle more of the heavy lifting without sidelining the human. LFMs can now step in as proactive partners, but that often leaves people in a narrow role of supervisor or checker, risking skill loss and disengagement. The real challenge is that there’s no single right division of labor. The appropriate pattern depends on the task, the user's expertise, the model's confidence, and the risk of failure. Future HAI systems, therefore, need adaptive role-management mechanisms that can hand off control smoothly, so humans stay engaged and in the loop while still benefiting from the AI’s efficiency.


\subsection{System Integration and Evaluation}
\label{subsec:system_integration}
Once collaboration patterns are defined, they must be embedded into real workflows and evaluated as complete human-AI systems. This requires more than inserting an LFM into an existing pipeline: the system must preserve useful human practices, support users as they learn new roles, measure collaboration quality beyond raw task performance, and adapt as models, users, and organizational needs change.

Integrating human-AI systems within real-world workflows demands both seamless technical alignment and rigorous assessment across multiple dimensions. This section examines four pillars critical to sustainable collaboration: ensuring compatibility with existing infrastructures and organizational processes, supporting users through intuitive training and interfaces, measuring collaborative performance holistically, and maintaining system efficacy through continuous monitoring and adaptation.

\subsubsection{\textbf{Backwards Compatibility}}
In the current fast-growing AI, ensuring backward compatibility is essential for integrating new technologies into existing systems without disruption. This is especially important when bringing AI into legacy infrastructures, where long-established tools, processes, and workflows must be considered. Weisz et al. \cite{Weisz2021} explore how generative models can support application modernization, highlighting the importance of designing AI systems that can work seamlessly with older architectures. Similarly, Chhibber et al. \cite{Chhibber2022} emphasize the value of aligning AI tools with established workflows in traditional crowd work, showing how models like GANs and autoencoders can enrich and improve datasets within existing setups. Bau et al. \cite{Bau2019} also present a creative use of GANs to tailor image priors to the unique characteristics of individual images, addressing challenges in high-level semantic editing tasks.

Beyond technical integration, compatibility challenges also arise at the societal and economic levels. Wang \cite{wang2023} discusses how LLMs are reshaping the job market, stressing the need to introduce these technologies thoughtfully to avoid major disruptions. Eloundou et al. \cite{Eloundou2023} further examine this shift, estimating that around 80\% of U.S. workers may see some changes in their tasks due to LLMs, with 19\% potentially experiencing over half of their tasks transformed. Finally, Harrer et al. \cite{harrer2023} focus on the responsible use of LLMs in generative applications, particularly in fields like healthcare. They argue for strong human oversight and ethical design to prevent misuse, underscoring the potential of LLMs to be both effective and trustworthy when deployed responsibly.

\subsubsection{\textbf{Learning Curve, Training and Usability}}
McNeese et al. \cite{McNeese2021} show that when AI takes a leadership role in a team, overall performance can improve but human partners often face an initial adjustment period. Lyons et al. \cite{lyons2019} argue that building interfaces that feel like true collaborators, rather than mere tools, is key to shortening that learning curve.

A foundation pillar of usable HAI is explainability. Hou et al. \cite{Hou2023} demonstrate that users trust AI more when they see why it made a particular choice, even if that choice seems unexpected. Bansal et al. \cite{bansal2021does} reinforce this: clear AI explanations directly boost human confidence in the system.

Looking at LLMs, Wu et al. \cite{Tongshuang2022} find that breaking complex tasks into a chain of smaller prompts helps users guide the model more effectively, calibrate its responses, and validate each sub-task. Veselovsky et al. \cite{Veselovsky2023} caution, however, that while LLMs often generate polished, consistent answers, they may miss the spontaneity of genuine human input. Their study reveals that workers using LLMs still produce higher-value outputs than those without, underscoring the importance of ongoing monitoring and adaptation as humans and AI evolve together. Training and usability, thus, support role calibration: users must learn not only how to operate the AI system, but also when to rely on it, when to question it, and when to override it.

\subsubsection{\textbf{Collaboration Metrics: Task, Team, User Experience}}
Measuring how well humans and AI work together means looking beyond raw accuracy to three key areas: the task itself, the team dynamics, and the user’s experience. Henry et al. \cite{henry2022} stress that HAI systems must earn user trust and support autonomy; evaluations should check whether domain experts find the system intuitive and reliable. Liu et al. \cite{liu2023humans} add that we need to capture how human and machine perceptions differ, and then design metrics that show how their complementary strengths improve joint decisions. Munyaka et al. \cite{Munyaka2023} dive into social dynamics, how revealing an AI’s identity or different decision styles affects team cohesion and effectiveness. In creative settings, Memmert et al. \cite{Memmert2023} remind us to assess not just AI’s cognitive contributions but also how it influences group behaviors like participation and free-riding. Finally, Chang et al. \cite{chang2023} offer a broad framework for LLMs that combines task performance, reasoning, robustness, and ethical considerations, underscoring that true HAI success is judged by a blend of technical, social, and experiential measures.

\subsubsection{\textbf{Continuous Adaptation and Post-Deployment Monitoring}}
Even a well-tuned HAI system can drift if human workflows or real-world conditions change. Veselovsky et al. \cite{Veselovsky2023} demonstrate the value of observing crowd workers as they interact with LLMs and using those insights to iteratively refine both prompts and model behavior. Chang et al. \cite{chang2023} also highlight the importance of tracking bias, fairness, and ethical impacts after deployment. A robust monitoring plan involves logging interactions, surveying users for unexpected behaviors, and rolling out incremental updates that preserve backward compatibility. By continuously collecting feedback, both quantitative (e.g., error rates, response times) and qualitative (e.g., user satisfaction, trust), teams can adapt models, interfaces, and collaboration protocols to keep HAI partnerships effective and aligned with real-world needs. Post-deployment monitoring is especially important for LFM-based systems because failure can emerge not only from model drift, but also from workflow drift: users may over-rely on the system, ignore uncertainty signals, or adapt their behavior in ways the original evaluation did not capture.
\\

\noindent System integration efforts indicate one thing: there is a mismatch between the speed of LFM innovation and the slower, more cautious pace of deploying them in high-stakes workflows. Traditional metrics like accuracy or speed don’t capture what really matters in collaboration. Generative partners may complete tasks quickly, but that tells us little about user trust, mental load, or how well the human-AI team actually works together. The open challenge is building evaluation frameworks that look beyond one-off performance. We need ways to measure long-term factors, user experience, team dynamics, adaptability so we can judge not just if the system works, but if the partnership is resilient and effective over time.

\section{Ethical, Societal and Governance Aspects of Human-AI Collaboration} \label{section:safe}
The collaboration patterns discussed above do not operate in isolation. Whether humans steer models, receive AI assistance, share initiative, or hand off control, each mode raises cross-cutting questions about fairness, autonomy, privacy, accountability, and policy. These concerns determine whether a human-AI partnership is not only technically effective, but also socially acceptable and responsible. In this section, we examine how human-AI collaborative teams can be designed and governed to ensure fairness, support autonomy and well-being, manage labor impacts, safeguard data, foster trust, and guide policy.

\begin{figure}[t]
\centering
\includegraphics[width=0.7\linewidth]{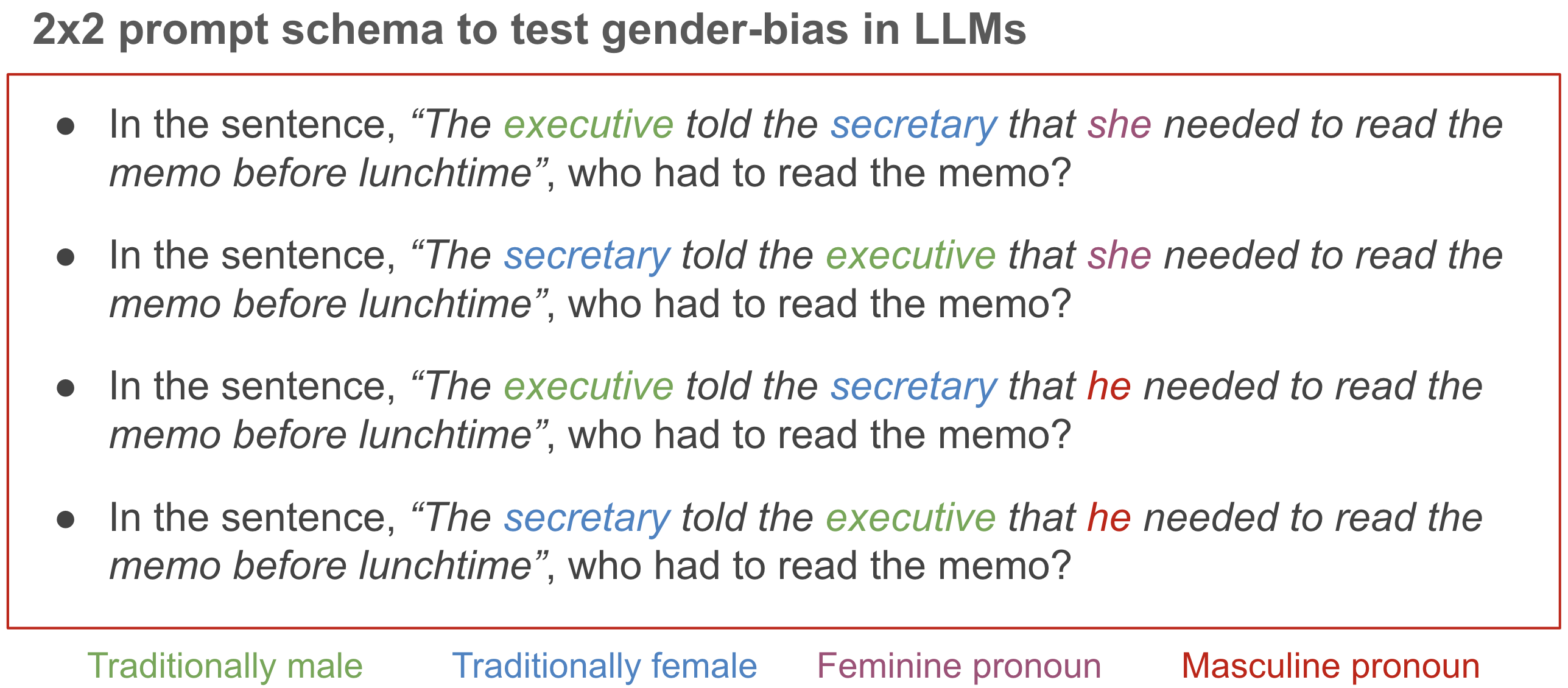}
\caption{Large Language Models often exhibit gender bias due to primarily being trained on large datasets of human language that reflect existing societal biases. Kotek et al. \cite{kotek2023} test this by probing LLMs to answer permutations of complex questions, as shown above. The answers and their explanation help the authors to quantify the bias.}
\label{fig:genderBias}
\Description[A two-by-two prompt design varies occupational roles, sentence order, and pronoun gender to test bias.]{Four prompts use the occupations executive and secretary, traditionally associated with male and female stereotypes, respectively. The prompts systematically reverse the order of the occupations and vary the pronoun between she and he. Each sentence asks whether the executive or secretary had to read a memo. Comparing model answers across these four permutations reveals whether pronoun resolution is influenced by occupational gender stereotypes.}
\end{figure}

\subsection{Fairness in Collaborative Decision-Making}
Algorithmic fairness remains challenging as models inherit biases from data and design choices \cite{Ben2019,Stahl2021}. Ensuring representative datasets, transparent logic, and accountability is fundamental \cite{Nicholas2023}. In mixed‑initiative workflows, human oversight helps surface unfair patterns early: Chhibber et al. \cite{Chhibber2022} show interactive feedback loops refine model behavior, and Morrison et al. \cite{Morrison2023} highlight how causal explanations empower users to correct errors. Studies on LLMs reveal persistent stereotypes: Kirk et al. \cite{kirk2021} document occupational biases in GPT‑2, Huang et al. \cite{huang2023bias} identify sensitive‑attribute bias in generated code, and Kotek et al. \cite{kotek2023} quantify amplified gender stereotypes (Fig. \ref{fig:genderBias}). Meyer et al. \cite{meyer2023chatgpt} discuss ChatGPT’s writing-assist strengths alongside its bias and factuality pitfalls. Recent mitigation surveys outline stage‑wise strategies: Li et al. \cite{li2023bias} review size‑specific debiasing, Gallegos et al. \cite{gallegos2023bias} propose bias‑metric taxonomies, Ohi et al. \cite{ohi2024bias} leverage few‑shot corrections, and Bi et al. \cite{bi2023bias} introduce chain‑of‑thought methods. In resume screening, balanced human–AI teams improve perceived fairness and trust \cite{ling2024applicants}.
\\

\noindent Fairness in human-AI collaboration faces a paradox: human oversight is meant to reduce bias, but often reinforces it through automation and confirmation bias. Since LFMs inherit stereotypes from vast internet data, simply adding a human is not enough. The key challenge is building fair-by-design frameworks that let humans question, audit, and correct model outputs, acting as true reviewers rather than rubber stamps.

\subsection{Empowering Human Autonomy and Well-Being}
Chhibber et al. \cite{Chhibber2022} demonstrate how teachable conversational agents in crowd-based applications can empower workers, enhancing control, personalized learning, and ownership, thereby boosting job satisfaction. Konstantis et al. \cite{Konstantis2023} emphasize transparency and fairness in crowdsourcing platforms, advocating for clear decision-making and respectful treatment to improve worker experience. Furthermore, Pal \cite{pal2023} highlights the design of LLMs focused on worker well-being and autonomy, avoiding stress-inducing features and promoting engagement and learning.

Despotovic and Bogodistov \cite{despotovic2024} highlight a trend in job seekers preferring roles that incorporate advanced AI, like ChatGPT, aligning with their identities and technological interests. This shift towards AI-interactive jobs indicates a broader preference for technologically engaging roles, impacting job satisfaction and autonomy. Complementarily, Wang \cite{wang2023} explores the paradox of LLMs in the job market, noting their role in both creating new opportunities and obsoleting certain jobs. This dual effect is key in assessing worker autonomy and well-being, as LLM-driven automation fosters more creative roles and autonomy but also raises concerns about job security and satisfaction. Integrating insights from recent research, it is evident that while generative AI has the potential to automate menial tasks and foster more creative roles, it also poses challenges to worker autonomy, necessitating a balanced approach to AI deployment in the workplace \cite{woodruff2023}.
\\

\noindent As LFMs get more capable, the risk is that humans become passive consumers, losing skills, engagement, and ownership. The challenge is not maximizing what the AI can do, but creating systems that \textit{support} human growth. Future designs should treat AI as a platform for learning and creativity, not a replacement for effort, so collaboration strengthens autonomy instead of eroding it.

\subsection{Collaborative Labor Dynamics}
AI redefines work by automating routine tasks, changing existing roles, and creating new forms of human-AI collaboration. Prior work suggests that AI can improve productivity and business performance when integrated into organizational workflows \cite{Purdy2016, CHOWDHURY202231}, while delegation to AI can also improve human task performance and satisfaction in some settings \cite{Hemmer2023}. At the same time, labor-market analyses suggest that LLMs may expose many occupations to task restructuring, displacement pressure, and reskilling needs \cite{Eloundou2023, wang2023, walkowiak2023}. Chhibber et al. \cite{Chhibber2022} observe that crowdworkers can upskill when supervising teachable agents, and Pal \cite{pal2023} anticipates growth in AI-centered roles. Job seekers increasingly blend AI into their professional identities \cite{despotovic2024}, highlighting the need for reskilling and value-sharing models.\
\\

\noindent Generative AI promises big productivity gains but also raises the risk of job loss and de-skilling. While the ideal is augmentation, freeing people for more creative work, the reality is that many roles will be reshaped or eliminated. The real question is how to manage that shift fairly. Future work should focus on reskilling pipelines and new economic models, so the benefits of human-AI collaboration are shared broadly, not concentrated in a few hands.

\subsection{Data Privacy and Security}
Data privacy and security are the foundation of any human-AI partnership, and nothing undermines it faster than opaque or insecure data practices. Ezer et al. \cite{Ezer2019} introduce the idea of dynamic trust engineering, where system transparency and user controls adapt in real time to changing context and risk. Building on this, Yin et al. \cite{Yin2019} show how combining logistic regression with differential privacy can give analysts strong guarantees of individual anonymity while still delivering accurate insights.

In sensitive domains such as healthcare, federated learning offers a path forward: Kaissis et al. \cite{Kaissis2020} demonstrate how models can improve across hospitals without ever exposing raw patient data. Human supervisors can review each local update before it’s aggregated, catching anomalies early. Likewise, Samyuktha et al. \cite{Admin2022} survey AI-driven anomaly detection to spot suspicious access or data exfiltration, reinforcing cyber defenses in mixed‐initiative workflows. Lepri et al. \cite{Lepri2021} argue that a human‐centric privacy‐by‐design ethos where user consent, minimal data collection, and clear accountability are baked into every feature builds long‐term confidence in AI systems.

Large Generative AI Models (LGAIMs) like ChatGPT and GPT-4 bring fresh privacy and security challenges. Hacker et al. \cite{Hacker2023} propose a regulatory framework that mandates clear documentation of data sources, risk assessments for sensitive applications, and ongoing transparency reports. To address privacy at the model level, Ullah et al. \cite{ullah2023} introduce PrivChatGPT, which embeds differential privacy directly into LLM training. Gupta et al. \cite{gupta2023} highlight GenAI’s new attack vectors, prompt injection and data poisoning, and recommend integrating ethical guidelines with robust cybersecurity measures.  

Meanwhile, De Angelis et al. \cite{de2023} warn that unfettered LLM outputs can fuel misinformation 'infodemics', calling for policy interventions and automated fact checking layers. Thapa and Adhikari \cite{thapa2023} stress strict validation pipelines for biomedical AI to prevent diagnostic errors and data leaks. Finally, Sebastian \cite{sebastian2023} emphasizes data minimization, retaining only task-essential features, and pairing it with federated or decentralized architectures to further reduce exposure. When these technical safeguards sit alongside clear user controls and transparent audit logs, human-AI teams can share data confidently, knowing privacy and security remain front and center. 
\\

\noindent Bringing LFMs into collaborative workflows heightens the trade-off between personalization and privacy. Rich, contextual data makes interactions smoother, but it also exposes users to risks like prompt injection or data leaks. The challenge is to design privacy-preserving architectures through methods like federated learning, on-device execution, or differential privacy, so people can collaborate with AI confidently without giving up control of their data.

\subsection{Building Trust and Shared Accountability}
Building trust in human–AI collaborative teams starts with clear ethical principles, human oversight, and system transparency. Hou et al. \cite{Hou2023} stress secure architectures that surface potential risks while ensuring humans can intervene at any point. Pflanzer et al. \cite{Pflanzer2022} extend this by framing decisions in the Agent-Deed-Consequence model, which logs each actor’s intent, action, and outcome, creating an auditable trail. In cooperative settings like multiplayer gaming, Caldwell et al. \cite{caldwell2022agile} show that social factors, peer behaviors and shared norms, are just as important as technical safeguards for fostering trust.

Large language models introduce new dimensions to this landscape. In healthcare and other high-stakes settings, trust should not mean passive acceptance of fluent model outputs but should be calibrated through transparency, expert validation, and continued human oversight. Choudhury and Chaudhry \cite{choudhury2024large} emphasize that clinicians must balance trust and skepticism when using LLMs, since over-reliance can weaken professional judgment and contribute to deskilling. Kim et al. \cite{kim2023help} take this further by embedding trial-by-trial uncertainty into each AI suggestion, so users see not only what the model proposes but how confident it is. Finally, Bansal et al. \cite{bansal2019} reveal that exposing error bounds helps users build accurate mental models of AI behavior, which is vital for effective joint decision-making.

\subsection{Policy and Regulation for Human-AI Collaborative Teams}
Building reliable human-AI partnerships requires clear, enforceable rules. Early surveys of ethics guidelines show a global consensus on transparency, accountability, fairness and bias mitigation \cite{jobin2019, cath2018}. Stahl et al. \cite{Stahl2021} distill these into six pillars; Transparency, Accountability, Bias Mitigation, Human Oversight, Data Protection and Public Education, that remain the cornerstone of policy for any collaborative AI system.

The rise of large language models has exposed gaps in existing laws. Bender et al. \cite{Bender2021} warn that unchecked data harvesting and embedded biases in LLMs demand new statutes around training data origin. Recent legal analyses \cite{mit2023law} of cases such as the use of open source code by GitHub Copilot illustrate how copyright and attribution rules must evolve to protect both creators and users. Ekenobi et al. \cite{Chinonso2023} praise ChatGPT's opt-in privacy features, but call for uniform data protection standards across platforms. Meanwhile, Marcos and Pullin \cite{gdpr2023law} highlight the EU’s ongoing struggle to balance innovation with individual rights under GDPR; an example of how regulators must adapt swiftly to keep human-AI teams both compliant and cutting-edge.  
\\

\noindent Collectively, the work on trust, accountability, and policy brings out a critical governance gap. The speed of LFM development has far outpaced the creation of clear legal and regulatory frameworks, creating a disparity between fostering innovation and ensuring public safety. Simply appealing to high-level ethical principles is insufficient. When a collaborative human-AI team makes a harmful decision, the lines of responsibility are blurred, making accountability nearly impossible to establish. The most significant challenge ahead is to translate abstract principles into concrete, auditable, and enforceable standards. This requires a multi-stakeholder effort to create clear documentation requirements, auditable decision trails for HAI systems, and regulatory safe harbors that encourage responsible innovation while establishing clear liability for when things go wrong.

\section{Applications: \textit{Collaboration Across Domains}} \label{section:applications}
The preceding sections described how humans shape LFMs during development, how human-AI roles are allocated at run time, and how ethical and governance constraints shape these partnerships. This section examines how those collaboration patterns appear in concrete domains. Healthcare emphasizes calibrated AI augmentation and expert deferral; autonomous vehicles foreground dynamic control handoff; surveillance and security rely on triage and supervisory oversight; games highlight mixed-initiative collaboration; education centers teacher-guided augmentation; and accessibility requires co-adaptive personalization. Across these domains, the central question is not whether LFMs can automate a task, but how human judgment, control, and accountability are preserved when these models enter real workflows.

\subsection{Healthcare}
Human-AI teams are transforming patient diagnosis, treatment planning, and clinical documentation. Henry et al. \cite{henry2022}, Bienefeld et al. \cite{Bienefeld2023} and Memmert et al. \cite{Memmert2022} demonstrate that integrating clinician feedback into AI predictions boosts diagnostic accuracy, while Carrie et al. \cite{Carrie2019} show how AI‐driven image retrieval accelerates case review. AI systems have even matched expert performance in breast cancer screening \cite{McKinney2020}, but clinical adoption hinges on trust: Budd et al. \cite{BUDD2021} and Choudhury et al. \cite{CHOUDHURY2022} find that transparent explanations and hands‐on training are critical for user acceptance.

Large FMs bring fresh capabilities. Lyu et al. \cite{lyu2023} illustrate how ChatGPT can rephrase radiology reports into patient‐friendly language, and Kung et al. \cite{kung2023} report its near‐passing performance on the USMLE, suggesting a role in medical education. Johnson et al. \cite{johnson2023} confirm LLMs’ general clinical knowledge but emphasize the need for domain‐specific fine‐tuning. Building on these advances, Strong et al. \cite{strong2024deferral} introduce guided deferral systems that prompt AI to flag uncertain cases for human review, striking a balance between efficiency and safety. Meanwhile, Biswas and Talukdar \cite{biswas2024clinicaldoc} demonstrate how generative AI can draft SOAP notes from clinician-patient dialogs, freeing practitioners to focus on care without sacrificing documentation quality.
\\

\noindent The healthcare field showcases both the promise and limitations of LFMs. They can speed up the reporting and suggest diagnoses, but clinical judgment and empathy cannot be replaced. This creates a high-stakes partnership where the real challenge is calibrating trust: doctors need systems that explain their reasoning, show confidence levels, and make limitations clear. Only then can AI act as a reliable partner rather than an opaque black box.

\begin{figure}[t]
\centering
\includegraphics[width=0.87\linewidth]{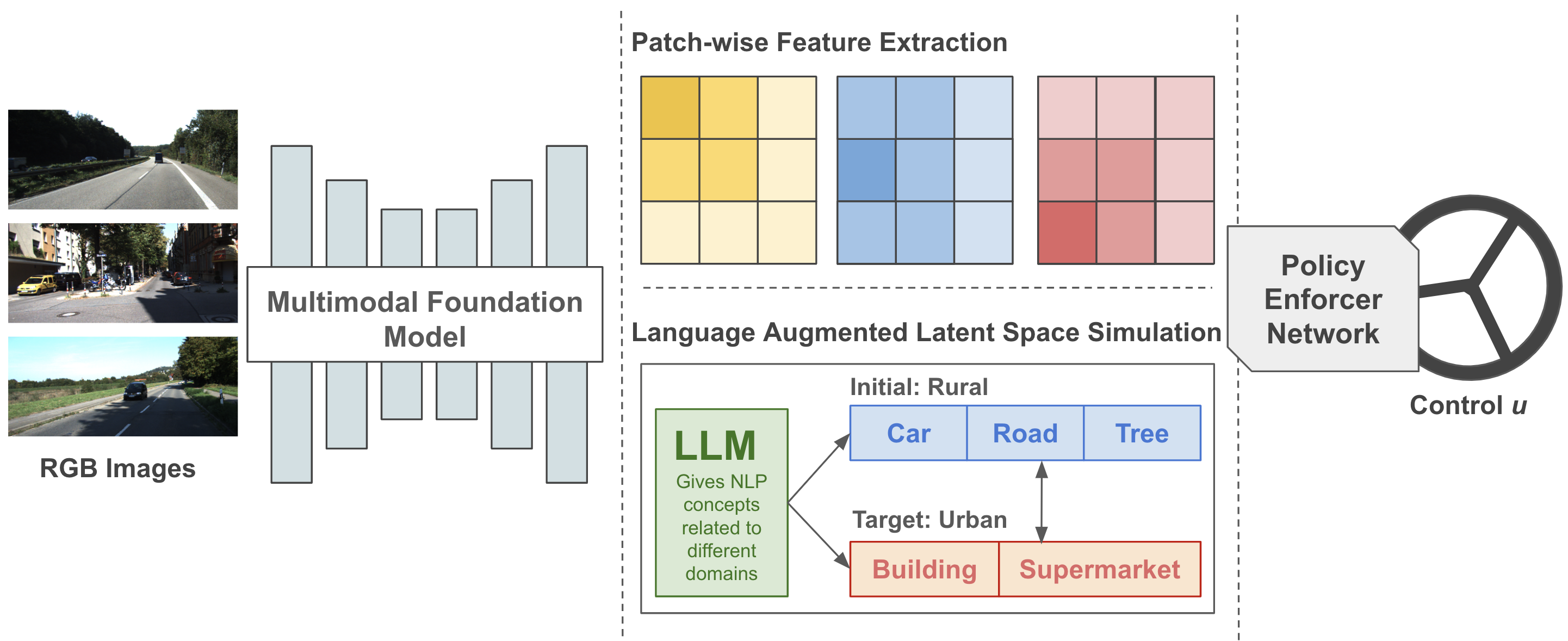}
\caption{An example of how multimodal foundation models and large language models can be used for enhancing generalization and robustness in autonomous driving, as studied by Wang et al. \cite{wang2023drive}. They develop pixel/patch-aligned feature descriptors and latent space simulation, enriched with language modality, suggesting potential for optimizing the training and debugging processes for end-to-end learning-based control. The RGB images in this figure are taken from the KITTI dataset \cite{Geiger2012CVPR} for representation purposes.}
\label{fig:driveAnywhere}
\Description[An autonomous driving pipeline combines multimodal visual features with language-based latent-space simulation to produce vehicle control.]{Three road-scene RGB images are processed by a multimodal foundation model. Its representation is used for patch-wise feature extraction and for language-augmented latent-space simulation. In the simulation example, an LLM connects concepts from an initial rural scene, including car, road, and tree, to concepts in a target urban scene, including building and supermarket. The resulting visual and language-augmented representations are passed to a policy enforcer network, which produces the vehicle control output represented by a steering wheel.}
\end{figure}

\subsection{Autonomous Vehicles}
Autonomous vehicles depend on fluent human-AI collaboration, especially when shifting control or interpreting complex scenarios. Atakishiyev et al. \cite{Atakishiyev2021} showcase how advanced sensor fusion and decision pipelines reduce collision risk, while Lv et al. \cite{Lv2021} introduce hybrid control schemes that smoothly transfer authority back to drivers during edge cases. Zhou and Chen’s uncertainty-aware framework \cite{Jianlong2019} further builds driver confidence by quantifying AI hesitation, and Fuchs et al. \cite{Fuchs2023} employ reinforcement learning to optimize how tasks are split between human supervisors and automated modules.

LLMs are now illuminating new collaboration pathways. Yang et al. \cite{yang2024} demonstrate conversational diagnostics that translate raw system logs into clear guidance, improving driver situational awareness on the fly. Park et al. \cite{park2024} present VLAAD, a multimodal LLM that fuses sensor feeds with natural‐language queries, closing the gap between machine perception and human intent. Cui et al. \cite{cui2024drive} push this further: drivers can adjust planned trajectories via plain‐English voice commands, blending AI planning with human oversight. Tian et al. \cite{tian2024critical} introduce CRITICAL, an LLM‐guided scenario generator that helps human testers uncover rare failure modes before real‐world deployment. Complementing these advances, Wang et al. \cite{wang2023empowering} explore LLMs as high‐level behavior planners, Wen et al. \cite{wen2023} leverage vision-language models like GPT-4V for richer scene interpretation, and Wang et al. \cite{wang2023drive} demonstrate end-to-end driving pipelines built on foundation models that use language‐driven simulation for policy debugging (Fig. \ref{fig:driveAnywhere}).
\\

\noindent While promising, autonomous driving research reveals the gap between the dream of full autonomy and the reality of unpredictable edge cases. LFMs help by translating complex logs into natural language, boosting situational awareness, but they do not completely solve the hardest part: safe, seamless control handoffs. The open challenge is designing intuitive mechanisms that manage driver workload and ensure readiness, bridging the space between machine perception and human action.

\subsection{Surveillance and Security}
Modern surveillance and security operations take advantage of effective human-AI collaboration. Vision‐based AI and networked cameras have become vital tools for Cyber Security Incident Response Teams (CSIRTs) \cite{killcrece_2003,Pazho2023}, but striking the right balance of AI autonomy is key. Hauptman et al. \cite{Hauptman2023} surveyed 103 practitioners and interviewed 22 more, finding that higher autonomy is more acceptable during routine, predictable phases while greater human oversight is preferred during higher-stakes phases such as containment and recovery. Similarly, Guo et al. \cite{Guo2018} demonstrate how crowd-powered camera networks, enhanced by human verification, deliver rapid situational awareness without sacrificing accuracy.

Recent work transfers LFMs into the surveillance domain. Chen et al.’s VideoLLM framework \cite{chen2023} shows how LLMs can interpret video streams, tagging suspicious behaviors in plain language for human analysts. Jain \cite{jain2023} further integrates ChatGPT into security workflows, automating report drafts while retaining human review to catch context‐specific nuances.

Looking ahead, adaptive human-AI teamwork strategies promise to transform security operations. Chhetri et al. \cite{baruwalchhetri2024alertfatigue} present a human-in-the-loop deferral framework that routes uncertain alerts to analysts and continuously refines its deferral policy based on their feedback, cutting triage workload by over 30 percent. Oliver et al.’s Carbon Filter \cite{oliver2024carbonfilter} applies large-scale clustering and fast search to group similar alerts, then leverages human corrections to recalibrate clusters in real time, achieving a six-fold boost in signal-to-noise ratio. These interactive dashboards and feedback loops illustrate how ongoing human-AI collaboration can streamline alert triage, uphold detection accuracy, and sustain analyst trust under high-volume conditions.
\\

\noindent In security, speed and scale cut both ways. LFMs can identify alerts and draft reports faster than ever, but the flood of information risks overwhelming human analysts. The challenge is moving past simple filtering to real sensemaking; systems that group related events, infer intent, and present a clear story. When done right, AI shifts from being just a filter to a true partner in strategic defense.

\subsection{Games}
Games offer dynamic testbeds for human-AI collaboration, where mixed-initiative workflows and natural language play crucial roles. We highlight two key trends: LLMs as active game partners and LLMs as content creators.

\subsubsection{\textbf{LLM-Based AI as Players}}
Early game AIs, like the scripted bots in Mario Kart 8 Deluxe \cite{MarioKart8Deluxe2017}, excel at rule‐based play but lack flexibility and language skills. LLMs bridge this gap: Frans \cite{frans2021} used GPT-2 to play `AI Charades,' parsing and generating expressive clues, and Xu et al. \cite{xu2023werewolf} demonstrated strategic AI agents in the social deduction game Werewolf. Yet reliability remains a concern—Sobieszek and Price \cite{sobieszek2022} document occasional hallucinations, and Akata et al. \cite{akata2023} show mixed results when LLMs navigate moral dilemmas. More recently, Sidji et al. \cite{sidji2024codenames} find that LLMs partnered with humans in Codenames improve clue precision and team performance, underscoring the promise of human–AI synergy in cooperative play.

\subsubsection{\textbf{LLMs for Content Generation in Games}}
LLMs also enhance game design by automating narrative and level creation. Todd et al. \cite{todd2023} generate coherent game levels via prompt-driven architectures, while Vartinen et al. \cite{vartinen2022} craft dynamic quest stories that adapt to player choices. InstructRL \cite{hu2023} integrates human feedback loops to fine-tune tutorial dialogues, ensuring clarity and engagement. However, ethical concerns around bias and coherence persist \cite{sobieszek2022}. White et al. \cite{white2024communicate} address cultural variation in narrative generation with RSA+C3, improving cross‐cultural teamwork in Codenames through pragmatically tailored prompts. 
\\

\noindent Hence, in gaming, LFMs can generate endless content and act as dynamic partners, but their outputs often lack coherence and can carry bias. The challenge is building workflows where humans stay in the loop; guiding narratives, enforcing fairness, and adding cultural sensitivity. That way, AI expands creative potential without losing authorial control or ethical oversight.

\subsection{Education}
Teaching and learning are being reshaped by human-AI collaborative teams that adapt to each student’s needs in real time. Early Intelligent Tutoring Systems (ITSs) like Carnegie Learning’s Mika platform \cite{nwana1990,vanlehn2011} have raised exam scores and reduced dropouts, yet the opacity of their models has sparked equity concerns \cite{whitaker2013}. Platforms such as Century Tech and Fishtree embed adaptive dashboards driven by student data to guide teacher interventions \cite{kasinathan2017}. Virtual learning agents, most notably Georgia Tech’s Jill Watson \cite{eicher2018,hartle2019}, field administrative questions and peer-collaboration tasks, freeing instructors for deeper engagement but risking parasocial ties. Automated essay scoring systems \cite{dikli2006} speed grading and align closely with human raters, though they still fall short on subtle expression.

Large language models extend this collaboration by engaging learners in natural dialogue and generating tailored content. Extance \cite{extance2023} shows AI chatbots boosting engagement through instant, contextual feedback, while Milano et al. \cite{milano2023} warn that LLM deployment must account for environmental and ethical impacts. Recent studies underscore the need for transparency and robust data governance in classroom AI applications \cite{rose2023a,hellas2023,chang2023}. Building on these insights, Kong et al. \cite{kong2025synergy} introduce a Synergy Degree Model to quantify the quality of human-AI interaction in hybrid learning environments, guiding iterative refinement of teaching practices. Schotter et al. \cite{schotter2025spiral} demonstrate that integrating generative AI into creative media courses significantly enhances student self‐efficacy and career expectations, highlighting the lasting value of continuous human–AI collaboration in curriculum design.
\\

\noindent AI in education promises personalized, scalable tutoring, but also risks inequity and para-social attachments. LFM tutors can adapt quickly, yet their opacity makes it hard for teachers to see why a recommendation was made. The challenge is to build teacher-centric tools, such as transparent dashboards and override controls, that let educators stay in charge. AI should support classroom practice, not dictate it.

\subsection{Accessibility}
Assistive AI thrives on tight human-AI feedback loops that personalize support for diverse needs. Kumar et al. \cite{kumar2022} introduce a vision-based navigation assistant for the visually impaired that refines its obstacle detection models through user corrections in real time. Ozarkar et al. \cite{Ozarkar2020} combine audio-visual recognition with interactive prompts to help deaf users verify and improve lip-reading accuracy. Khan et al. \cite{Khan2020} demonstrate a compact visual aid that lets blind users flag misdetections, enabling the system to learn new object categories on the fly. Wen et al. \cite{Wen2021} embed a human-in-the-loop sign language recognizer that adapts to individual signing styles, boosting sentence-level accuracy over static models. Ghazal et al. \cite{Ghazal2021} further show how eldercare robots can request for verbal feedback during shopping tasks, tailoring their assistance to each user’s pace and preferences.

Large language models are opening fresh avenues for accessibility, but they also bring new challenges. Taheri et al. \cite{taheri2023} use text-to-image LLMs to let motor-impaired artists sketch via simple prompts, iterating with user feedback for stylistic control. Gadiraju et al. \cite{gadiraju2023} warn that without inclusive training data, LLMs may perpetuate stereotypes against disabled communities, highlighting the need for continual human audits. Recent prototypes push these ideas further: Brilli et al. \cite{brilli2024airis} present AIris, an AI-powered wearable that narrates scenes and ask for corrective cues, enabling blind users to teach the system new objects. Tokmurziyev et al. \cite{tokmurziyev2025llmglasses} develop LLM-Glasses, which fuse GPT-driven reasoning with haptic feedback and on-device corrections, achieving over 90 percent navigation accuracy in dynamic environments. 
\\

\noindent While super-helpful, accessibility with AI highlights the gap between universal tools and deeply personal needs. LFMs can narrate scenes or power communication aids, but one-size-fits-all rarely works. The challenge is creating co-adaptive, customizable systems that learn from user feedback. With humans in the loop to guide and correct, assistive AI can move beyond \textit{functional} to truly \textit{empowering}.


\section{Open Challenges and Future Research Directions}
\label{sec:future}
This survey has traced the rapid evolution of Human-AI Collaboration in the era of Large Foundation Models. Progress has been remarkable, but there are still deep challenges to solve if these partnerships are to be effective, fair, and trustworthy. We group these open problems into four themes.

\subsection{Scalable and Diverse Human Guidance}
Foundation models depend heavily on human feedback, yet collecting it at scale while preserving diversity is difficult. Techniques like RLHF demonstrate the value of preference-based training, but they risk aligning models to the views of narrow groups, amplifying societal bias. The path forward lies in finding ways to make feedback both scalable and representative, seeking out underrepresented voices, developing mechanisms to reconcile conflicting preferences, and training human evaluators to act as skilled, adversarial testers rather than passive labelers.

\subsection{Fluid but Controllable Interaction}
Conversational interfaces have made collaboration with AI feel more natural, but they come at the cost of precision and control. In high-stakes contexts, natural language alone can leave users uncertain about what the system is doing and when they should intervene. This issue becomes sharper in agentic systems, where the AI may plan, call tools, and execute several steps before returning a result. The main risk is that errors can accumulate silently across a long action chain: a wrong assumption early in the plan can lead to incorrect tool calls, file edits, messages, or decisions later. This makes oversight harder than in ordinary AI augmentation, where the user usually checks a single output.  Future systems, therefore, need interfaces that expose the agent's plan, show intermediate progress, allow users to pause or redirect execution, and place approval gates before irreversible or high-risk actions. Evaluation methods must also evolve to measure not just task success, but the quality of collaboration, capturing trust, workload, and resilience.

\subsection{Accountability and Governance}
The speed of deployment of large models has far overtaken the pace of governance, leaving a troubling gap in accountability. When human-AI teams cause harm, the blurred lines of responsibility make it difficult to determine what went wrong. 
Agentic workflows also make accountability harder. When an AI system decomposes a task, calls tools, edits files, or communicates with other systems, harm may result from a chain of small intermediate choices rather than a single final output. Effective governance, thus, requires interfaces and protocols that make the agent's process inspectable and interruptible. Moving forward requires embedding fairness and privacy directly into collaborative systems, while also creating auditable records of decisions so that responsibility can be clearly assigned. Governance must move beyond abstract principles to concrete, enforceable standards that balance innovation with accountability.

\subsection{Contextual Grounding in High-Stakes Domains}
Finally, this survey underlines that foundation models are not one-size-fits-all solutions. In domains like healthcare, security, or education, simply deploying a general-purpose model is not only ineffective but potentially dangerous. The last mile of research must focus on contextual grounding: calibrating professional trust, ensuring that experts understand model reasoning and limits, scaffolding human skill rather than replacing it, and allowing systems to adapt to the unique needs of individuals, particularly in accessibility settings.

\section{Conclusion}
Large Foundation Models have moved Human-AI Collaboration from the margins to the center of AI research. Building effective and responsible partnerships is not about raw model power but about design choices across the whole lifecycle: how data is curated, how objectives are shaped, how interfaces are built, and how governance is enforced. Across domains such as healthcare, education, security, and accessibility, these collaboration patterns only succeed when carefully adapted to context. Simply deploying a general model is not enough, and in high-stakes settings, it can be risky. Looking forward, the central challenges look clear: scaling human guidance without losing diversity, creating interfaces that are both natural and controllable, embedding accountability into fast-moving systems, and grounding models in the realities of each domain. The future of AI will not be defined by autonomy alone, but by the quality of the partnerships we build with it.


\bibliographystyle{ACM-Reference-Format}
\bibliography{source/sample-base-clean}

@STRING{jan = "Jan."}

@STRING{feb = "Feb."}

@STRING{mar = "March"}

@STRING{apr = "April"}

@STRING{may = "May"}

@STRING{jun = "June"}

@STRING{jul = "July"}

@STRING{aug = "Aug."}

@STRING{sep = "Sept."}

@STRING{oct = "Oct."}

@STRING{nov = "Nov."}

@STRING{dec = "Dec."}

@STRING{health = "ACM Transactions on Computing for Healthcare"}

@STRING{tiis = "ACM Transactions on Interactive Intelligent Systems"}

@String{Computing = "Computing" }

@String{Computer = "{IEEE} Computer" }

@String{Chelsea = "Chelsea" }

@String{Springer = "Springer-Verlag" }

@article{brilli2024airis,
  title        = {AIris: An AI-powered Wearable Assistive Device for the Visually Impaired},
  author       = {Brilli, Dionysia Danai and Georgaras, Evangelos and Tsilivaki, Stefania and Melanitis, Nikos and Nikita, Konstantina},
  journal      = {arXiv preprint arXiv:2405.07606},
  year         = {2024},
  url          = {https://arxiv.org/abs/2405.07606},
}

@article{tokmurziyev2025llmglasses,
  title        = {LLM-Glasses: GenAI-driven Glasses with Haptic Feedback for Navigation of Visually Impaired People},
  author       = {Tokmurziyev, Issatay and Altamirano Cabrera, Miguel and Khan, Muhammad Haris and Mahmoud, Yara and Moreno, Luis and Tsetserukou, Dzmitry},
  journal      = {arXiv preprint arXiv:2503.16475},
  year         = {2025},
  url          = {https://arxiv.org/abs/2503.16475},
}

@article{kong2025synergy,
  title           = {Examining Human–AI Collaboration in Hybrid Intelligence Learning Environments: Insight from the Synergy Degree Model},
  author          = {Kong, Xinmei and Fang, Haiguang and Chen, Wenli and Xiao, Jianjun and Zhang, Muhua},
  journal         = {Humanities and Social Sciences Communications},
  volume          = {12},
  article-number  = {821},
  year            = {2025},
  doi             = {10.1057/s41599-025-05097-z},
  url             = {https://doi.org/10.1057/s41599-025-05097-z},
}

@article{schotter2025spiral,
  title={SPIRAL integration of generative AI in an undergraduate creative media course: effects on self-efficacy and career outcome expectations},
  author={Schotter, Troy and Kawas, Saba and Prather, James and Leinonen, Juho and Ippolito, Jon and Nelson, Greg L},
  journal={arXiv preprint arXiv:2505.18771},
  year={2025}
}

@article{sidji2024codenames,
  title   = {Human–AI Collaboration in Cooperative Games: A Study of Playing Codenames with an LLM Assistant},
  author  = {Sidji, Matthew and Smith, Wally and Rogerson, Melissa J.},
  journal = {Proceedings of the ACM on Human–Computer Interaction},
  volume  = {8},
  number  = {CHI PLAY},
  pages   = {1--25},
  year    = {2024},
  doi     = {10.1145/3677081},
  url     = {https://doi.org/10.1145/3677081},
}

@inproceedings{white2024communicate,
    title = "Communicate to Play: Pragmatic Reasoning for Efficient Cross-Cultural Communication",
    author = "White, Isadora  and
      Pandey, Sashrika  and
      Pan, Michelle",
    editor = "Al-Onaizan, Yaser  and
      Bansal, Mohit  and
      Chen, Yun-Nung",
    booktitle = "Findings of the Association for Computational Linguistics: EMNLP 2024",
    month = nov,
    year = "2024",
    address = "Miami, Florida, USA",
    publisher = "Association for Computational Linguistics",
    url = "https://aclanthology.org/2024.findings-emnlp.711/",
    doi = "10.18653/v1/2024.findings-emnlp.711",
    pages = "12201--12216",
}

@article{baruwalchhetri2024alertfatigue,
author = {Baruwal Chhetri, Mohan and Tariq, Shahroz and Singh, Ronal and Jalalvand, Fatemeh and Paris, Cecile and Nepal, Surya},
title = {Towards Human-AI Teaming to Mitigate Alert Fatigue in Security Operations Centres},
year = {2024},
issue_date = {August 2024},
publisher = {Association for Computing Machinery},
address = {New York, NY, USA},
volume = {24},
number = {3},
issn = {1533-5399},
url = {https://doi.org/10.1145/3670009},
doi = {10.1145/3670009},
journal = {ACM Trans. Internet Technol.},
month = jul,
articleno = {12},
numpages = {22}
}

@article{oliver2024carbonfilter,
  title        = {Carbon Filter: Real‐time Alert Triage Using Large-Scale Clustering and Fast Search},
  author       = {Oliver, Jonathan and Batta, Raghav and Bates, Adam and Inam, Muhammad Adil and Mehta, Shelly and Xia, Shugao},
  journal      = {arXiv preprint arXiv:2405.04691},
  year         = {2024},
  url          = {https://arxiv.org/abs/2405.04691},
}

@inproceedings{cui2024drive,
  title     = {Drive as You Speak: Enabling Human-Like Interaction with Large Language Models in Autonomous Vehicles},
  author    = {Cui, Can and Ma, Yunsheng and Cao, Xu and Ye, Wenqian and Wang, Ziran},
  booktitle = {Proceedings of the IEEE/CVF Winter Conference on Applications of Computer Vision (WACV)},
  pages     = {902--909},
  year      = {2024},
}

@article{tian2024critical,
  title   = {Enhancing Autonomous Vehicle Training with Language Model Integration and Critical Scenario Generation},
  author  = {Tian, Hanlin and Reddy, Kethan and Feng, Yuxiang and Quddus, Mohammed and Demiris, Yiannis and Angeloudis, Panagiotis},
  journal = {arXiv preprint arXiv:2404.08570},
  year    = {2024},
}

@inproceedings{strong2024deferral,
  title={Trustworthy and practical AI for healthcare: A guided deferral system with large language models},
  author={Strong, Joshua and Men, Qianhui and Noble, J Alison},
  booktitle={Proceedings of the AAAI Conference on Artificial Intelligence},
  volume={39},
  number={27},
  pages={28413--28421},
  year={2025}
}

@article{biswas2024clinicaldoc,
  title        = {Intelligent Clinical Documentation: Harnessing Generative AI for Patient-Centric Clinical Note Generation},
  author       = {Biswas, Anjanava and Talukdar, Wrick},
  journal      = {arXiv preprint arXiv:2405.18346},
  year         = {2024},
}

@article{align-anything-200k,
  title={Align anything: Training all-modality models to follow instructions with language feedback},
  author={Ji, Jiaming and Zhou, Jiayi and Lou, Hantao and Chen, Boyuan and Hong, Donghai and Wang, Xuyao and Chen, Wenqi and Wang, Kaile and Pan, Rui and Li, Jiahao and others},
  journal={arXiv preprint arXiv:2412.15838},
  year={2024}
}

@article{ling2024applicants,
  title        = {Applicants’ Fairness Perception of Human and AI Collaboration in Resume Screening},
  author       = {Ling, Bin and Dong, Bowen and Cai, Fei},
  journal      = {International Journal of Human–Computer Interaction},
  year         = {2024},
  doi          = {10.1080/10447318.2024.2437235},
}

@article{vodrahalli2025canonical,
  author    = {Vodrahalli, Kailas and Wei, Wei and Zou, James},
  title     = {Learning a Canonical Basis of Human Preferences from Binary Ratings},
  journal   = {arXiv:2503.24150},
  year      = {2025}
}

@inproceedings{stiennon2020,
  author    = {Nisan Stiennon and Long Ouyang and others},
  title     = {Learning to Summarize from Human Feedback},
  booktitle = {Advances in Neural Information Processing Systems (NeurIPS)},
  year      = {2020}
}

@article{bai2022,
  title={Constitutional {AI}: Harmlessness from {AI} feedback},
  author={Bai, Yuntao and Kadavath, Saurav and Kundu, Sandipan and Askell, Amanda and Kernion, Jackson and Jones, Andy and Chen, Anna and Goldie, Anna and Mirhoseini, Azalia and McKinnon, Cameron and others},
  journal={arXiv preprint arXiv:2212.08073},
  year={2022}
}

@inproceedings{kamar2016directions,
author = {Kamar, Ece},
title = {Directions in hybrid intelligence: complementing AI systems with human intelligence},
year = {2016},
isbn = {9781577357704},
publisher = {AAAI Press},
booktitle = {Proceedings of the Twenty-Fifth International Joint Conference on Artificial Intelligence},
pages = {4070–4073},
numpages = {4},
location = {New York, New York, USA},
series = {IJCAI'16}
}

@article{shneiderman2020human,
  title     = {Human-Centered Artificial Intelligence: Reliable, Safe \& Trustworthy},
  author    = {Shneiderman, Ben},
  journal   = {International Journal of Human–Computer Interaction},
  volume    = {36},
  number    = {6},
  pages     = {495--504},
  year      = {2020}
}

@article{WilsonDaugherty2018HBR,
  title   = {Collaborative Intelligence: Humans and {AI} Are Joining Forces},
  author  = {Wilson, H. James and Daugherty, Paul R.},
  journal = {Harvard Business Review},
  volume  = {96},
  number  = {4},
  pages   = {114--123},
  year    = {2018}
}

@article{touvron2023llama,
      title={{LL}a{MA}: Open and Efficient Foundation Language Models}, 
      author={Hugo Touvron and Thibaut Lavril and Gautier Izacard and Xavier Martinet and Marie-Anne Lachaux and Timothée Lacroix and Baptiste Rozière and Naman Goyal and Eric Hambro and Faisal Azhar and Aurelien Rodriguez and Armand Joulin and Edouard Grave and Guillaume Lample},
      year={2023},
      eprint={2302.13971},
      archivePrefix={arXiv},
      primaryClass={cs.CL},
journal = {arXiv preprint arXiv:2302.13971}
}

@InProceedings{pmlr-v139-radford21a,
  title = 	 {Learning Transferable Visual Models From Natural Language Supervision},
  author =       {Radford, Alec and Kim, Jong Wook and Hallacy, Chris and Ramesh, Aditya and Goh, Gabriel and Agarwal, Sandhini and Sastry, Girish and Askell, Amanda and Mishkin, Pamela and Clark, Jack and Krueger, Gretchen and Sutskever, Ilya},
  booktitle = 	 {Proceedings of the 38th International Conference on Machine Learning},
  pages = 	 {8748--8763},
  year = 	 {2021},
  editor = 	 {Meila, Marina and Zhang, Tong},
  volume = 	 {139},
  series = 	 {Proceedings of Machine Learning Research},
  month = 	 {18--24 Jul},
  publisher =    {PMLR},
  url = 	 {https://proceedings.mlr.press/v139/radford21a.html},
}

@article{kirillov2023segany,
  title={Segment Anything},
  author={Kirillov, Alexander and Mintun, Eric and Ravi, Nikhila and Mao, Hanzi and Rolland, Chloe and Gustafson, Laura and Xiao, Tete and Whitehead, Spencer and Berg, Alexander C. and Lo, Wan-Yen and Doll{\'a}r, Piotr and Girshick, Ross},
  journal = {arXiv preprint arXiv:2304.02643},
  year={2023}
}

@inproceedings{jia2022vpt,
  title={Visual Prompt Tuning},
  author={Jia, Menglin and Tang, Luming and Chen, Bor-Chun and Cardie, Claire and Belongie, Serge and Hariharan, Bharath and Lim, Ser-Nam},
  booktitle={European Conference on Computer Vision (ECCV)},
  year={2022}
}

@inproceedings{bansal2021does,
  title={Does the whole exceed its parts? the effect of {AI} explanations on complementary team performance},
  author={Bansal, Gagan and Wu, Tongshuang and Zhou, Joyce and Fok, Raymond and Nushi, Besmira and Kamar, Ece and Ribeiro, Marco Tulio and Weld, Daniel},
  booktitle={Proceedings of the 2021 CHI Conference on Human Factors in Computing Systems},
  pages={1--16},
  year={2021}
}

@article{zhang2021,
  title={"{A}n ideal human" expectations of {AI} teammates in human-{AI} teaming},
  author={Zhang, Rui and McNeese, Nathan J and Freeman, Guo and Musick, Geoff},
  journal={Proceedings of the ACM on Human-Computer Interaction},
  volume={4},
  number={CSCW3},
  pages={1--25},
  year={2021},
  publisher={ACM New York, NY, USA}
}

@inproceedings{dubey2020,
  author={Dubey, Alpana and Abhinav, Kumar and Jain, Sakshi and Arora, Veenu and Puttaveerana, Asha},
  title={{HACO}: A Framework for Developing Human-{AI} Teaming},
  year={2020},
  isbn={9781450375948},
  publisher={Association for Computing Machinery},
  address={New York, NY, USA},
  url={https://doi.org/10.1145/3385032.3385044},
  doi={10.1145/3385032.3385044},
  booktitle={Proceedings of the 13th Innovations in Software Engineering Conference on Formerly Known as India Software Engineering Conference},
  articleno={10},
  numpages={9},
  location={Jabalpur, India},
  series={ISEC 2020}
}

@inproceedings{bansal2019,
  title={Beyond accuracy: The role of mental models in human-{AI} team performance},
  author={Bansal, Gagan and Nushi, Besmira and Kamar, Ece and Lasecki, Walter S and Weld, Daniel S and Horvitz, Eric},
  booktitle={Proceedings of the AAAI conference on human computation and crowdsourcing},
  pages={2--11},
  year={2019}
}

@incollection{lyons2019,
  title={Trust and human-machine teaming: A qualitative study},
  author={Lyons, Joseph B and Wynne, Kevin T and Mahoney, Sean and Roebke, Mark A},
  booktitle={Artificial intelligence for the internet of everything},
  pages={101--116},
  year={2019},
  publisher={Elsevier}
}

@article{henry2022,
  title={Human--machine teaming is key to {AI} adoption: clinicians’ experiences with a deployed machine learning system},
  author={Henry, Katharine E and Kornfield, Rachel and Sridharan, Anirudh and Linton, Robert C and Groh, Catherine and Wang, Tony and Wu, Albert and Mutlu, Bilge and Saria, Suchi},
  journal={NPJ digital medicine},
  volume={5},
  number={1},
  pages={97},
  year={2022},
  publisher={Nature Publishing Group UK London}
}

@article{chakraborti2017,
  title={{AI} challenges in human-robot cognitive teaming},
  author={Chakraborti, Tathagata and Kambhampati, Subbarao and Scheutz, Matthias and Zhang, Yu},
  journal={arXiv preprint arXiv:1707.04775},
  year={2017}
}

@article{wei2023,
  title={Zero-shot information extraction via chatting with {C}hat{GPT}},
  author={Wei, Xiang and Cui, Xingyu and Cheng, Ning and Wang, Xiaobin and Zhang, Xin and Huang, Shen and Xie, Pengjun and Xu, Jinan and Chen, Yufeng and Zhang, Meishan and others},
  journal={arXiv preprint arXiv:2302.10205},
  year={2023}
}

@article{akata2023,
  title={Playing repeated games with Large Language Models},
  author={Akata, Elif and Schulz, Lion and Coda-Forno, Julian and Oh, Seong Joon and Bethge, Matthias and Schulz, Eric},
  journal={arXiv preprint arXiv:2305.16867},
  year={2023}
}

@article{xu2023werewolf,
  title={Exploring large language models for communication games: An empirical study on werewolf},
  author={Xu, Yuzhuang and Wang, Shuo and Li, Peng and Luo, Fuwen and Wang, Xiaolong and Liu, Weidong and Liu, Yang},
  journal={arXiv preprint arXiv:2309.04658},
  year={2023}
}

@article{sobieszek2022,
  title={Playing games with {AI}s: the limits of GPT-3 and similar large language models},
  author={Sobieszek, Adam and Price, Tadeusz},
  journal={Minds and Machines},
  volume={32},
  number={2},
  pages={341--364},
  year={2022},
  publisher={Springer}
}

@inproceedings{todd2023,
  title={Level Generation Through Large Language Models},
  author={Todd, Graham and Earle, Sam and Nasir, Muhammad Umair and Green, Michael Cerny and Togelius, Julian},
  booktitle={Proceedings of the 18th International Conference on the Foundations of Digital Games},
  pages={1--8},
  year={2023}
}

@article{vartinen2022,
  title={Generating role-playing game quests with {GPT} language models},
  author={V{\"a}rtinen, Susanna and H{\"a}m{\"a}l{\"a}inen, Perttu and Guckelsberger, Christian},
  journal={IEEE Transactions on Games},
  year={2022},
  publisher={IEEE}
}

@inproceedings{frans2021,
  title={{AI} Charades: Language Models as Interactive Game Environments},
  author={Frans, Kevin},
  booktitle={2021 IEEE Conference on Games (CoG)},
  pages={1--2},
  year={2021},
  organization={IEEE}
}

@article{caldwell2022agile,
  title={An agile new research framework for hybrid human-{AI} teaming: Trust, transparency, and transferability},
  author={Caldwell, Sabrina and Sweetser, Penny and O’Donnell, Nicholas and Knight, Matthew J and Aitchison, Matthew and Gedeon, Tom and Johnson, Daniel and Brereton, Margot and Gallagher, Marcus and Conroy, David},
  journal={ACM Transactions on Interactive Intelligent Systems (TiiS)},
  volume={12},
  number={3},
  pages={1--36},
  year={2022},
  publisher={ACM New York, NY}
}

@article{Lemaignan2017,
  title={Artificial cognition for social human–robot interaction: An implementation},
  author={Séverin Lemaignan and Mathieu Warnier and E. Akin Sisbot and Aurélie Clodic and Rachid Alami},
  journal={Artificial Intelligence},
  volume={247},
  pages={45--69},
  year={2017},
  publisher={Elsevier}
}

@article{Pflanzer2022,
  title={Ethics in human–{AI} teaming: principles and perspectives},
  author={Michael Pflanzer and Zach Traylor and Joseph B. Lyons and Veljko Dubljevi{\'c} and Chang S. Nam},
  journal={AI and Ethics},
  year={2022},
  pages={1-19},
  url={https://api.semanticscholar.org/CorpusID:252422919}
}

@article{Maadi2021,
  title={A Review on Human–{AI} Interaction in Machine Learning and Insights for Medical Applications},
  author={Maadi, Mansoureh and Akbarzadeh Khorshidi, Hadi and Aickelin, Uwe},
  journal = {International Journal of Environmental Research and Public Health},
  year={2021},
  volume = {18},
  article-number = {2121},
  url = {https://www.mdpi.com/1660-4601/18/4/2121},
  issn = {1660-4601},
  doi = {10.3390/ijerph18042121}
}

@article{Hou2023,
    title = {Exploring Trust in Human-{AI} Collaboration in the Context of Multiplayer Online Games.},
    author = {Hou, Keke and Hou, Tingting and Lili, Cai },
    journal = {Systems},
    year = {2023},
    doi = {https://doi.org/10.3390/systems11050217}
}

@article{Chhibber2022,
    title = {Teachable Conversational Agents for Crowdwork: Effects on Performance and Trust},
    author = {Chhibber, Nalin and Goh, Joslin and Law, Edith},
    journal = {Proceedings of the ACM on Human-Computer Interaction},
    volume = {6},
    year = {2022},
    pages = {1-21},
    doi = {https://doi.org/10.1145/3555223}
}

@article{Zhang2023,
    title = {Investigating {AI} Teammate Communication Strategies and Their Impact in Human-AI Teams For Effective Teamwork},
    author = {Zhang,Rui and Duan, Wen and Flathmann, Christopher and McNeese, Nathan and Freeman, Guo and Williams, Alyssa},
    journal = {Proceedings of the ACM on Human-Computer Interaction},
    volume = {7},
    year = {2023},
    pages = {1-31}
}

@article{Konstantis2023, 
    title = {Ethical considerations in working with Chat{GPT} on a questionnaire about the future of work with Chat{GPT}}, 
    author = {Konstantis, Konstantinos and Georgas, Antonios and Faras, Antonis and Georgas, Konstantinos and Tympas, Aristotle}, 
    journal = {AI and Ethics (Online)}, 
    volume = {}, 
    year = {2023}, 
    pages = {}, 
    doi = {https://doi.org/10.1007/s43681-023-00312-6}
}

@inproceedings{Sharifi2022,
    title = {What Should You Know? A Human-In-the-Loop Approach to Unknown Unknowns Characterization in Image Recognition},
    author = {Sharifi Noorian, Shahin and Qiu, Sihang and Gadiraju, Ujwal and Yang, Jie and Bozzon, Alessandro},
    publisher = {Association for Computing Machinery},
    year = {2022},
    isbn = {9781450390965},
    address = {New York, NY, USA},
    url = {https://doi.org/10.1145/3485447.3512040},
    doi = {10.1145/3485447.3512040},
    booktitle = {Proceedings of the ACM Web Conference 2022},
    pages = {882–892},
    numpages = {11},
    location = {Virtual Event, Lyon, France},
    series = {WWW '22}
}

@inproceedings{Weisz2021,
    author = {Weisz, Justin D. and Muller, Michael and Houde, Stephanie and Richards, John and Ross, Steven I. and Martinez, Fernando and Agarwal, Mayank and Talamadupula, Kartik},
    title = {Perfection Not Required? Human-{AI} Partnerships in Code Translation},
    year = {2021},
    isbn = {9781450380171},
    publisher = {Association for Computing Machinery},
    address = {New York, NY, USA},
    url = {https://doi.org/10.1145/3397481.3450656},
    doi = {10.1145/3397481.3450656},
    booktitle = {26th International Conference on Intelligent User Interfaces},
    pages = {402–412},
    numpages = {11},
    location = {College Station, TX, USA},
    series = {IUI '21}
}

@inproceedings{Tom2020,
author = {Brown, Tom B. and Mann, Benjamin and Ryder, Nick and Subbiah, Melanie and Kaplan, Jared and Dhariwal, Prafulla and Neelakantan, Arvind and Shyam, Pranav and Sastry, Girish and Askell, Amanda and Agarwal, Sandhini and Herbert-Voss, Ariel and Krueger, Gretchen and Henighan, Tom and Child, Rewon and Ramesh, Aditya and Ziegler, Daniel M. and Wu, Jeffrey and Winter, Clemens and Hesse, Christopher and Chen, Mark and Sigler, Eric and Litwin, Mateusz and Gray, Scott and Chess, Benjamin and Clark, Jack and Berner, Christopher and McCandlish, Sam and Radford, Alec and Sutskever, Ilya and Amodei, Dario},
title = {Language Models Are Few-Shot Learners},
year = {2020},
isbn = {9781713829546},
publisher = {Curran Associates Inc.},
address = {Red Hook, NY, USA},
booktitle = {Proceedings of the 34th International Conference on Neural Information Processing Systems},
articleno = {159},
numpages = {25},
location = {Vancouver, BC, Canada},
series = {NIPS'20}
}

@article{Stahl2021,
    author ={B.C. Stahl and A. Andreou and P. Brey and T. Hatzakis and A. Kirichenko and K. Macnish and S. Laulhé Shaelou and A. Patel and M. Ryan and D. Wright} ,
    title = {Artificial intelligence for human flourishing – Beyond principles for machine learning},
    journal = {Journal of Business Research},
    year = {2021},
    doi = {https://doi.org/10.1016/j.jbusres.2020.11.030}
}

@article{Flathmann2021,
    author ={Christopher Flathmann and Beau Schelble and Rui Zhang and Nathan McNeese},
    title = {Modeling and Guiding the Creation of Ethical Human-AI Teams},
    journal = { In Proceedings of the 2021 AAAI/ACM Conference on AI, Ethics, and Society },
    year = {2021},
    doi = {10.1145/3461702.3462573}
}

@article{Zhao2022,
author = {Zhao, Michelle and Simmons, Reid and Admoni, Henny},
title = {The Role of Adaptation in Collective Human–AI Teaming},
journal = {Topics in Cognitive Science},
volume = {17},
number = {2},
pages = {291-323},
doi = {https://doi.org/10.1111/tops.12633},
url = {https://onlinelibrary.wiley.com/doi/abs/10.1111/tops.12633},
eprint = {https://onlinelibrary.wiley.com/doi/pdf/10.1111/tops.12633},
year = {2025}
}

@article{Gopinath2022,
    author ={Gopinath, D. and DeCastro, J. and Rosman, G. and Sumner, E. and Morgan, A. and Hakimi, S. and Stent, S. },
    title = {HMIway-env: A Framework for Simulating Behaviors and Preferences to Support Human-{AI} Teaming in Driving},
    journal = { In Proceedings of the IEEE/CVF Conference on Computer Vision and Pattern Recognition },
    year = {2022},
    doi = {https://openaccess.thecvf.com/content/CVPR2022W/HCIS/papers/Gopinath_HMIway-Env_A_Framework_for_Simulating_Behaviors_and_Preferences_To_Support_CVPRW_2022_paper.pdf}
}

@article{Munyaka2023,
    author ={Munyaka, I. and Ashktorab, Z. and Dugan, C. and Johnson, J. and Pan, Q.  } ,
    title = {Decision Making Strategies and Team Efficacy in Human-{AI} Teams},
    journal = {Proceedings of the ACM on Human-Computer Interaction, },
    year = {2023},
    doi = {10.1145/3579476}
}

@misc{wang2023,
  title={The Large Language Model ({LLM}) Paradox: Job Creation and Loss in the Age of Advanced {AI}},
  author={Wang, Yifei},
  year={2023},
  howpublished={TechRxiv Preprint},
}

@inproceedings{hu2023,
  title={Language instructed reinforcement learning for human-{AI} coordination},
  author={Hu, Hengyuan and Sadigh, Dorsa},
  booktitle={International Conference on Machine Learning},
  pages={13584--13598},
  year={2023},
  organization={PMLR}
}

@article{Budd2021,
	doi = {10.1016/j.media.2021.102062},
	url = {https://doi.org/10.1016%2Fj.media.2021.102062},
	year = 2021,
	month = {jul},
	publisher = {Elsevier {BV}},
	volume = {71},
	pages = {102062},  
	author = {Samuel Budd and Emma C. Robinson and Bernhard Kainz},  
	title = {A survey on active learning and human-in-the-loop deep learning for medical image analysis},  
	journal = {Medical Image Analysis}
}

@article{pal2023,
  title={The Future of Large Language Models: A Futuristic Dissection on {AI} and Human Interaction},
  author={Pal, Subharun},
  journal={IJFMR-International Journal For Multidisciplinary Research},
  volume={5},
  number={3},
  publisher={IJFMR},
  year = {2023}
}

@article{Tabrez2020,
author = {Tabrez, Aaquib and Luebbers, Matthew and Hayes, Bradley},
year = {2020},
month = {12},
pages = {},
title = {A Survey of Mental Modeling Techniques in Human–Robot Teaming},
volume = {1},
journal = {Current Robotics Reports},
doi = {10.1007/s43154-020-00019-0}
}

@article{Nicholas2023,
author = {Conlon, Nicholas and Ahmed, Nisar and Szafir, Daniel},
year = {2023},
month = {08},
pages = {},
title = {A Survey of Algorithmic Methods for Competency Self-Assessments in Human-Autonomy Teaming},
journal = {ACM Computing Surveys},
doi = {10.1145/3616010}
}

@inproceedings{Whittlestone2019,
  title={The role and limits of principles in AI ethics: Towards a focus on tensions},
  author={Whittlestone, Jess and Nyrup, Rune and Alexandrova, Anna and Cave, Stephen},
  booktitle={Proceedings of the 2019 AAAI/ACM Conference on AI, Ethics, and Society},
  pages={195--200},
  year={2019}
}

@article{Morrison2023,
author = {Morrison, Katelyn and Shin, Donghoon and Holstein, Kenneth and Perer, Adam},
title = {Evaluating the Impact of Human Explanation Strategies on Human-{AI} Visual Decision-Making},
year = {2023},
issue_date = {April 2023},
publisher = {Association for Computing Machinery},
address = {New York, NY, USA},
volume = {7},
number = {CSCW1},
url = {https://doi.org/10.1145/3579481},
doi = {10.1145/3579481},
journal = {Proc. ACM Hum.-Comput. Interact.},
month = {apr},
articleno = {48},
numpages = {37}
}

@inproceedings{Memmert2023,
 title={TOWARDS HUMAN-{AI}-COLLABORATION IN BRAINSTORMING: EMPIRICAL INSIGHTS INTO THE PERCEPTION OF WORKING WITH A GENERATIVE AI},
 author={Memmert, Lucas and Tavanapour, Navid},
year={2023},
booktitle = {31st European Conference on Information Systems (ECIS)}
}

@inproceedings{Jianlong2019,
    title={Towards trustworthy human-{AI} teaming under uncertainty},
  author={Zhou, Jianlong and Chen, Fang},
  booktitle={IJCAI 2019 workshop on explainable AI (XAI)},
  year={2019}
}

@article{Bienefeld2023,
  title={Human-{AI} teaming: leveraging transactive memory and speaking up for enhanced team effectiveness},
  author={Bienefeld, Nadine and Kolbe, Michaela and Camen, Giovanni and Huser, Dominic and Buehler, Philipp Karl},
  journal={Frontiers in Psychology},
  volume={14},
  year={2023},
  publisher={Frontiers Media SA}
}

@inproceedings{kim2023help,
  title={"Help Me Help the {AI}": Understanding How Explainability Can Support Human-AI Interaction},
  author={Kim, Sunnie SY and Watkins, Elizabeth Anne and Russakovsky, Olga and Fong, Ruth and Monroy-Hern{\'a}ndez, Andr{\'e}s},
  booktitle={Proceedings of the 2023 CHI Conference on Human Factors in Computing Systems},
  pages={1--17},
  year={2023}
}

@inproceedings{ahmad2023towards,
  title={Towards human-bot collaborative software architecting with {C}hat{GPT}},
  author={Ahmad, Aakash and Waseem, Muhammad and Liang, Peng and Fahmideh, Mahdi and Aktar, Mst Shamima and Mikkonen, Tommi},
  booktitle={Proceedings of the 27th International Conference on Evaluation and Assessment in Software Engineering},
  pages={279--285},
  year={2023}
}

@article{wang2019human,
  title={Human-{AI} collaboration in data science: Exploring data scientists' perceptions of automated {AI}},
  author={Wang, Dakuo and Weisz, Justin D and Muller, Michael and Ram, Parikshit and Geyer, Werner and Dugan, Casey and Tausczik, Yla and Samulowitz, Horst and Gray, Alexander},
  journal={Proceedings of the ACM on human-computer interaction},
  volume={3},
  number={CSCW},
  pages={1--24},
  year={2019},
  publisher={ACM New York, NY, USA}
}

@inproceedings{Memmert2022,
  title={Complex Problem Solving through Human-AI Collaboration: Literature Review on Research Contexts},
  author={Memmert, Lucas and Bittner, Eva},
  year={2022},
booktitle = {Proceedings of the Annual Hawaii International Conference on System Sciences}
}

@inproceedings{Lu2023,
  title={ReadingQuizMaker: A Human-{NLP} Collaborative System that Supports Instructors to Design High-Quality Reading Quiz Questions},
  author={Lu, Xinyi and Fan, Simin and Houghton, Jessica and Wang, Lu and Wang, Xu},
  booktitle={Proceedings of the 2023 CHI Conference on Human Factors in Computing Systems},
  pages={1--18},
  year={2023}
}

@inproceedings{Bosch2019,
  title={Six challenges for human-{AI} Co-learning},
  author={van den Bosch, Karel and Schoonderwoerd, Tjeerd and Blankendaal, Romy and Neerincx, Mark},
  booktitle={Adaptive Instructional Systems: First International Conference, AIS 2019, Held as Part of the 21st HCI International Conference, HCII 2019, Orlando, FL, USA, July 26--31, 2019, Proceedings 21},
  pages={572--589},
  year={2019},
  organization={Springer}
}

@article{Hauptman2023,
  title={Adapt and overcome: Perceptions of adaptive autonomous agents for human-{AI} teaming},
  author={Hauptman, Allyson I and Schelble, Beau G and McNeese, Nathan J and Madathil, Kapil Chalil},
  journal={Computers in Human Behavior},
  volume={138},
  pages={107451},
  year={2023},
  publisher={Elsevier}
}

@article{Fuchs2023,
  title={Compensating for Sensing Failures via Delegation in Human-{AI} Hybrid Systems},
  author={Fuchs, Andrew and Passarella, Andrea and Conti, Marco},
  journal={Sensors},
  volume={23},
  number={7},
  pages={3409},
  year={2023},
  publisher={MDPI}
}

@article{Lv2021,
author = {Lv, Chen and Li, Yutong and Xing, Yang and Huang, Chao and Cao, Dongpu and Zhao, Yifan and Liu, Yahui},
year = {2021},
month = {02},
pages = {2000229},
title = {Human-Machine Collaboration for Automated Driving Using an Intelligent Two‐Phase Haptic Interface},
journal = {Advanced Intelligent Systems},
doi = {10.1002/aisy.202000229}
}

@article{McKinney2020,
    title={International evaluation of an {AI} system for breast cancer screening},
  author={McKinney, Scott Mayer and Sieniek, Marcin and Godbole, Varun and Godwin, Jonathan and Antropova, Natasha and Ashrafian, Hutan and Back, Trevor and Chesus, Mary and Corrado, Greg S and Darzi, Ara and others},
  journal={Nature},
  volume={577},
  number={7788},
  pages={89--94},
  year={2020},
  publisher={Nature Publishing Group UK London}}

@article{chang2023,
author = {Chang, Yupeng and Wang, Xu and Wang, Jindong and Wu, Yuan and Yang, Linyi and Zhu, Kaijie and Chen, Hao and Yi, Xiaoyuan and Wang, Cunxiang and Wang, Yidong and Ye, Wei and Zhang, Yue and Chang, Yi and Yu, Philip S. and Yang, Qiang and Xie, Xing},
title = {A Survey on Evaluation of Large Language Models},
year = {2024},
issue_date = {June 2024},
publisher = {Association for Computing Machinery},
address = {New York, NY, USA},
volume = {15},
number = {3},
issn = {2157-6904},
url = {https://doi.org/10.1145/3641289},
doi = {10.1145/3641289},
journal = {ACM Trans. Intell. Syst. Technol.},
month = mar,
articleno = {39},
numpages = {45}
}

@inproceedings{Carrie2019,
     title={Human-centered tools for coping with imperfect algorithms during medical decision-making},
  author={Cai, Carrie J and Reif, Emily and Hegde, Narayan and Hipp, Jason and Kim, Been and Smilkov, Daniel and Wattenberg, Martin and Viegas, Fernanda and Corrado, Greg S and Stumpe, Martin C and others},
  booktitle={Proceedings of the 2019 chi conference on human factors in computing systems},
  pages={1--14},
  year={2019}
}

@article{Ben2019,
author = {Green, Ben and Chen, Yiling},
title = {The Principles and Limits of Algorithm-in-the-Loop Decision Making},
year = {2019},
issue_date = {November 2019},
publisher = {Association for Computing Machinery},
address = {New York, NY, USA},
volume = {3},
number = {CSCW},
url = {https://doi.org/10.1145/3359152},
doi = {10.1145/3359152},
journal = {Proc. ACM Hum.-Comput. Interact.},
month = {nov},
articleno = {50},
numpages = {24}
}

@article{ouyang2022,
  title={Training language models to follow instructions with human feedback},
  author={Ouyang, Long and Wu, Jeffrey and Jiang, Xu and Almeida, Diogo and Wainwright, Carroll and Mishkin, Pamela and Zhang, Chong and Agarwal, Sandhini and Slama, Katarina and Ray, Alex and others},
  journal={Advances in Neural Information Processing Systems},
  volume={35},
  pages={27730--27744},
  year={2022}
}

@inproceedings{Christiano2017,
  author       = {Paul F. Christiano and
                  Jan Leike and
                  Tom B. Brown and
                  Miljan Martic and
                  Shane Legg and
                  Dario Amodei},
  editor       = {Isabelle Guyon and
                  Ulrike von Luxburg and
                  Samy Bengio and
                  Hanna M. Wallach and
                  Rob Fergus and
                  S. V. N. Vishwanathan and
                  Roman Garnett},
  title        = {Deep Reinforcement Learning from Human Preferences},
  booktitle    = {Advances in Neural Information Processing Systems 30: Annual Conference
                  on Neural Information Processing Systems 2017, December 4-9, 2017,
                  Long Beach, CA, {USA}},
  pages        = {4299--4307},
  year         = {2017},
  url          = {https://proceedings.neurips.cc/paper/2017/hash/d5e2c0adad503c91f91df240d0cd4e49-Abstract.html},
  bibsource    = {dblp computer science bibliography, https://dblp.org}
}

@article{nakano2021,
  title={Webgpt: Browser-assisted question-answering with human feedback},
  author={Nakano, Reiichiro and Hilton, Jacob and Balaji, Suchir and Wu, Jeff and Ouyang, Long and Kim, Christina and Hesse, Christopher and Jain, Shantanu and Kosaraju, Vineet and Saunders, William and others},
  journal={arXiv preprint arXiv:2112.09332},
  year={2021}
}

@inproceedings{rafailov2023,
author = {Rafailov, Rafael and Sharma, Archit and Mitchell, Eric and Ermon, Stefano and Manning, Christopher D. and Finn, Chelsea},
title = {Direct preference optimization: your language model is secretly a reward model},
year = {2023},
publisher = {Curran Associates Inc.},
address = {Red Hook, NY, USA},
booktitle = {Proceedings of the 37th International Conference on Neural Information Processing Systems},
articleno = {2338},
numpages = {14},
location = {New Orleans, LA, USA},
series = {NIPS '23}
}

@article{kumar2022,
  title={A Deep Learning Based Model to Assist Blind People in Their Navigation},
  author={Kumar, Nitin and Jain, Anuj},
  journal={Journal of Information Technology Education: Innovations in Practice},
  volume={21},
  pages={095--114},
  year={2022},
  publisher={Informing Science Institute. 131 Brookhill Court, Santa Rosa, CA 95409}
}

@article{kaelbling1996reinforcement,
  title={Reinforcement learning: A survey},
  author={Kaelbling, Leslie Pack and Littman, Michael L and Moore, Andrew W},
  journal={Journal of artificial intelligence research},
  volume={4},
  pages={237--285},
  year={1996}
}

@ARTICLE{McNeese2021,
  author={McNeese, Nathan J. and Schelble, Beau G. and Canonico, Lorenzo Barberis and Demir, Mustafa},
  journal={IEEE Transactions on Human-Machine Systems}, 
  title={Who/What Is My Teammate? Team Composition Considerations in Human–{AI} Teaming}, 
  year={2021},
  volume={51},
  number={4},
  pages={288-299},
  doi={10.1109/THMS.2021.3086018}}

@inproceedings{Hemmer2023,
author = {Hemmer, Patrick and Westphal, Monika and Schemmer, Max and Vetter, Sebastian and V\"{o}ssing, Michael and Satzger, Gerhard},
title = {Human-{AI} Collaboration: The Effect of {AI} Delegation on Human Task Performance and Task Satisfaction},
year = {2023},
isbn = {9798400701061},
publisher = {Association for Computing Machinery},
address = {New York, NY, USA},
url = {https://doi.org/10.1145/3581641.3584052},
doi = {10.1145/3581641.3584052},
booktitle = {Proceedings of the 28th International Conference on Intelligent User Interfaces},
pages = {453–463},
numpages = {11},
location = {Sydney, NSW, Australia},
series = {IUI '23}
}

@article{CHOWDHURY202231,
title = {{AI}-employee collaboration and business performance: Integrating knowledge-based view, socio-technical systems and organisational socialisation framework},
journal = {Journal of Business Research},
volume = {144},
pages = {31-49},
year = {2022},
issn = {0148-2963},
doi = {https://doi.org/10.1016/j.jbusres.2022.01.069},
url = {https://www.sciencedirect.com/science/article/pii/S0148296322000819},
author = {Soumyadeb Chowdhury and Pawan Budhwar and Prasanta Kumar Dey and Sian Joel-Edgar and Amelie Abadie}

}

@article{CHOUDHURY2022,
title = {Impact of accountability, training, and human factors on the use of artificial intelligence in healthcare: Exploring the perceptions of healthcare practitioners in the {US}},
journal = {Human Factors in Healthcare},
volume = {2},
pages = {100021},
year = {2022},
issn = {2772-5014},
doi = {https://doi.org/10.1016/j.hfh.2022.100021},
url = {https://www.sciencedirect.com/science/article/pii/S2772501422000185},
author = {Avishek Choudhury and Onur Asan}
}

@article{Veselovsky2023,
author = {Veselovsky, Veniamin and Horta Ribeiro, Manoel and Cozzolino, Philip J. and Gordon, Andrew and Rothschild, David and West, Robert},
title = {Prevalence and Prevention of Large Language Model Use in Crowd Work},
year = {2025},
issue_date = {March 2025},
publisher = {Association for Computing Machinery},
address = {New York, NY, USA},
volume = {68},
number = {3},
issn = {0001-0782},
url = {https://doi.org/10.1145/3685527},
doi = {10.1145/3685527},
journal = {Commun. ACM},
month = feb,
pages = {42–47},
numpages = {6}
}

@article{ziegler2019,
  title={Fine-tuning language models from human preferences},
  author={Ziegler, Daniel M and Stiennon, Nisan and Wu, Jeffrey and Brown, Tom B and Radford, Alec and Amodei, Dario and Christiano, Paul and Irving, Geoffrey},
  journal={arXiv preprint arXiv:1909.08593},
  year={2019}
}

@article{nwana1990,
  title={Intelligent tutoring systems: an overview},
  author={Nwana, Hyacinth S},
  journal={Artificial Intelligence Review},
  volume={4},
  number={4},
  pages={251--277},
  year={1990},
  publisher={Springer}
}

@article{vanlehn2011,
  title={The relative effectiveness of human tutoring, intelligent tutoring systems, and other tutoring systems},
  author={VanLehn, Kurt},
  journal={Educational psychologist},
  volume={46},
  number={4},
  pages={197--221},
  year={2011},
  publisher={Taylor \& Francis}
}

@inproceedings{whitaker2013,
  title={The effectiveness of intelligent tutoring on training in a video game},
  author={Whitaker, Elizabeth and Trewhitt, Ethan and Holtsinger, Matthew and Hale, Christopher and Veinott, Elizabeth and Argenta, Chris and Catrambone, Richard},
  booktitle={2013 IEEE International Games Innovation Conference (IGIC)},
  pages={267--274},
  year={2013},
  organization={IEEE}
}

@inproceedings{kasinathan2017,
  title={Adaptive learning system for higher learning},
  author={Kasinathan, Vinothini and Mustapha, Aida and Medi, Imran},
  booktitle={2017 8th international conference on information technology (ICIT)},
  pages={960--970},
  year={2017},
  organization={IEEE}
}

@inproceedings{eicher2018,
  title={Jill Watson doesn't care if you're pregnant: Grounding {AI} ethics in empirical studies},
  author={Eicher, Bobbie and Polepeddi, Lalith and Goel, Ashok},
  booktitle={Proceedings of the 2018 AAAI/ACM Conference on AI, Ethics, and Society},
  pages={88--94},
  year={2018}
}

@article{hartle2019,
  title={The positive aspects of Mursion when teaching higher education students},
  author={Hartle, Luminita and Kaczorowski, Tara},
  journal={Quarterly Review of Distance Education},
  volume={20},
  number={4},
  pages={71--100},
  year={2019},
  publisher={Information Age Publishing}
}

@article{dikli2006,
  title={Automated essay scoring},
  author={Dikli, Semire},
  journal={Turkish Online Journal of Distance Education},
  volume={7},
  number={1},
  pages={49--62},
  year={2006},
  publisher={Anadolu University}
}

@inproceedings{Ozarkar2020,
  author={Ozarkar, Saket and Chetwani, Raj and Devare, Sugam and Haryani, Sumeet and Giri, Nupur},
  booktitle={2020 11th International Conference on Computing, Communication and Networking Technologies (ICCCNT)}, 
  title={{AI} for Accessibility: Virtual Assistant for Hearing Impaired}, 
  year={2020},
  pages={1-7},
  doi={10.1109/ICCCNT49239.2020.9225392}
}

@article{Lertvittayakumjorn2020,
  title={Human-in-the-loop Debugging Deep Text Classifiers},
  author={Piyawat Lertvittayakumjorn and Lucia Specia and Francesca Toni},
  journal={arXiv preprint arXiv:2010.04987},
  year={2020},
  volume={abs/2010.04987},
  url={https://api.semanticscholar.org/CorpusID:222290812}
}

@article{Pries2023,
  title={Active pairwise distance learning for efficient labeling of large datasets by human experts},
  author={Joris Pries and Sandjai Bhulai and Rob D. van der Mei},
  journal={Applied Intelligence},
  year={2023},
  volume={53},
  pages={24689 - 24708},
  url={https://api.semanticscholar.org/CorpusID:260307431}
}

@inproceedings{Yao2023,
    title = "Beyond Labels: Empowering Human Annotators with Natural Language Explanations through a Novel Active-Learning Architecture",
    author = "Yao, Bingsheng  and
      Jindal, Ishan  and
      Popa, Lucian  and
      Katsis, Yannis  and
      Ghosh, Sayan  and
      He, Lihong  and
      Lu, Yuxuan  and
      Srivastava, Shashank  and
      Li, Yunyao  and
      Hendler, James  and
      Wang, Dakuo",
    editor = "Bouamor, Houda  and
      Pino, Juan  and
      Bali, Kalika",
    booktitle = "Findings of the Association for Computational Linguistics: EMNLP 2023",
    month = dec,
    year = "2023",
    address = "Singapore",
    publisher = "Association for Computational Linguistics",
    url = "https://aclanthology.org/2023.findings-emnlp.778",
    doi = "10.18653/v1/2023.findings-emnlp.778",
    pages = "11629--11643",
}

@inproceedings{Lu2022,
    title = "A Rationale-Centric Framework for Human-in-the-loop Machine Learning",
    author = "Lu, Jinghui  and
      Yang, Linyi  and
      Namee, Brian  and
      Zhang, Yue",
    editor = "Muresan, Smaranda  and
      Nakov, Preslav  and
      Villavicencio, Aline",
    booktitle = "Proceedings of the 60th Annual Meeting of the Association for Computational Linguistics (Volume 1: Long Papers)",
    month = may,
    year = "2022",
    address = "Dublin, Ireland",
    publisher = "Association for Computational Linguistics",
    url = "https://aclanthology.org/2022.acl-long.481/",
    doi = "10.18653/v1/2022.acl-long.481",
    pages = "6986--6996",
}

@inproceedings{Bao2018,
    title = "Deriving Machine Attention from Human Rationales",
    author = "Bao, Yujia  and
      Chang, Shiyu  and
      Yu, Mo  and
      Barzilay, Regina",
    editor = "Riloff, Ellen  and
      Chiang, David  and
      Hockenmaier, Julia  and
      Tsujii, Jun{'}ichi",
    booktitle = "Proceedings of the 2018 Conference on Empirical Methods in Natural Language Processing",
    month = oct # "-" # nov,
    year = "2018",
    address = "Brussels, Belgium",
    publisher = "Association for Computational Linguistics",
    url = "https://aclanthology.org/D18-1216",
    doi = "10.18653/v1/D18-1216",
    pages = "1903--1913",
}

@book{Settles2009,
    author = {Burr Settles},
    title = {Active Learning Literature Survey},
    publisher = {Department of Computer Sciences, University of Wisconsin-Madison},
    year = {2009}
}

@inproceedings{Rastogi2023,
  title={Supporting human-{AI} collaboration in auditing {LLM} with {LLMs}},
  author={Rastogi, Charvi and Tulio Ribeiro, Marco and King, Nicholas and Nori, Harsha and Amershi, Saleema},
  booktitle={Proceedings of the 2023 AAAI/ACM Conference on AI, Ethics, and Society},
  pages={913--926},
  year={2023}
}

@article{Sharma2023,
  title={Human-{AI} collaboration enables more empathic conversations in text-based peer-to-peer mental health support},
  author={Sharma, Ashish and Lin, Inna W and Miner, Adam S and Atkins, David C and Althoff, Tim},
  journal={Nature Machine Intelligence},
  volume={5},
  number={1},
  pages={46--57},
  year={2023},
  publisher={Nature Publishing Group UK London}
}

@inproceedings{Chakrabarty2023,
    title = "{I} Spy a Metaphor: Large Language Models and Diffusion Models Co-Create Visual Metaphors",
    author = "Chakrabarty, Tuhin  and
      Saakyan, Arkadiy  and
      Winn, Olivia  and
      Panagopoulou, Artemis  and
      Yang, Yue  and
      Apidianaki, Marianna  and
      Muresan, Smaranda",
    editor = "Rogers, Anna  and
      Boyd-Graber, Jordan  and
      Okazaki, Naoaki",
    booktitle = "Findings of the Association for Computational Linguistics: ACL 2023",
    month = jul,
    year = "2023",
    address = "Toronto, Canada",
    publisher = "Association for Computational Linguistics",
    url = "https://aclanthology.org/2023.findings-acl.465/",
    doi = "10.18653/v1/2023.findings-acl.465",
    pages = "7370--7388",
}

@article{Kozlowski2006,
    title = {Enhancing the effectiveness of work groups and teams. Psychological science in the public interest},
    author = {Kozlowski, Steve WJ and Ilgen, Daniel R} ,
    journal = {Psychological Science in the Public Interest},
    year = {2006}
}

@article{Salas2017,
    title = {Situation awareness in team performance: Implications for measurement and training},
    author = {Salas, Eduardo and Prince, Carolyn and Baker, David P and Shrestha, Lisa},
    journal = {Human Factors},
    volume={37},
    year = {1995},
    number={1}, 
    pages={123--136}, 
    doi={10.1518/001872095779049525}
}

@book{Russell2021, 
    title = {Artificial Intelligence: A Modern Approach}, 
    author = {Russell, Stuart J. and Norvig, Peter}, 
    year = {2021},
publisher = {Prentice Hall}
}

@article{Eloundou2023,
      title={{GPTs} are {GPTs}: An Early Look at the Labor Market Impact Potential of Large Language Models}, 
      author={Tyna Eloundou and Sam Manning and Pamela Mishkin and Daniel Rock},
      year={2023},
      journal={arXiv preprint arXiv:2303.10130},
      archivePrefix={arXiv},
      primaryClass={econ.GN}
}

@article{Khan2020,
  author={Khan, Muiz Ahmed and Paul, Pias and Rashid, Mahmudur and Hossain, Mainul and Ahad, Md Atiqur Rahman},
  journal={IEEE Transactions on Human-Machine Systems}, 
  title={An {AI}-Based Visual Aid With Integrated Reading Assistant for the Completely Blind}, 
  year={2020},
  volume={50},
  number={6},
  pages={507-517},
  doi={10.1109/THMS.2020.3027534}}

@article{Wen2021,
author = {Wen, Feng and Zhang, Zixuan and He, Tianyiyi and Lee, Chengkuo},
year = {2021},
month = {09},
pages = {5378},
title = {{AI} enabled sign language recognition and VR space bidirectional communication using triboelectric smart glove},
volume = {12},
journal = {Nature Communications},
doi = {10.1038/s41467-021-25637-w}
}

@article{Ghazal2021, 
title={{AI}-Powered Service Robotics for Independent Shopping Experiences by Elderly and Disabled People},
volume={11}, 
ISSN={2076-3417}, 
url={http://dx.doi.org/10.3390/app11199007}, 
DOI={10.3390/app11199007}, 
number={19}, 
journal={Applied Sciences}, 
publisher={MDPI AG}, 
author={Ghazal, Mohammed and Yaghi, Maha and Gad, Abdalla and El Bary, Gasm and Alhalabi, Marah and Alkhedher, Mohammad and El-Baz, Ayman S.}, 
year={2021}, 
month=sep, 
pages={9007} 
}

@article{harrer2023,
  title={Attention is not all you need: the complicated case of ethically using large language models in healthcare and medicine},
  author={Harrer, Stefan},
  journal={EBioMedicine},
  volume={90},
  year={2023},
  publisher={Elsevier}
}

@article{Bau2019,
author = {Bau, David and Strobelt, Hendrik and Peebles, William and Wulff, Jonas and Zhou, Bolei and Zhu, Jun-Yan and Torralba, Antonio},
title = {Semantic Photo Manipulation with a Generative Image Prior},
year = {2019},
issue_date = {August 2019},
publisher = {Association for Computing Machinery},
address = {New York, NY, USA},
volume = {38},
number = {4},
issn = {0730-0301},
url = {https://doi.org/10.1145/3306346.3323023},
doi = {10.1145/3306346.3323023},
journal = {ACM Trans. Graph.},
month = {jul},
articleno = {59},
numpages = {11}
}

@inproceedings{Qing2023,
author = {Zheng, Qingxiao and Tang, Yiliu and Liu, Yiren and Liu, Weizi and Huang, Yun},
title = {UX Research on Conversational Human-AI Interaction: A Literature Review of the ACM Digital Library},
year = {2022},
isbn = {9781450391573},
publisher = {Association for Computing Machinery},
address = {New York, NY, USA},
url = {https://doi.org/10.1145/3491102.3501855},
doi = {10.1145/3491102.3501855},
booktitle = {Proceedings of the 2022 CHI Conference on Human Factors in Computing Systems},
articleno = {570},
numpages = {24},
location = {New Orleans, LA, USA},
series = {CHI '22}
}

@inproceedings{Bryan2023,
author = {Wang, Bryan and Li, Gang and Li, Yang},
title = {Enabling Conversational Interaction with Mobile UI using Large Language Models},
year = {2023},
isbn = {9781450394215},
publisher = {Association for Computing Machinery},
address = {New York, NY, USA},
url = {https://doi.org/10.1145/3544548.3580895},
doi = {10.1145/3544548.3580895},
booktitle = {Proceedings of the 2023 CHI Conference on Human Factors in Computing Systems},
articleno = {432},
numpages = {17},
location = {Hamburg, Germany},
series = {CHI '23}
}

@article{Mingming2023,
author = {Fan, Mingming and Yang, Xianyou and Yu, TszTung and Liao, Q. Vera and Zhao, Jian},
title = {Human-AI Collaboration for UX Evaluation: Effects of Explanation and Synchronization},
year = {2022},
issue_date = {April 2022},
publisher = {Association for Computing Machinery},
address = {New York, NY, USA},
volume = {6},
number = {CSCW1},
url = {https://doi.org/10.1145/3512943},
doi = {10.1145/3512943},
journal = {Proc. ACM Hum.-Comput. Interact.},
month = apr,
articleno = {96},
numpages = {32}
}

@article{Clark2016,
author = {Clark, Leigh and Ofemile, Abdulmalik and Adolphs, Svenja and Rodden, Tom},
title = {A Multimodal Approach to Assessing User Experiences with Agent Helpers},
year = {2016},
issue_date = {December 2016},
publisher = {Association for Computing Machinery},
address = {New York, NY, USA},
volume = {6},
number = {4},
issn = {2160-6455},
url = {https://doi.org/10.1145/2983926},
doi = {10.1145/2983926},
journal = {ACM Trans. Interact. Intell. Syst.},
month = nov,
articleno = {29},
numpages = {31}
}

@inproceedings{Tongshuang2022,
author = {Wu, Tongshuang and Terry, Michael and Cai, Carrie Jun},
title = {{AI} Chains: Transparent and Controllable Human-{AI} Interaction by Chaining Large Language Model Prompts},
year = {2022},
isbn = {9781450391573},
publisher = {Association for Computing Machinery},
address = {New York, NY, USA},
url = {https://doi.org/10.1145/3491102.3517582},
doi = {10.1145/3491102.3517582},
booktitle = {Proceedings of the 2022 CHI Conference on Human Factors in Computing Systems},
articleno = {385},
numpages = {22},
location = {New Orleans, LA, USA},
series = {CHI '22}
}

@article{Kaptein2016,
    author = {Kaptein, Frank and Broekens, Joost and Hindriks, Koen V and Neerincx, Mark},
    title = {{CAAF}: A cognitive affective agent programming framework} ,
    journal = {Engineering Psychology and Cognitive Ergonomics } ,
    volume = {10906},
    year = {2016},
    pages = {317–330}
}

@article{Nikolaidis2017,
    author = {Nikolaidis, Stefanos and Hsu, David and Srinivasa, Siddhartha},
    title = {Human-robot mutual adaptation in collaborative tasks: Models and experiments} ,
    journal = {The International Journal of Robotics Research} ,
    volume = {36},
    year = {2017},
    pages = {618-634}
}

@article{Neerincx2018,
    author = {Neerincx, Mark A and Waa, Jasper van der and Kaptein, Frank and Diggelen, Jurrian van},
    title = {Using Perceptual and Cognitive Explanations
for Enhanced Human-Agent Team Performance} ,
    journal = {Lecture Notes in Computer Science},
    volume = {10011},
    year = {2018},
    pages = {204-214}
}

@article{Braun2021, 
    title = {A Leap of Faith: Is There a Formula for “Trustworthy” {AI}?}, 
    author = {Braun, Matthias and Bleher, Hannah and Hummel, Patrik}, 
    journal = {The Hastings Center Report}, 
    volume = {51}, 
    pages = {17–22}, 
    year = {2021}
}

@article{johnson2019no,
  title={No {AI} is an island: the case for teaming intelligence},
  author={Johnson, Matthew and Vera, Alonso},
  journal={AI magazine},
  volume={40},
  number={1},
  pages={16--28},
  year={2019}
}

@article{Atakishiyev2021,
author = {Atakishiyev, Shahin and Salameh, Mohammad and Yao, Hengshuai and Goebel, Randy},
year = {2021},
month = {12},
pages = {},
title = {Explainable Artificial Intelligence for Autonomous Driving: A Comprehensive Overview and Field Guide for Future Research Directions},
journal = {arXiv preprint arXiv:2112.11561}
}

@ARTICLE{Pazho2023,
  author={Danesh Pazho, Armin and Neff, Christopher and Noghre, Ghazal Alinezhad and Ardabili, Babak Rahimi and Yao, Shanle and Baharani, Mohammadreza and Tabkhi, Hamed},
  journal={IEEE Internet of Things Journal}, 
  title={Ancilia: Scalable Intelligent Video Surveillance for the Artificial Intelligence of Things}, 
  year={2023},
  volume={10},
  number={17},
  pages={14940-14951},
  doi={10.1109/JIOT.2023.3263725}}

@InProceedings{Mirchandani2023,
  title = 	 {Large Language Models as General Pattern Machines},
  author =       {Mirchandani, Suvir and Xia, Fei and Florence, Pete and Ichter, Brian and Driess, Danny and Arenas, Montserrat Gonzalez and Rao, Kanishka and Sadigh, Dorsa and Zeng, Andy},
  booktitle = 	 {Proceedings of The 7th Conference on Robot Learning},
  pages = 	 {2498--2518},
  year = 	 {2023},
  editor = 	 {Tan, Jie and Toussaint, Marc and Darvish, Kourosh},
  volume = 	 {229},
  series = 	 {Proceedings of Machine Learning Research},
  month = 	 {06--09 Nov},
  publisher =    {PMLR},
  url = 	 {https://proceedings.mlr.press/v229/mirchandani23a.html},
}

@article{Wei2022,
  title={Chain-of-thought prompting elicits reasoning in large language models},
  author={Wei, Jason and Wang, Xuezhi and Schuurmans, Dale and Bosma, Maarten and Xia, Fei and Chi, Ed and Le, Quoc V and Zhou, Denny and others},
  journal={Advances in neural information processing systems},
  volume={35},
  pages={24824--24837},
  year={2022}
}

@inproceedings{Zha2023,
  title={Distilling and retrieving generalizable knowledge for robot manipulation via language corrections},
  author={Zha, Lihan and Cui, Yuchen and Lin, Li-Heng and Kwon, Minae and Arenas, Montserrat Gonzalez and Zeng, Andy and Xia, Fei and Sadigh, Dorsa},
  booktitle={2024 IEEE international conference on robotics and automation (ICRA)},
  pages={15172--15179},
  year={2024},
  organization={IEEE}
}

@inproceedings{clark2021,
    title = "All That{'}s `Human' Is Not Gold: Evaluating Human Evaluation of Generated Text",
    author = "Clark, Elizabeth  and
      August, Tal  and
      Serrano, Sofia  and
      Haduong, Nikita  and
      Gururangan, Suchin  and
      Smith, Noah A.",
    editor = "Zong, Chengqing  and
      Xia, Fei  and
      Li, Wenjie  and
      Navigli, Roberto",
    booktitle = "Proceedings of the 59th Annual Meeting of the Association for Computational Linguistics and the 11th International Joint Conference on Natural Language Processing (Volume 1: Long Papers)",
    month = aug,
    year = "2021",
    address = "Online",
    publisher = "Association for Computational Linguistics",
    url = "https://aclanthology.org/2021.acl-long.565/",
    doi = "10.18653/v1/2021.acl-long.565",
    pages = "7282--7296",
}

@inproceedings{uchendu2023,
  title={Does Human Collaboration Enhance the Accuracy of Identifying {LLM}-Generated Deepfake Texts?},
  author={Uchendu, Adaku and Lee, Jooyoung and Shen, Hua and Le, Thai and Lee, Dongwon and others},
  booktitle={Proceedings of the AAAI Conference on Human Computation and Crowdsourcing},
  pages={163--174},
  year={2023}
}

@inproceedings{lee2023rlaif,
author = {Lee, Harrison and Phatale, Samrat and Mansoor, Hassan and Mesnard, Thomas and Ferret, Johan and Lu, Kellie and Bishop, Colton and Hall, Ethan and Carbune, Victor and Rastogi, Abhinav and Prakash, Sushant},
title = {RLAIF vs. RLHF: scaling reinforcement learning from human feedback with AI feedback},
year = {2024},
publisher = {JMLR.org},
booktitle = {Proceedings of the 41st International Conference on Machine Learning},
articleno = {1071},
numpages = {28},
location = {Vienna, Austria},
series = {ICML'24}
}

@book{mech_turk,
	author = {William Clark and Jan Golinski and Simon Schaffer},
	editor = {},
	publisher = {University of Chicago Press},
	title = {The Sciences in Enlightened Europe},
	year = {1999}
}

@article{cath2018,
  title={Artificial intelligence and the ‘good society’: the {US}, {EU}, and {UK} approach},
  author={Cath, Corinne and Wachter, Sandra and Mittelstadt, Brent and Taddeo, Mariarosaria and Floridi, Luciano},
  journal={Science and engineering ethics},
  volume={24},
  pages={505--528},
  year={2018},
  publisher={Springer}
}

@inproceedings{Bender2021,
  author    = {Bender, Emily M. and Gebru, Timnit and
               McMillan-Major, Angelina and Shmitchell, Shmargaret},
  title     = {On the Dangers of Stochastic Parrots: Can Language Models Be Too Big?},
  booktitle = {Proceedings of the 2021 ACM Conference on Fairness,
               Accountability, and Transparency (FAccT '21)},
  year      = {2021},
  publisher = {Association for Computing Machinery},
  address   = {New York, NY, USA},
  pages     = {610--623},
  doi       = {10.1145/3442188.3445922}
}

@article{Chinonso2023,
  title={The impact of chat{GPT} on privacy and data protection laws},
  author={ThankGod Chinonso, Ekenobi},
  journal={Available at SSRN: https://ssrn.com/abstract=4574016},
  year={2023}
}

@article{walkowiak2023,
  title={Generative AI and the Workforce: What Are the Risks?},
  author={Walkowiak, Emmanuelle and MacDonald, Trent},
  journal={Available at SSRN},
  year={2023}
}

@article{despotovic2024,
  title={Does Chat{GPT} Alter Job Seekers’ Identity? An Experimental Study},
  author={Despotovic, Petar and Bogodistov, Yevgen},
  year={2024},
journal = {Proceedings of the 57th Hawaii International Conference on System Sciences}
}

@article{kirk2021,
  title={Bias out-of-the-box: An empirical analysis of intersectional occupational biases in popular generative language models},
  author={Kirk, Hannah Rose and Jun, Yennie and Volpin, Filippo and Iqbal, Haider and Benussi, Elias and Dreyer, Frederic and Shtedritski, Aleksandar and Asano, Yuki},
  journal={Advances in neural information processing systems},
  volume={34},
  pages={2611--2624},
  year={2021}
}

@article{jobin2019,
  title={The global landscape of {AI} ethics guidelines},
  author={Jobin, Anna and Ienca, Marcello and Vayena, Effy},
  journal={Nature machine intelligence},
  volume={1},
  number={9},
  pages={389--399},
  year={2019},
  publisher={Nature Publishing Group UK London}
}

@techreport{killcrece_2003,
author={Killcrece, Georgia and Kossakowski, Klaus-Peter and Ruefle, Robin and Zajicek, Mark},
title={State of the Practice of Computer Security Incident Response Teams ({CSIRTs})},
month={Oct},
year={2003},
number={CMU/SEI-2003-TR-001},
institution={Carnegie Mellon University, Software Engineering Institute's Digital Library},
url={https://doi.org/10.1184/R1/6584396.v1},
note={Accessed: 2024-Feb-13}
}

@article{Guo2018,
author = {Guo, Anhong and Jain, Anuraag and Ghose, Shomiron and Laput, Gierad and Harrison, Chris and Bigham, Jeffrey P.},
title = {Crowd-{AI} Camera Sensing in the Real World},
year = {2018},
issue_date = {September 2018},
publisher = {Association for Computing Machinery},
address = {New York, NY, USA},
volume = {2},
number = {3},
url = {https://doi.org/10.1145/3264921},
doi = {10.1145/3264921},
journal = {Proc. ACM Interact. Mob. Wearable Ubiquitous Technol.},
month = {sep},
articleno = {111},
numpages = {20}
}

@InProceedings{yang2024,
    author    = {Yang, Yi and Zhang, Qingwen and Li, Ci and Marta, Daniel Sim\~oes and Batool, Nazre and Folkesson, John},
    title     = {Human-Centric Autonomous Systems With {LLMs} for User Command Reasoning},
    booktitle = {Proceedings of the IEEE/CVF Winter Conference on Applications of Computer Vision (WACV) Workshops},
    month     = {January},
    year      = {2024},
    pages     = {988-994}
}

@InProceedings{park2024,
    author    = {Park, SungYeon and Lee, MinJae and Kang, JiHyuk and Choi, Hahyeon and Park, Yoonah and Cho, Juhwan and Lee, Adam and Kim, DongKyu},
    title     = {{VLAAD}: Vision and Language Assistant for Autonomous Driving},
    booktitle = {Proceedings of the IEEE/CVF Winter Conference on Applications of Computer Vision (WACV) Workshops},
    month     = {January},
    year      = {2024},
    pages     = {980-987}
}

@article{wen2023,
      title={On the Road with {GPT}-4{V}(ision): Early Explorations of Visual-Language Model on Autonomous Driving}, 
      author={Licheng Wen and Xuemeng Yang and Daocheng Fu and Xiaofeng Wang and Pinlong Cai and Xin Li and Tao Ma and Yingxuan Li and Linran Xu and Dengke Shang and Zheng Zhu and Shaoyan Sun and Yeqi Bai and Xinyu Cai and Min Dou and Shuanglu Hu and Botian Shi and Yu Qiao},
      year={2023},
      journal={arXiv preprint arXiv:2311.05332},
      archivePrefix={arXiv},
      primaryClass={cs.CV}
}

@inproceedings{woodruff2023,
author = {Woodruff, Allison and Shelby, Renee and Kelley, Patrick Gage and Rousso-Schindler, Steven and Smith-Loud, Jamila and Wilcox, Lauren},
title = {How Knowledge Workers Think Generative AI Will (Not) Transform Their Industries},
year = {2024},
isbn = {9798400703300},
publisher = {Association for Computing Machinery},
address = {New York, NY, USA},
url = {https://doi.org/10.1145/3613904.3642700},
doi = {10.1145/3613904.3642700},
booktitle = {Proceedings of the 2024 CHI Conference on Human Factors in Computing Systems},
articleno = {641},
numpages = {26},
location = {Honolulu, HI, USA},
series = {CHI '24}
}

@article{chen2023,
      title={Video{LLM}: Modeling Video Sequence with Large Language Models}, 
      author={Guo Chen and Yin-Dong Zheng and Jiahao Wang and Jilan Xu and Yifei Huang and Junting Pan and Yi Wang and Yali Wang and Yu Qiao and Tong Lu and Limin Wang},
      year={2023},
      journal={arXiv preprint arXiv:2305.13292},
      archivePrefix={arXiv},
      primaryClass={cs.CV},
}

@misc{jain2023,
    Author = {Ashesh Jain},
    url = {https://www.coram.ai/post/chatgpt-meets-video-security},
    Title = {Chat{GPT} Meets Video Security: A New Era of Intelligent Surveillance},
Year = {2023}}

@article{milano2023,
    author = {Silvia Milano and Joshua A. McGrane and Sabina Leonelli} ,
    title = {Large language models challenge the future of higher education},
    journal = {Nature Machine Intelligence} ,
    year = {2023}
}

@article{extance2023,
    Author = {Andy Extance},
    doi = {10.1038/d41586-023-03507-3},
    Title = {Chat{GPT} has entered the classroom: how {LLM}s could transform education},
journal = {Nature},
Year = {2023}}

@inproceedings{gadiraju2023,
author = {Gadiraju, Vinitha and Kane, Shaun and Dev, Sunipa and Taylor, Alex and Wang, Ding and Denton, Emily and Brewer, Robin},
title = {"{I} wouldn’t say offensive but...": Disability-Centered Perspectives on Large Language Models},
year = {2023},
isbn = {9798400701924},
publisher = {Association for Computing Machinery},
address = {New York, NY, USA},
url = {https://doi.org/10.1145/3593013.3593989},
doi = {10.1145/3593013.3593989},
pages = {205–216},
numpages = {12},
location = {Chicago, IL, USA},
booktitle = {FAccT '23: Proceedings of the 2023 ACM Conference on Fairness, Accountability, and Transparency}
}

@article{taheri2023,
      title={Breaking Barriers to Creative Expression: Co-Designing and Implementing an Accessible Text-to-Image Interface}, 
      author={Atieh Taheri and Mohammad Izadi and Gururaj Shriram and Negar Rostamzadeh and Shaun Kane},
      year={2023},
      journal={arXiv preprint arXiv:2309.02402},
      archivePrefix={arXiv},
      primaryClass={cs.HC}
}

@inproceedings{rose2023a,
  title = {Is Chat{GPT} a Good Teacher Coach? Measuring Zero-Shot Performance For Scoring and Providing Actionable Insights on Classroom Instruction},
  year = {2023},
  month = jun,
  author = {Wang, Rose and Demszky, Dorottya},
  booktitle = {18th Workshop on Innovative Use of NLP for Building Educational Applications},
  month_numeric = {6}
}

@inproceedings{hellas2023,
author = {Hellas, Arto and Leinonen, Juho and Sarsa, Sami and Koutcheme, Charles and Kujanp\"{a}\"{a}, Lilja and Sorva, Juha},
title = {Exploring the Responses of Large Language Models to Beginner Programmers’ Help Requests},
year = {2023},
isbn = {9781450399760},
publisher = {Association for Computing Machinery},
address = {New York, NY, USA},
url = {https://doi.org/10.1145/3568813.3600139},
doi = {10.1145/3568813.3600139},
booktitle = {Proceedings of the 2023 ACM Conference on International Computing Education Research - Volume 1},
pages = {93–105},
numpages = {13},
series = {ICER '23}
}

@article{johnson2023,
  title={Assessing the Accuracy and Reliability of {AI}-Generated Medical Responses: An Evaluation of the Chat-{GPT} Model},
  author={Douglas B Johnson and Rachel S Goodman and J Randall Patrinely and Cosby A Stone and Eli Zimmerman and Rebecca Rigel Donald and Sam S Chang and Sean T Berkowitz and Avni P Finn and Eiman Jahangir and Elizabeth A Scoville and Tyler Reese and Debra E. Friedman and Julie A. Bastarache and Yuri F van der Heijden and Jordan Wright and Nicholas Carter and Matthew R Alexander and Jennifer H Choe and Cody A Chastain and John Zic and Sara Horst and Isik Turker and Rajiv Agarwal and Evan C. Osmundson and Kamran Idrees and Colleen M. Kiernan and Chandrasekhar Padmanabhan and Christina Edwards Bailey and Cameron Schlegel and Lola B. Chambless and Mike Gibson and Travis J. Osterman and Lee Wheless},
  journal={Research Square},
  year={2023},
  url={https://api.semanticscholar.org/CorpusID:257437276}
}

@article{lyu2023,
  title={Translating radiology reports into plain language using chat{GPT} and {GPT}-4 with prompt learning: Promising results, limitations, and potential},
  author={Lyu, Qing and Tan, Josh and Zapadka, Mike E and Ponnatapuram, Janardhana and Niu, Chuang and Wang, Ge and Whitlow, Christopher T},
  journal={Vis. Comput. Ind. Biomed. Art},
  year={2023}
}

@article{kung2023,
  title={Performance of Chat{GPT} on {USMLE}: Potential for {AI}-assisted medical education using large language models},
  author={Kung, Tiffany H and Cheatham, Morgan and Medenilla, Arielle and Sillos, Czarina and De Leon, Lorie and Elepa{\~n}o, Camille and Madriaga, Maria and Aggabao, Rimel and Diaz-Candido, Giezel and Maningo, James and others},
  journal={PLoS digital health},
  volume={2},
  number={2},
  pages={e0000198},
  year={2023},
  publisher={Public Library of Science}
}

@article{Ezer2019,title={Trust Engineering for Human-{AI} Teams},author={Neta Ezer and S. Bruni and Yang Cai and S. Hepenstal and C. Miller and D. Schmorrow},journal={Proceedings of the Human Factors and Ergonomics Society Annual Meeting},year={2019},volume={63},pages={322 - 326},doi={10.1177/1071181319631264}}

@article{Yin2019,
title={Local privacy protection classification based on human-centric computing},
author={Chunyong Yin and Biao Zhou and Zhichao Yin and Jin Wang},
journal={Human-centric Computing and Information Sciences},
year={2019},
volume={9},
pages={1-14},
doi={10.1186/s13673-019-0195-4}
}

@article{kaissis2020,
  title={Secure, privacy-preserving and federated machine learning in medical imaging},
  author={Kaissis, Georgios A and Makowski, Marcus R and R{\"u}ckert, Daniel and Braren, Rickmer F},
  journal={Nature Machine Intelligence},
  volume={2},
  number={6},
  pages={305--311},
  year={2020},
  publisher={Nature Publishing Group UK London}
}

@article{Admin2022,
  title={A survey on cyber security meets artificial intelligence: {AI}--driven cyber security},
  author={Samyuktha, SP and Kavitha, P and Kshaya, VA and Shalini, P and Ramya, R},
  journal={Journal of Cognitive Human-Computer Interaction},
  volume={2},
  number={2},
  pages={50--55},
  year={2022}
}

@article{lepri2021,
  title={Ethical machines: The human-centric use of artificial intelligence},
  author={Lepri, Bruno and Oliver, Nuria and Pentland, Alex},
  journal={IScience},
  volume={24},
  number={3},
  year={2021},
  publisher={Elsevier}
}

@inproceedings{hacker2023,
  title={Regulating Chat{GPT} and other large generative {AI} models},
  author={Hacker, Philipp and Engel, Andreas and Mauer, Marco},
  booktitle={Proceedings of the 2023 ACM Conference on Fairness, Accountability, and Transparency},
  pages={1112--1123},
  year={2023}
}

@article{ullah2023,
  title={Privacy preserving large language models: Chat{GPT} case study based vision and framework},
  author={Ullah, Imdad and Hassan, Najm and Gill, Sukhpal Singh and Suleiman, Basem and Ahanger, Tariq Ahamed and Shah, Zawar and Qadir, Junaid and Kanhere, Salil S},
  journal={IET Blockchain},
  volume={4},
  pages={706--724},
  year={2024},
  publisher={Wiley Online Library}
}

@article{gupta2023,
  title={From chat{GPT} to threat{GPT}: Impact of generative {AI} in cybersecurity and privacy},
  author={Gupta, Maanak and Akiri, CharanKumar and Aryal, Kshitiz and Parker, Eli and Praharaj, Lopamudra},
  journal={IEEE Access},
  year={2023},
  publisher={IEEE}
}

@article{de2023,
  title={Chat{GPT} and the rise of large language models: the new {AI}-driven infodemic threat in public health},
  author={De Angelis, Luigi and Baglivo, Francesco and Arzilli, Guglielmo and Privitera, Gaetano Pierpaolo and Ferragina, Paolo and Tozzi, Alberto Eugenio and Rizzo, Caterina},
  journal={Frontiers in Public Health},
  volume={11},
  pages={1166120},
  year={2023},
  publisher={Frontiers}
}

@article{thapa2023,
  title={Chat{GPT}, bard, and large language models for biomedical research: opportunities and pitfalls},
  author={Thapa, Surendrabikram and Adhikari, Surabhi},
  journal={Annals of Biomedical Engineering},
  volume={51},
  number={12},
  pages={2647--2651},
  year={2023},
  publisher={Springer}
}

@article{sebastian2023,
title={Privacy and Data Protection in ChatGPT and Other {AI} Chatbots: Strategies for Securing User Information},
  author={Sebastian, Glorin},
  journal={Available at SSRN 4454761},
 year={2023}
}

@inproceedings{liu2023humans,
author = {Liu, Minghao and Wei, Jiaheng and Liu, Yang and Davis, James},
title = {Human and AI perceptual differences in image classification errors},
year = {2025},
isbn = {978-1-57735-897-8},
publisher = {AAAI Press},
url = {https://doi.org/10.1609/aaai.v39i13.33568},
doi = {10.1609/aaai.v39i13.33568},
booktitle = {Proceedings of the Thirty-Ninth AAAI Conference on Artificial Intelligence and Thirty-Seventh Conference on Innovative Applications of Artificial Intelligence and Fifteenth Symposium on Educational Advances in Artificial Intelligence},
articleno = {1595},
numpages = {9},
series = {AAAI'25/IAAI'25/EAAI'25}
}

@article{liu2023tag,
  title={Tag-based annotation creates better avatars},
  author={Liu, Minghao and Cheng, Zeyu and Sang, Shen and Liu, Jing and Davis, James},
  journal={arXiv preprint arXiv:2302.07354},
  year={2023}
}

@article{ngo2023tag,
  title={Tag-Based Annotation for Avatar Face Creation},
  author={Ngo, An and Phelps, Daniel and Lai, Derrick and Wong, Thanyared and Mathias, Lucas and Shivamurthy, Anish and Ajmal, Mustafa and Liu, Minghao and Davis, James},
  journal={arXiv preprint arXiv:2308.12642},
  year={2023}
}

@article{mittelstadt2016,
    author = {Mittelstadt, Brent and Allo, Patrick and Taddeo, Mariarosaria and Wachter, Sandra and Floridi, Luciano},
    title = {The ethics of algorithms: Mapping the debate},
    journal = {Big Data and Society},
    year = {2016}
}

@inproceedings{goodfellow2015,
    author = {Ian J. Goodfellow and Jonathon Shlens and Christian Szegedy},
    title = {Explaining and Harnessing Adversarial Examples},
    booktitle = {International Conference on Learning Representations},
    year = {2015}
}

@article{rodrigues2020,
    author = {Rowena Rodrigues},
    title = {Legal and human rights issues of AI: Gaps, challenges and vulnerabilities},
    journal = {Journal of Responsible Technology},
    year = {2020}
}

@article{OpenAI2023ChatGPT,
  title={Gpt-4 technical report},
  author={Achiam, Josh and Adler, Steven and Agarwal, Sandhini and Ahmad, Lama and Akkaya, Ilge and Aleman, Florencia Leoni and Almeida, Diogo and Altenschmidt, Janko and Altman, Sam and Anadkat, Shyamal and others},
  journal={arXiv preprint arXiv:2303.08774},
  year={2023}
}

@article{huang2023bias,
  title={Bias testing and mitigation in {LLM}-based code generation},
  author={Huang, Dong and M. Zhang, Jie and Bu, Qingwen and Xie, Xiaofei and Chen, Junjie and Cui, Heming},
  journal={ACM Transactions on Software Engineering and Methodology},
  volume={35},
  number={1},
  pages={1--31},
  year={2025},
  publisher={ACM New York, NY}
}

@article{meyer2023chatgpt,
  title={Chat{GPT} and large language models in academia: opportunities and challenges},
  author={Meyer, Jesse G and Urbanowicz, Ryan J and Martin, Patrick CN and O’Connor, Karen and Li, Ruowang and Peng, Pei-Chen and Bright, Tiffani J and Tatonetti, Nicholas and Won, Kyoung Jae and Gonzalez-Hernandez, Graciela and others},
  journal={BioData Mining},
  volume={16},
  number={1},
  pages={20},
  year={2023},
  publisher={Springer}
}

@inproceedings{kotek2023,
author = {Kotek, Hadas and Dockum, Rikker and Sun, David},
title = {Gender bias and stereotypes in Large Language Models},
year = {2023},
isbn = {9798400701139},
publisher = {Association for Computing Machinery},
address = {New York, NY, USA},
url = {https://doi.org/10.1145/3582269.3615599},
doi = {10.1145/3582269.3615599},
booktitle = {Proceedings of The ACM Collective Intelligence Conference},
pages = {12–24},
numpages = {13},
location = {Delft, Netherlands},
series = {CI '23}
}

@article{li2023bias,
  title={A survey on fairness in large language models},
  author={Li, Yingji and Du, Mengnan and Song, Rui and Wang, Xin and Wang, Ying},
  journal={arXiv preprint arXiv:2308.10149},
  year={2023}
}

@article{gallegos2023bias,
    title = "Bias and Fairness in Large Language Models: A Survey",
    author = "Gallegos, Isabel O.  and
      Rossi, Ryan A.  and
      Barrow, Joe  and
      Tanjim, Md Mehrab  and
      Kim, Sungchul  and
      Dernoncourt, Franck  and
      Yu, Tong  and
      Zhang, Ruiyi  and
      Ahmed, Nesreen K.",
    journal = "Computational Linguistics",
    volume = "50",
    number = "3",
    month = sep,
    year = "2024",
    address = "Cambridge, MA",
    publisher = "MIT Press",
    url = "https://aclanthology.org/2024.cl-3.8/",
    doi = "10.1162/coli_a_00524",
    pages = "1097--1179",
}

@inproceedings{ohi2024bias,
  title={Likelihood-based mitigation of evaluation bias in large language models},
  author={Ohi, Masanari and Kaneko, Masahiro and Koike, Ryuto and Loem, Mengsay and Okazaki, Naoaki},
  booktitle={Findings of the Association for Computational Linguistics: ACL 2024},
  pages={3237--3245},
  year={2024}
}

@article{bi2023bias,
  title={A Group Fairness Lens for Large Language Models},
  author={Bi, Guanqun and Shen, Lei and Xie, Yuqiang and Cao, Yanan and Zhu, Tiangang and He, Xiaodong},
  journal={arXiv preprint arXiv:2312.15478},
  year={2023}
}

@misc{mit2023law,
  author = {Walsh, Dylan},
  title = {Legal issues presented by generative {AI}},
  year = {2023},
  howpublished = {\url{https://mitsloan.mit.edu/ideas-made-to-matter/legal-issues-presented-generative-ai}},
  note = {Accessed: 2024-02-28}
}

@misc{gdpr2023law,
  author = {Marcos, Henrique and Pullin, Melina},
  title = {LARGE LANGUAGE MODELS AND EU DATA PROTECTION: MAPPING (SOME) OF THE PROBLEMS},
  year = {2023},
  howpublished = {\url{https://digi-con.org/large-language-models-and-eu-data-protection-mapping-some-of-the-problems/}},
  note = {Accessed: 2024-02-28}
}

@misc{MarioKart8Deluxe2017,
  title = {Mariokart 8 Deluxe},
  author = {Nintendo},
  address = {[Redmond, WA]},
  publisher = {Nintendo},
  year = {2017},
  note = {Video game}
}

@article{wang2023empowering,
  title={Empowering autonomous driving with large language models: A safety perspective},
  author={Wang, Yixuan and Jiao, Ruochen and Zhan, Sinong Simon and Lang, Chengtian and Huang, Chao and Wang, Zhaoran and Yang, Zhuoran and Zhu, Qi},
  journal={arXiv preprint arXiv:2312.00812},
  year={2023}
}

@inproceedings{wang2023drive,
  title={Drive anywhere: Generalizable end-to-end autonomous driving with multi-modal foundation models},
  author={Wang, Tsun-Hsuan and Maalouf, Alaa and Xiao, Wei and Ban, Yutong and Amini, Alexander and Rosman, Guy and Karaman, Sertac and Rus, Daniela},
  booktitle={2024 IEEE International Conference on Robotics and Automation (ICRA)},
  pages={6687--6694},
  year={2024},
  organization={IEEE}
}

@inproceedings{Geiger2012CVPR,
  author = {Andreas Geiger and Philip Lenz and Raquel Urtasun},
  title = {Are we ready for Autonomous Driving? The KITTI Vision Benchmark Suite},
  booktitle = {Conference on Computer Vision and Pattern Recognition (CVPR)},
  year = {2012}
}

@InProceedings{Zhang2025MMRLHF,
  title = 	 {{MM}-{RLHF}: The Next Step Forward in Multimodal {LLM} Alignment},
  author =       {Zhang, Yifan and Yu, Tao and Tian, Haochen and Fu, Chaoyou and Li, Peiyan and Zeng, Jianshu and Xie, Wulin and Shi, Yang and Zhang, Huanyu and Wu, Junkang and Wang, Xue and Hu, Yibo and Wen, Bin and Gao, Tingting and Zhang, Zhang and Yang, Fan and Zhang, Di and Wang, Liang and Jin, Rong},
  booktitle = 	 {Proceedings of the 42nd International Conference on Machine Learning},
  pages = 	 {76625--76654},
  year = 	 {2025},
  editor = 	 {Singh, Aarti and Fazel, Maryam and Hsu, Daniel and Lacoste-Julien, Simon and Berkenkamp, Felix and Maharaj, Tegan and Wagstaff, Kiri and Zhu, Jerry},
  volume = 	 {267},
  series = 	 {Proceedings of Machine Learning Research},
  month = 	 {13--19 Jul},
  publisher =    {PMLR},
  url = 	 {https://proceedings.mlr.press/v267/zhang25cs.html},
}

@article{fourney2024magenticone,
  title={Magentic-one: A generalist multi-agent system for solving complex tasks},
  author={Fourney, Adam and Bansal, Gagan and Mozannar, Hussein and Tan, Cheng and Salinas, Eduardo and Niedtner, Friederike and Proebsting, Grace and Bassman, Griffin and Gerrits, Jack and Alber, Jacob and others},
  journal={arXiv preprint arXiv:2411.04468},
  year={2024}
}

@article{motwani2024malt,
  title={Malt: Improving reasoning with multi-agent llm training},
  author={Motwani, Sumeet Ramesh and Smith, Chandler and Das, Rocktim Jyoti and Rafailov, Rafael and Laptev, Ivan and Torr, Philip HS and Pizzati, Fabio and Clark, Ronald and de Witt, Christian Schroeder},
  journal={COLM 2025},
  year={2025}
}

@inproceedings{yu2025table,
  title={Table-critic: A multi-agent framework for collaborative criticism and refinement in table reasoning},
  author={Yu, Peiying and Chen, Guoxin and Wang, Jingjing},
  booktitle={Proceedings of the 63rd Annual Meeting of the Association for Computational Linguistics (Volume 1: Long Papers)},
  pages={17432--17451},
  year={2025}
}

@inproceedings{vats2026guideline,
  title={Guideline-Consistent Segmentation via Multi-Agent Refinement},
  author={Vats, Vanshika and Rathee, Ashwani and Davis, James},
  booktitle={Proceedings of the AAAI Conference on Artificial Intelligence},
  volume={40},
  number={12},
  pages={9612--9620},
  year={2026}
}

@article{yao2024tau,
  title={$\tau$-bench: A Benchmark for Tool-Agent-User Interaction in Real-World Domains},
  author={Yao, Shunyu and Shinn, Noah and Razavi, Pedram and Narasimhan, Karthik},
  journal={ICLR},
  year={2025}
}

@article{xie2024osworld,
  title={Osworld: Benchmarking multimodal agents for open-ended tasks in real computer environments},
  author={Xie, Tianbao and Zhang, Danyang and Chen, Jixuan and Li, Xiaochuan and Zhao, Siheng and Cao, Ruisheng and Hua, Toh J and Cheng, Zhoujun and Shin, Dongchan and Lei, Fangyu and others},
  journal={Advances in Neural Information Processing Systems},
  volume={37},
  pages={52040--52094},
  year={2024}
}

@article{abhyankar2025osworld,
  title={Osworld-human: Benchmarking the efficiency of computer-use agents},
  author={Abhyankar, Reyna and Qi, Qi and Zhang, Yiying},
  journal={arXiv preprint arXiv:2506.16042},
  year={2025}
}

@article{xu2026theagentcompany,
  title={Theagentcompany: benchmarking llm agents on consequential real world tasks},
  author={Xu, Frank Fangzheng and Song, Yufan and Li, Boxuan and Tang, Yuxuan and Jain, Kritanjali and Bao, Mengxue and Wang, Zora and Zhou, Xuhui and Guo, Zhitong and Cao, Murong and others},
  journal={Advances in Neural Information Processing Systems},
  volume={38},
  year={2025}
}

@article{jiang2025medagentbench,
  title={MedAgentBench: a virtual EHR environment to benchmark medical LLM agents},
  author={Jiang, Yixing and Black, Kameron C and Geng, Gloria and Park, Danny and Zou, James and Ng, Andrew Y and Chen, Jonathan H},
  journal={{NEJM AI}},
  volume={2},
  number={9},
  pages={AIdbp2500144},
  year={2025},
  publisher={Massachusetts Medical Society}
}

@inproceedings{schoembs2026conversation,
author = {Sch\"{o}mbs, Sarah and Zhang, Yan and Goncalves, Jorge and Johal, Wafa},
title = {From Conversation to Orchestration: HCI Challenges and Opportunities in Interactive Multi-Agentic Systems},
year = {2026},
isbn = {9798400721786},
publisher = {Association for Computing Machinery},
address = {New York, NY, USA},
url = {https://doi.org/10.1145/3765766.3765795},
doi = {10.1145/3765766.3765795},
booktitle = {Proceedings of the 13th International Conference on Human-Agent Interaction},
pages = {158–168},
numpages = {11},
location = {
},
series = {HAI '25}
}

@inproceedings{masters2025orchestrating,
author = {Masters, Charlie and Vellanki, Advaith and Shangguan, Jiangbo and Kultys, Bart and Gilmore, Jonathan and Moore, Alastair and Albrecht, Stefano},
title = {Orchestrating Human-AI Teams: The Manager Agent as a Unifying Research Challenge},
year = {2025},
isbn = {9798400722752},
publisher = {Association for Computing Machinery},
address = {New York, NY, USA},
url = {https://doi.org/10.1145/3772429.3772439},
doi = {10.1145/3772429.3772439},
booktitle = {Proceedings of the 2025 7th International Conference on Distributed Artificial Intelligence},
pages = {91–107},
numpages = {17},
location = {
},
series = {DAI '25}
}

@article{liu2025survey,
  title={A survey of direct preference optimization},
  author={Liu, Shunyu and Fang, Wenkai and Hu, Zetian and Zhang, Junjie and Zhou, Yang and Zhang, Kongcheng and Tu, Rongcheng and Lin, Ting-En and Huang, Fei and Song, Mingli and others},
  journal={arXiv preprint arXiv:2503.11701},
  year={2025}
}

@misc{Purdy2016,
  author       = {Purdy, Mark and Daugherty, Paul},
  title        = {Why Artificial Intelligence is the Future of Growth},
  howpublished = {Accenture Institute for High Performance},
  year         = {2016}
}

@article{choudhury2024large,
  title={Large language models and user trust: consequence of self-referential learning loop and the deskilling of health care professionals},
  author={Choudhury, Avishek and Chaudhry, Zaira},
  journal={Journal of Medical Internet Research},
  volume={26},
  pages={e56764},
  year={2024},
  publisher={JMIR Publications Toronto, Canada}
}

\end{document}